\documentclass{article}

\usepackage[main, final]{neurips_2026}

\usepackage[utf8]{inputenc}
\usepackage[T1]{fontenc}
\usepackage{hyperref}
\usepackage{url}
\usepackage{booktabs}
\usepackage{amsfonts}
\usepackage{nicefrac}
\usepackage{microtype}
\usepackage{xcolor}

\usepackage{amsmath}
\usepackage{amssymb}
\usepackage{graphicx}
\usepackage{subcaption}
\usepackage{multirow}
\usepackage{algorithm}
\usepackage{algorithmic}

\newcommand{\tisae}{TI-SAE}
\newcommand{\predselector}{prediction selector}
\newcommand{\Hgrid}{\mathbf{H}_{\text{grid}}}
\newcommand{\Hhat}{\hat{\mathbf{H}}_{\text{grid}}}
\newcommand{\Ahat}{\hat{\mathbf{A}}}
\newcommand{\reals}{\mathbb{R}}
\newcommand{\loss}{\mathcal{L}}

\title{Temporal Island Sparse Autoencoders for Interpreting Clinical Time-Series Models}

\author{%
  \textbf{Muhan Yeo}\textsuperscript{1,2} \quad
  \textbf{Bong Gyun Kang}\textsuperscript{1} \quad
  \textbf{HyunGi Kim}\textsuperscript{1,3} \quad
  \textbf{Sungroh Yoon}\textsuperscript{1}\thanks{Corresponding author.} \\
  \\
  \textsuperscript{1}Department of Electrical and Computer Engineering and Interdisciplinary Program in \\ Artificial Intelligence, Seoul National University, Seoul, Republic of Korea \\
  \textsuperscript{2}McWilliams School of Biomedical Informatics, \\ University of Texas Health Science Center at Houston, Houston, TX, USA \\
  \textsuperscript{3}UNIST InnoCORE AI-Space Solar Initiative, \\ Ulsan National Institute of Science and Technology (UNIST), Ulsan, Republic of Korea \\
  \\
  \texttt{Muhan.Yeo@uth.tmc.edu} \\
  \texttt{\{luckypanda,rlagusrl0128,sryoon\}@snu.ac.kr}
}

\begin{document}

\maketitle

\begin{abstract}
Deep learning models for electronic health records achieve strong predictive performance but provide little insight into \emph{what clinical concepts} they represent internally.
Post-hoc attribution methods produce per-cell importance scores that lack temporal coherence, cross-variable structure, and reusability across patients.
We introduce the \textbf{Temporal Island Sparse Autoencoder} (\tisae{}), which learns a backbone-specific, patient-shared \emph{clinical concept dictionary} from a frozen predictive model's representations.
Each dictionary entry (latent) is defined by learned temporal Gaussian islands---specifying \emph{when} the concept is active---and a sparse variable-composition vector---specifying \emph{which variables} are involved.
A decoupled selector then identifies which dictionary entries drive each patient's prediction, separating representation learning from attribution.
We evaluate \tisae{} across five backbone architectures spanning different representation strategies on MIMIC-III and MIMIC-IV (three prediction tasks each).
\tisae{} produces structured clinical concepts that go beyond transient attribution heatmaps while achieving strong performance on standard faithfulness benchmarks.

\end{abstract}

\section{Introduction}
\label{sec:intro}

Deep learning models for electronic health records (EHR) have achieved strong predictive performance in clinical tasks including mortality, decompensation, and readmission~\citep{rajkomar2018scalable, harutyunyan2019multitask, tipirneni2022self}.
However, these models remain opaque. For clinical models to move toward actual insight, clinicians need to understand a model's behavior in terms they already use---clinical patterns with temporal extent and cross-variable structure~\citep{tonekaboni2019clinicians, ghassemi2021false}.

Post-hoc explanation methods such as Integrated Gradients~\citep{sundararajan2017axiomatic}, SHAP~\citep{lundberg2017unified}, Dynamask~\citep{crabbe2021explaining}, and TimeX~\citep{queen2023encoding} produce importance scores over the time$\times$variable input grid.
These attributions answer \emph{which individual inputs matter} but not \emph{what clinical concepts the model uses}.
Point attributions lack (a)~temporal and multivariable structure, which are prerequisites for clinical interpretation, and (b)~compositionality into concepts that generalize across patients.
Moreover, most time-series explanation methods assume dense, regularly-sampled inputs, yet clinical event sequences are sparse and irregular, forcing ad-hoc adaptations.

Sparse autoencoders (SAEs) offer a principled approach to this problem by decomposing a model's internal representations into sparse, interpretable features via dictionary learning~\citep{bricken2023monosemanticity, templeton2024scaling}.
However, standard SAEs often recover token-specific or highly local features, which are not well-aligned with temporally extended clinical concepts~\citep{bhalla2026temporal}. 
Moreover, clinical episodes span hundreds of variables over extended time windows, and extracting a coherent concept from raw patient trajectories is infeasible without knowing which variables and what time window to examine.
This motivates an SAE architecture that structurally encodes spatiotemporal composition.

\begin{figure}[t]
    \includegraphics[width=1.0\textwidth]{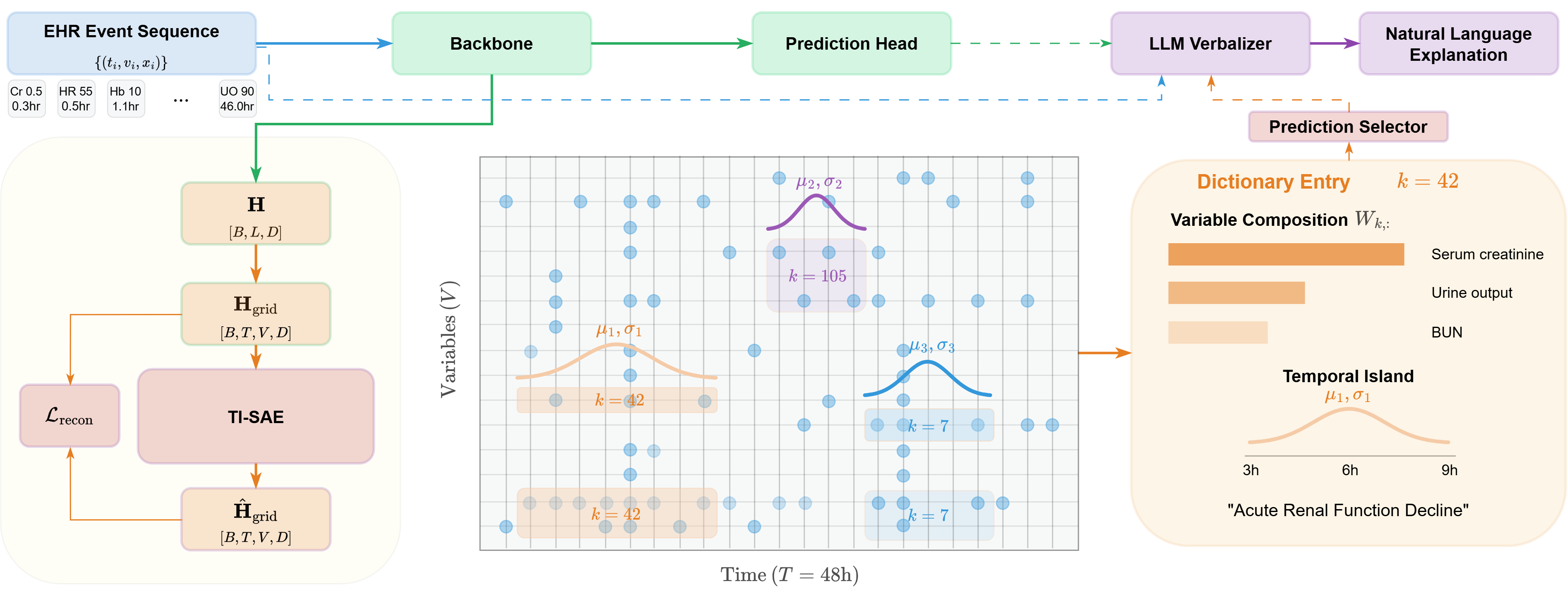}
    \centering
    \caption{
    \textbf{Overview of the Temporal Island SAE (\tisae{}).}
    A frozen backbone transforms an EHR event sequence into a hidden representation $\mathbf{H}[B,L,D]$, which is then scattered into a dense representation grid $\Hgrid[B,T,V,D]$.
    \tisae{} learns a dictionary of $K$ latents, each with $M$ temporal Gaussian islands (colored curves) defining when the concept is active, and sparse variable-composition weights $\mathbf{w}_k$ defining which variables are involved.
    Each island corresponds to an interpretable clinical concept with explicit temporal extent and variable structure.
    }
    \label{fig:overview}
\end{figure}

We propose the \textbf{Temporal Island Sparse Autoencoder} (\tisae{}), which learns a backbone-specific, patient-shared clinical concept dictionary---a finite vocabulary of the latent patterns reused across patients (Figure~\ref{fig:overview}).
Each dictionary entry (latent $k$) has two components: (a)~$M$ learned \emph{temporal Gaussian islands} with centers~$\mu_{k,m}$ and widths~$\sigma_{k,m}$ that define \emph{when} the concept is active, and (b)~a sparse variable-composition vector $\mathbf{w}_k \in \reals^V$ that defines \emph{which variables} are involved.
The \emph{island} structure, our key architectural novelty, encodes spatiotemporal organization directly in the latent space via temporally localized Gaussian windows and sparse variable composition.

We evaluate \tisae{} across five backbone architectures on MIMIC-III and MIMIC-IV with three prediction tasks each.
\tisae{} achieves the highest faithfulness (AOPC) on four of five backbones, while simultaneously providing a decomposable clinical concept dictionary that no baseline method can produce.
The learned dictionary is judged clinically plausible by four clinicians (mean Likert 4.07--4.40), transfers to unseen data, and produces superior concept-level explanations in both LLM-judge and clinician evaluations.
Code is available at \url{https://github.com/MuhanY/TI-SAE}.

\section{Related work}
\label{sec:related}

\subsection{EHR model architectures}
\label{sec:related_ehr}

Each clinical event carries a variable identity, a measured value, and a timestamp.
Code-level models such as BEHRT~\citep{li2020behrt},
CLMBR~\citep{steinberg2021language}, and MOTOR~\citep{steinberg2024motor} encode only discrete medical concept codes without measurement values, often targeting long outpatient trajectories.
For inpatient settings where measurement magnitudes and their temporal evolution are central to prediction, models must explicitly represent all three components.

\paragraph{Event representation.}
Early approaches to irregular clinical time series operated on dense time$\times$variable grids with imputation or learned decay for unobserved cells~\citep{che2018grud}.
Modern architectures instead process variable-length event sequences directly, adopting different strategies for representing each event.
Continuous value embedding (CVE) methods such as STraTS~\citep{tipirneni2022self} embed raw scalar values and timestamps through learned MLPs, producing a single token per event; HyMaTE~\citep{mottalib2025hymate} extends this approach with a hybrid Mamba/transformer encoder.
Discrete methods quantize values into bins:
ETHOS~\citep{renc2024ethos} serializes each event into up to seven tokens, while EHRMamba~\citep{fallahpour2024ehrmamba} inserts time-interval tokens between events.
Grid-native architectures such as DuETT~\citep{labach2023duett} and BAT~\citep{devries2025bat} operate directly on the time$\times$variable matrix with factored bi-axial attention.

\begin{table}[t]
\caption{\textbf{Comparison of explanation methods along structural properties.} \emph{Temporal structure}: attributions are organized in time, not
independent per cell. \emph{Decomposable}: explanations decompose into components that generalize across patients. \emph{Post-hoc}: no
backbone modification. \emph{No concept-label supervision}: does not require external concept-label annotation.}
\label{tab:method_comparison}
\centering
\small
\begin{tabular}{lcccc}
\toprule
Method family & {Temporal struct.} & {Decomposable} & {Post-hoc} & {No concept sup.} \\
\midrule
Attention & & & \checkmark & \checkmark \\
Gradient-based & (\checkmark) & & \checkmark & \checkmark \\
Perturbation-based & (\checkmark) & & \checkmark & \checkmark \\
Learned selection & (\checkmark) & & \checkmark & \checkmark \\
Concept bottleneck & & \checkmark & & \\
\midrule
\textbf{\tisae{} (ours)} & \checkmark & \checkmark & \checkmark & \checkmark \\
\bottomrule
\end{tabular}
\end{table}

\subsection{Post-hoc explainability for time series}
\label{sec:related_posthoc}

\paragraph{Gradient-based methods.}
Integrated Gradients~\citep{sundararajan2017axiomatic}, DeepLIFT~\citep{shrikumar2017learning}, and GradSHAP assign per-feature importance scores via backpropagation.
Applied to time series, they produce $(T \times V)$ attribution maps with no temporal structure---each cell is scored independently.
TSR~\citep{ismail2020benchmarking} combines gradient methods via two-step saliency rescaling but the output remains a flat attribution map.

\paragraph{Perturbation-based methods.}
KernelSHAP~\citep{lundberg2017unified} and TimeSHAP~\citep{bento2021timeshap} estimate feature importance by sampling input subsets and observing output changes; TimeSHAP adapts this for sequences by grouping time steps, but produces flat cell-level attributions.
Dynamask~\citep{crabbe2021explaining} learns continuous perturbation masks with temporal smoothness regularization.
FIT~\citep{tonekaboni2020went} uses KL-divergence shifts in the predictive distribution; WinIT~\citep{leung2023temporal} aggregates feature-removal scores over time windows.
ContraLSP~\citep{liu2024explaining} adds contrastive regularization to mask optimization.
These methods improve \emph{how} attributions are computed but do not decompose explanations into structured concepts.

\paragraph{Learned selection.}
TimeX~\citep{queen2023encoding} trains an interpretable surrogate model via model behavior consistency.
TimeX++~\citep{liu2024timex++} reformulates selection through an information bottleneck.
L2X~\citep{chen2018learning} introduces Gumbel-softmax selection for instance-wise feature importance.
They operate at the cell level without concept decomposition.

\subsection{Sparse autoencoders for interpretability}
\label{sec:related_sae}

Dictionary learning for mechanistic interpretability has been transformative in understanding language models.
\citet{bricken2023monosemanticity} showed that SAEs decompose MLP activations into monosemantic features, with \citet{templeton2024scaling} scaling this to production models.
Gated SAEs~\citep{rajamanoharan2024improving} improve reconstruction--sparsity tradeoffs via gated encoding.
TopK SAEs~\citep{gao2025scaling} enforce fixed sparsity via top-$k$ activation selection.
\citet{zhang2025large} extended SAEs to vision-language models, validating the approach beyond text.
Most recently, \citet{mishra2026dissecting} applied SAEs to time-series foundation models, revealing causal feature hierarchies.
No prior work applies SAEs to clinical time-series models with explicit spatiotemporal structure. 

\subsection{Concept-based explanations}
\label{sec:related_concepts}

TCAV~\citep{kim2018interpretability} tests whether models use human-defined concepts via directional derivatives.
ACE~\citep{ghorbani2019towards} automates concept extraction via superpixel clustering.
Concept Bottleneck Models (CBMs)~\citep{koh2020concept} enforce a concept layer during training, with post-hoc extensions removing the training requirement~\citep{yuksekgonul2023posthoc}.

\tisae{} differs from CBMs in several ways: (a)~it is fully post-hoc on a frozen backbone (no architectural modification or retraining), (b)~concepts are \emph{discovered} from data via dictionary learning rather than defined a priori, (c)~each concept has learned temporal extent (Gaussian islands) rather than being a static attribute, and (d)~concept labels are post-hoc validation, not the concept definition itself.
Table~\ref{tab:method_comparison} summarizes these structural distinctions.

\section{Method}
\label{sec:method}

\subsection{Problem setup and notation}
\label{sec:notation}

We consider inpatient EHR episodes consisting of irregular multivariate observations $\{(t_i, v_i, x_i)\}_{i=1}^{L}$, where $t_i \in [0, T)$ is the timestamp, $v_i \in \{1, \dots, V\}$ is the variable index, and $x_i \in \reals$ is the observed value.
The clinical outcome is a binary label $y \in \{0, 1\}$.
Variables are organized into $C$ clinical systems (e.g., cardiovascular, renal, respiratory) via a known ontology mapping $\phi: \{1, \dots, V\} \to \{1, \dots, C\}$, where some variables are shared across two channels.

A pretrained backbone $f_\theta$ transforms events into contextual embeddings $\mathbf{H} \in \reals^{B \times L \times D}$, which are pooled into a patient-level summary $\mathbf{q} \in \reals^D$ and
passed to a prediction head $\hat{y} = \sigma(g_\psi(\mathbf{q}))$.
The embeddings are organized into a dense time$\times$variable grid $\Hgrid \in \reals^{B \times T \times V \times D}$, with an observation mask $\mathbf{O} \in \{0,1\}^{B \times T \times V}$ indicating which cells contain real observations.
The backbone is frozen throughout; \tisae{} operates entirely on
$\Hgrid$.

\subsection{TI-SAE architecture}
\label{sec:architecture}

\begin{figure}[t]
    \includegraphics[width=1.0\textwidth]{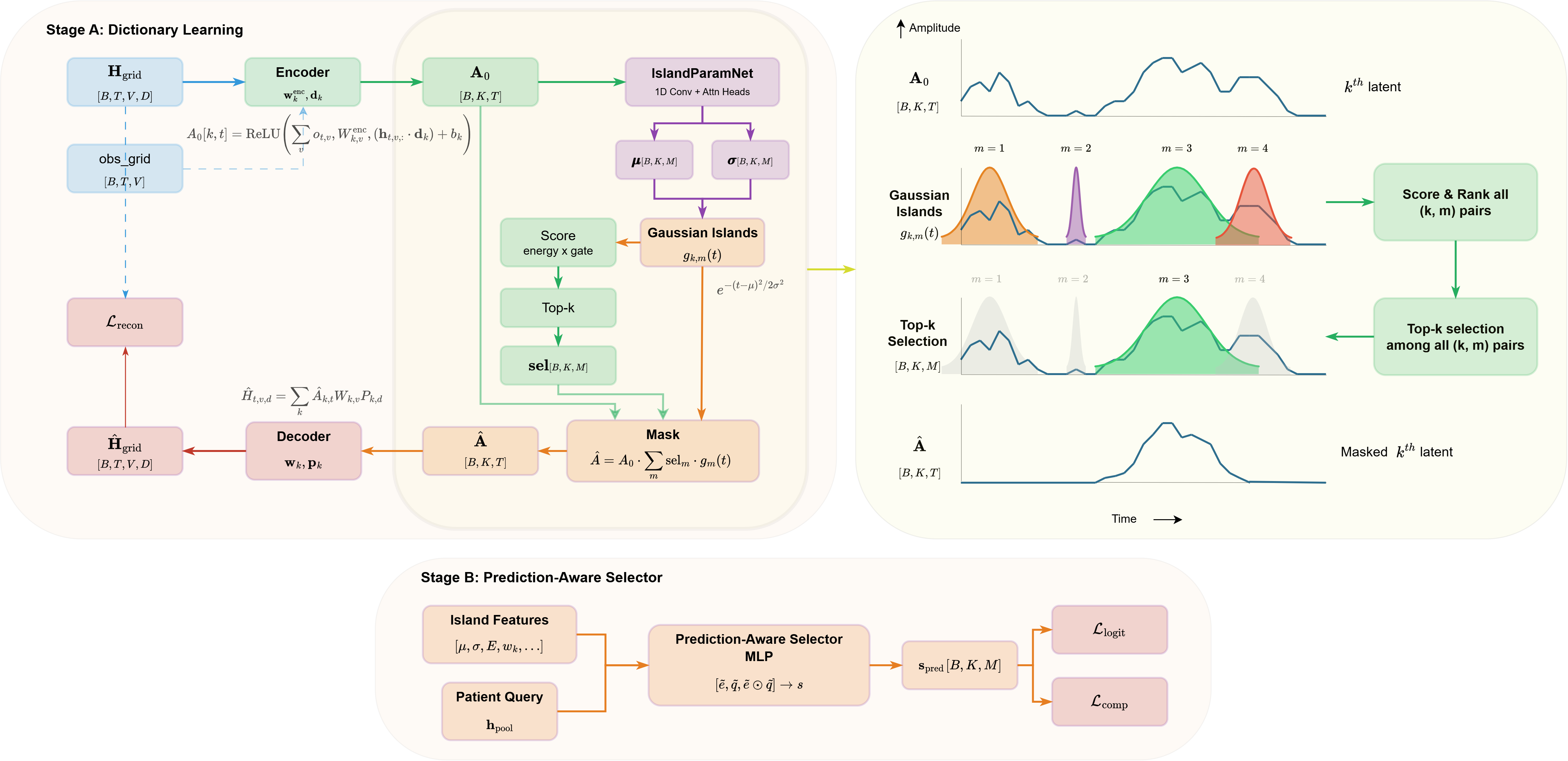}
    \centering
    \caption{
    \textbf{\tisae{} architecture.}
    The encoder maps $\Hgrid$ to activation traces $\mathbf{A}_0[B,K,T]$.
    The IslandParamNet produces $M$ Gaussian islands per latent with learned centers $\mu$ and widths $\sigma$.
    Energy-gated selection picks top-k islands, defining temporal masks.
    The decoder reconstructs $\Hhat$ via sparse routing weights $\mathbf{W}$ and atoms $\mathbf{P}$.
    The prediction-aware selector is trained separately on the frozen latent basis to identify prediction-relevant islands.
    }
    \label{fig:architecture}
\end{figure}

\tisae{} decomposes $\Hgrid$ into $K$ latents, each with $M$ temporal islands.
The architecture has four components: encoder, island parameter network, island selector, and decoder (Figure~\ref{fig:architecture}).

\subsubsection{Encoder}
\label{sec:encoder}

The encoder maps $\Hgrid$ to pre-selection activation traces $\mathbf{A}_0 \in \reals^{B \times K \times T}$. Each latent $k$ has a sparse variable-routing vector $\mathbf{w}^{\text{enc}}_k \in \reals^V$ (top-$\kappa_{\text{enc}}$, row-normalized, non-negative) and a projection vector $\mathbf{d}_k \in \reals^D$:
\begin{equation}
    A_0[b,k,t] = \text{ReLU}\!\left( \sum_{v=1}^{V} O[b,t,v] \cdot W^{\text{enc}}[k,v] \cdot \left(\Hgrid[b,t,v,:] \cdot \mathbf{d}_k\right) + b_k \right)
    \label{eq:encoder}
\end{equation}
Each latent attends to a sparse subset of variables (via $\mathbf{w}^{\text{enc}}_k$), projected onto $\mathbf{d}_k$ to produce a scalar activation.

\subsubsection{Temporal Gaussian islands}
\label{sec:islands}

The \textbf{Island Parameter Network} transforms activation traces into per-latent temporal parameters.
For each latent $k$ and island $m \in \{1, \dots, M\}$, it produces a center $\mu_{k,m}$ and width $\sigma_{k,m}$:
\begin{align}
    \mathbf{h}_k(t) &= \text{Conv1D}(\mathbf{A}_0[b,k,:]) \in \reals^{H \times T} \label{eq:island_h} \\
    \mu_{k,m} &= \sum_{t=0}^{T-1} \text{softmax}_t\!\big((\boldsymbol{\ell}_\mu^{(m)})^\top \mathbf{h}_k(t)\big) \cdot t \label{eq:mu} \\
    g_{k,m}(t) &= \exp\!\left(-\frac{(t - \mu_{k,m})^2}{2\sigma_{k,m}^2}\right) \label{eq:gaussian}
\end{align}
A 1D convolutional feature extractor maps each activation trace to a temporal feature map $\mathbf{h}_k(t)$ (Eq.~\ref{eq:island_h}).
Each island head $m$ applies a pointwise projection $\boldsymbol{\ell}_\mu^{(m)}$ to produce attention logits over time, and $\mu_{k,m}$ is the softmax-weighted average time position (Eq.~\ref{eq:mu}).
The width $\sigma_{k,m}$ is estimated via EM: iterating between soft assignment of time steps to islands and variance estimation, with $\sigma$ clamped to $[\sigma_{\min}, \sigma_{\max}]$.
Each Gaussian island $g_{k,m}(t)$ defines a soft temporal window.
The window is a temporal support gate indicating \emph{when} the concept is active. The concept's activation trajectory is carried by the encoder activation $\mathbf{A}_0$ within the window.

\subsubsection{Island selection}
\label{sec:selection}

From $K \times M$ candidate islands per sample, we select a budget of $n_{\text{global}}$ islands via differentiable top-$k$ with straight-through estimation (STE)~\citep{bengio2013estimating}.
Each island $(k, m)$ is scored by its energy-gated activation:
\begin{equation}
    s_{k,m} = \frac{\sum_t A_0[b,k,t] \cdot g_{k,m}(t)}{\sqrt{\sum_t g_{k,m}(t)}} \cdot \text{gate}_{k,m}
    \label{eq:score}
\end{equation}
where the numerator measures total activation energy within the Gaussian window, the denominator normalizes by effective window length, and $\text{gate}_{k,m} \in [0,1]$ is a learned budget-control signal.

The selected islands define a temporal mask for each latent:
\begin{equation}
    \text{mask}[b,k,t] = \text{clamp}\!\left(\sum_{m=1}^{M} \text{sel}_{k,m} \cdot g_{k,m}(t),\ 0,\ 1\right)
    \label{eq:mask}
\end{equation}
and the masked activation is $\Ahat[b,k,t] = A_0[b,k,t] \cdot \text{mask}[b,k,t]$.

\subsubsection{Decoder}
\label{sec:decoder}

The decoder reconstructs $\Hgrid$ from the masked activations. Each latent $k$ has an atom $\mathbf{p}_k \in \reals^D$ and reuses the same underlying routing parameters as the encoder, with top-$\kappa_{\text{dec}}$ sparsification ($\kappa_{\text{dec}} <
  \kappa_{\text{enc}}$):
\begin{equation}
    \Hhat[b,t,v,d] = \sum_{k=1}^{K} \Ahat[b,k,t] \cdot W[k,v] \cdot P[k,d] + b_d
    \label{eq:decoder}
\end{equation}
where $\mathbf{b} \in \reals^D$ is a shared bias.

\subsection{Stage A: dictionary learning}
\label{sec:stage_a}

\tisae{} is trained to reconstruct frozen backbone representations.
Before encoding, we subtract a population-level per-variable mean $\bar{\mathbf{H}} \in \reals^{V \times D}$ computed from training data, so the SAE operates on $\Hgrid - \bar{\mathbf{H}}$; the mean is added back after decoding.
The primary loss is per-cell normalized mean squared error (NMSE) on observed grid cells:
\begin{equation}
    \loss_{\text{recon}} = \frac{1}{|\mathcal{O}|} \sum_{(b,t,v) \in \mathcal{O}} \frac{\|\Hhat[b,t,v,:] - \Hgrid[b,t,v,:]\|_c^2}{\|\Hgrid[b,t,v,:]\|_c^2 + \epsilon}
    \label{eq:recon}
\end{equation}
where $\|\cdot\|_c$ denotes centered (mean-subtracted) norm and $\mathcal{O} = \{(b,t,v) : O[b,t,v] = 1\}$.

\paragraph{Dead latent revival.}
Latents that are not selected for $>$$\delta$ batches are declared ``dead.''
An auxiliary loss $\loss_{\text{auxk}}$ trains dead latents to reconstruct the residual $\Hgrid - \Hhat$, reviving unused capacity~\citep{gao2025scaling}.

\paragraph{Regularizers.}
$\mu$-repulsion ($\lambda_{\text{repul}}$) pushes island centers apart within each latent, preventing multiple islands converge to the same time position.
Decoder diversity ($\lambda_{\text{div}}$) penalizes cosine similarity between $\mathbf{W}$ rows, encouraging variable-composition diversity across latents.
Mask coverage ($\lambda_{\text{mcov}}$) penalizes the total Gaussian mass of selected islands, preventing \tisae{} from increasing temporal coverage to favor reconstruction.
Channel purity ($\lambda_{\text{ch\_purity}}$) encourages each latent's routing weights to concentrate on variables from a single clinical channel.
Full regularizer formulas are in Appendix~\ref{app:regularizers}.

The Stage~A total loss is:
\begin{equation}
  \loss_A = \lambda_{\text{recon}} \loss_{\text{recon}} + 
  \lambda_{\text{auxk}} \loss_{\text{auxk}} + \lambda_{\text{repul}} \loss_{\text{repul}} + \lambda_{\text{div}} \loss_{\text{div}} + 
  \lambda_{\text{mcov}} \loss_{\text{mcov}} +
  \lambda_{\text{ch}} \loss_{\text{ch}}
  \label{eq:stage_a}
\end{equation}

\subsection{Stage B: prediction-aware selector}
\label{sec:predselector}

The dictionary from Stage~A captures what the model represents; a prediction-aware concept selector (hereafter, the \predselector{}) identifies which concepts drive each prediction.
The entire \tisae{} is frozen during training.

Each candidate island $(k, m)$ is described by a feature vector $\mathbf{e}_{k,m}$ combining its temporal parameters ($\mu$, $\sigma$), activation statistics, the routing weights $\mathbf{w}_k$, and observation coverage within the island window (full specification in Appendix~\ref{app:predselector}).
A patient-level query $\mathbf{q}$ is obtained from the backbone's attention pooling.
Island features and patient query are projected into a shared space, combined via element-wise interaction, and scored:
\begin{equation}
    s^{\text{pred}}_{k,m} = \text{MLP}\!\left([\mathbf{U}_e \mathbf{e}_{k,m};\ \mathbf{U}_q \mathbf{q};\ \mathbf{U}_e \mathbf{e}_{k,m} \odot \mathbf{U}_q \mathbf{q}]\right)
    \label{eq:pred_score}
\end{equation}
where $\mathbf{U}_e$ and $\mathbf{U}_q$ are learned projection matrices.

Islands are selected via softmax-weighted scoring with temperature $\tau$, producing continuous selection weights that sum to $n_{\text{global}}^{\text{pred}}$.

\paragraph{Losses.}
Let $z_{\text{keep}}$, $z_{\text{drop}}$, and $z_{\text{full}}$ denote the prediction logit when using only selected islands, only the complement, and all islands respectively, with teacher logit $z_{\text{teacher}}$ from the frozen backbone.
\begin{align}
    \loss_{\text{logit}} &= \|z_{\text{keep}} - z_{\text{teacher}}\|^2 \label{eq:logit} \\
    \loss_{\text{comp}} &= \text{ReLU}\!\left(\delta - c \cdot (z_{\text{full}} - z_{\text{drop}})\right) \label{eq:comp_loss}
\end{align}
where $c = \text{sign}(z_{\text{teacher}})$ is the teacher's predicted class direction and $\delta$ is a hinge margin.
The total \predselector{} loss is $\loss_B = \lambda_{\text{logit}} \loss_{\text{logit}} + \lambda_{\text{comp}} \loss_{\text{comp}}$.

\subsection{Latent characterization}
\label{sec:verbalization}

The learned dictionary is characterized post-hoc.
Each latent $k$ is described by actual raw-value patterns of its variable composition ($\mathbf{w}_k$) and temporal extents (island $\mu$, $\sigma$) from top-$N$ exemplar episodes ranked by activation and the prediction-aware selection.
In this study, an LLM labeler produces per-latent clinical concept labels and descriptions from these summaries (Appendix~\ref{app:dictionary_labeling}).

\section{Experimental setup}
\label{sec:setup}

\subsection{Dataset}
\label{sec:dataset}
We use MIMIC-III~\citep{johnson2016mimic} and MIMIC-IV~\citep{johnson2023mimiciv}, publicly available critical care databases.
We extract 48-hour ICU episodes with $V{=}135$ clinical variables (MIMIC-III) and $V{=}132$ variables (MIMIC-IV).
We evaluate on three binary prediction tasks: in-hospital mortality, 30-day readmission, and length-of-stay $>$7 days.
MIMIC-III contains $N{=}28{,}626$ episodes (14.1\% mortality); MIMIC-IV contains $N{=}46{,}140$ episodes (14.4\% mortality).
Both are split into train/validation/test (64/16/20\%) by patient.

\subsection{Model training}
\label{sec:training}

We evaluate \tisae{} across five backbone configurations (Section~\ref{sec:related_ehr}): a discrete transformer (with and without pretraining), STraTS~\citep{tipirneni2022self}, HyMaTE~\citep{mottalib2025hymate}, and BAT~\citep{devries2025bat}.
\tisae{} ($K{=}1280$, $M{=}4$, $n_{\text{global}}{=}128$) and the prediction selector ($n_{\text{global}}^{\text{pred}}{=}64$) are trained on each frozen backbone.
Full architecture specifications and training details are in Appendix~\ref{app:training}.

\paragraph{Baselines.}
We compare against backbone attention, Integrated
Gradients~\citep{sundararajan2017axiomatic}, 
TimeSHAP~\citep{bento2021timeshap},
Dynamask~\citep{crabbe2021explaining},
TSR~\citep{ismail2020benchmarking}, Cluster-TCAV~\citep{ghorbani2019towards,kim2018interpretability},
FIT~\citep{tonekaboni2020went}, WinIT~\citep{leung2023temporal},
and random sanity checks, all evaluated at matched top-$f$ fractions.

\subsection{Evaluation metrics}
\label{sec:eval_metrics}

\subsubsection{Faithfulness}
\label{sec:faithfulness}

Comprehensiveness measures the prediction change when the top-$f$ most important events are removed; sufficiency measures the change when only those events are kept.
We evaluate under three perturbation strategies that define how removed events are treated: \emph{removal} (event deleted entirely), \emph{marginal} (replaced with the per-variable median from the training set), and \emph{neighborhood-based} (replaced with the patient's temporally nearest observed value of the same variable,
approximating conditional sampling $P(x_v \mid x_{\text{observed}})$).
We report neighborhood-based sampling as the primary strategy, as it best approximates realistic clinical counterfactuals.
We summarize the comprehensiveness curve with AOPC (Area Over the Perturbation Curve).

\subsubsection{Clinical plausibility}
\label{sec:plausibility}

We evaluate whether attributions align with established clinical severity indicators, using SOFA organ-system severity~\citep{vincent1996sofa} as soft ground truth generated on both MIMIC-III and MIMIC-IV datasets. We measure whether each method's top-attributed cells correspond to clinically severe observations (AUROC and AUPRC).
Ground truth generation details are in Appendix~\ref{app:plausibility}.

\subsubsection{Verbalized explanation quality}
\label{sec:layer3}

For patient-level explanations, the top \predselector{}-selected latents are combined with raw clinical values and the model's prediction, then passed to an LLM verbalizer to produce natural language explanations. 
We evaluate verbalization quality by comparing explanations under two conditions: \emph{attribution-augmented} (top-$K$ attributed cells)
and \emph{\tisae{}-augmented} (top \predselector{}-selected concepts with dictionary labels), both receiving the same patient data table with matched information budget. 
This comparison mimics a clinician reading each method's explanation, so the inputs intentionally differ in the two conditions.
A pairwise LLM judge and four tertiary-hospital clinicians independently evaluate each dimension: \emph{concept-level reasoning}, \emph{clinical actionability}, and \emph{faithfulness}.
Full protocol in Appendix~\ref{app:verbalization}.

\section{Results}
\label{sec:results}

\begin{figure}[t]
    \includegraphics[width=1.0\textwidth]{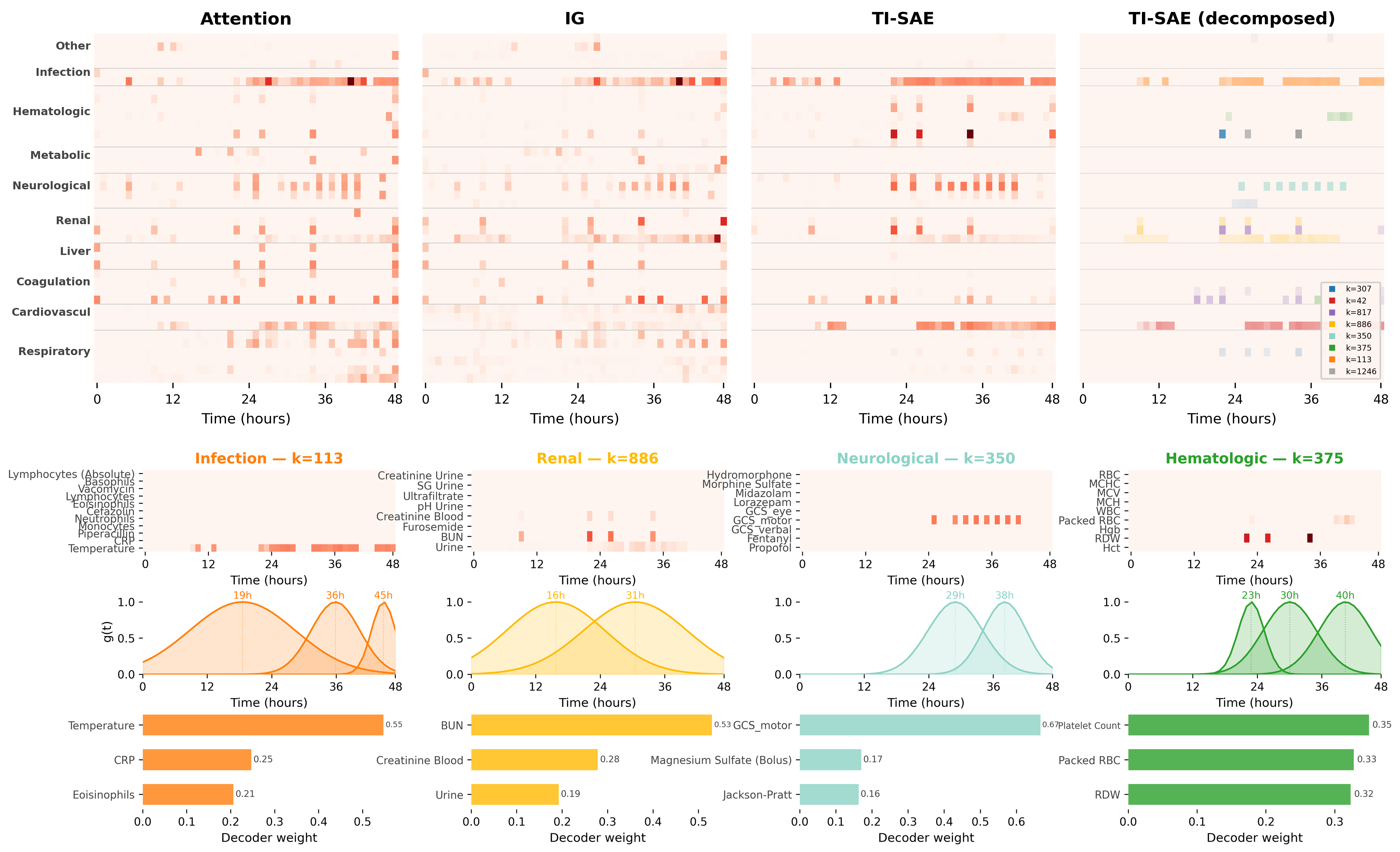}
    \centering
    \caption{
    \textbf{Qualitative comparison of attribution methods and \tisae{} decomposition.}
    (a)~Attribution heatmaps for an example mortality episode.
    \tisae{} produces structured attribution that decomposes into per-channel latent
    contributions (rightmost column: each color represents one latent).
    (b)~Four selected latents from the decomposition. Top row: attribution restricted to the latent. Middle row: learned Gaussian islands defining when the concept is active. Bottom row: decoder variable-composition weights $\mathbf{w}_k$.
    }
    \label{fig:qualitative}
\end{figure}

We evaluate \tisae{} across five backbone architectures on MIMIC-III and MIMIC-IV (three prediction tasks each).
Figure~\ref{fig:qualitative} illustrates that \tisae{}'s attribution decomposes into per-channel latent contributions with learned temporal windows and variable composition.

\begin{table}[t]
\caption{\textbf{Faithfulness and clinical plausibility across backbone architectures.}
AOPC: neighborhood-based perturbation.
AUROC/AUPRC: plausibility against SOFA-derived clinical prior.
Averaged over tasks and datasets.
Best per column in \textbf{bold}, second-best \underline{underlined}.}
\label{tab:main_results}
\centering
\small
\setlength{\tabcolsep}{2.5pt}
\begin{tabular}{l c@{\hskip 6pt}c@{\hskip 6pt}c@{\hskip 6pt}c@{\hskip 6pt}c ccccc}
\toprule
& \multicolumn{5}{c}{Faithfulness (AOPC $\uparrow$)} & \multicolumn{5}{c}{Clinical Plausibility (AUROC/AUPRC $\uparrow$)} \\
\cmidrule(lr){2-6} \cmidrule(lr){7-11}
& Disc. & D+PT & HyM. & BAT & STraTS & Discrete & Disc.+PT & HyMaTE & BAT & STraTS \\
\midrule
\tisae{} (soft) & \textbf{.086} & \textbf{.076} & \textbf{.097} & \textbf{.043} & .016 & \textbf{.593}/\textbf{.286} &
\underline{.589}/\underline{.255} & \underline{.614}/\textbf{.276} & \underline{.588}/\underline{.251} & .536/.227 \\
\tisae{} (hard) & \underline{.079} & \underline{.068} & \underline{.090} & \underline{.041} & .016 & .569/\underline{.267} & .568/.243 & .581/.257 &
.567/.243 & .535/.228 \\
Attention & .053 & .019 & .051 & -- & .016 & .570/.242 & .539/.219 & .554/.218 & -- & .551/.217 \\
IG & .060 & .051 & .049 & .031 & \underline{.067} & .545/.225 & .549/.229 & .596/.244 & .584/.239 & .577/.223 \\
TSR & .063 & .054 & .062 & .037 & \textbf{.071} & .566/.243 & .573/.241 & \textbf{.625}/.272 & \textbf{.608}/\textbf{.261} &
\underline{.595}/\underline{.234} \\
DynaMask & .019 & .032 & .012 & .028 & .034 & .510/.201 & .503/.200 & .494/.187 & .493/.187 & .494/.187 \\
TimeSHAP & .009 & .011 & .012 & .016 & .029 & .512/.202 & .517/.206 & .516/.205 & .519/.209 & .513/.201 \\
WinIT & .020 & .030 & .032 & .021 & .032 & \underline{.578}/.252 & \textbf{.611}/\textbf{.280} & .611/\underline{.273} & .565/.247 &
\textbf{.621}/\textbf{.280} \\
Random & .002 & .003 & .002 & .006 & .007 & .502/.190 & .502/.190 & .502/.190 & .502/.190 & .502/.190 \\
\bottomrule
\end{tabular}
\end{table}

\subsection{Faithfulness and clinical plausibility}
\label{sec:results_main}

Table~\ref{tab:main_results} presents faithfulness (AOPC under neighborhood-based perturbation) and clinical plausibility (AUROC and AUPRC against SOFA severity) across backbones, averaged over tasks and datasets.
\tisae{} achieves the highest AOPC on four of five backbones (Discrete, Disc.+PT, HyMaTE, BAT), exceeding baselines.
On STraTS, \tisae{}'s advantage is reduced; we find that STraTS's residual-averaging architecture does not preserve the per-variable locality that \tisae{}'s routing requires (Appendix~\ref{app:ablations}).
On clinical plausibility, \tisae{} achieves the highest AUPRC and AUROC on the discrete backbone, and competitive scores on the others.

\subsection{Dictionary validation}
\label{sec:results_dictionary}
The learned dictionary captures clinically coherent patterns: high-mortality latents correspond to vasopressor-dependent shock and respiratory failure, while protective latents capture hemodynamic stability and normal acid-base status (Appendix~\ref{app:dictionary}).
Four independent clinicians rated a sample of 30 dictionary labels on a 5-point Likert scale, yielding mean ratings of 4.07--4.40 (Appendix~\ref{app:clinician_eval}).

The learned concepts are reproducible across independent training runs, both when retraining the full pipeline (decoder cosine .724, variable overlap 77\%) and per stage (dictionary alone: decoder cosine .735, variable overlap 82\%; selector on a fixed dictionary: top-64 Jaccard .708, rank correlation .87; Appendix~\ref{app:seed_stability}), and zero-ablating importance-ranked concepts degrades predictions monotonically and more than random-order removal, confirming that individual concepts carry non-redundant predictive contributions (Appendix~\ref{app:concept_knockout}).
The learned dictionary shows cross-dataset transfer: faithfulness and plausibility preserved when the MIMIC-III--trained model is applied to disjoint MIMIC-IV without retraining (Table~\ref{tab:cross_dataset}, Appendix~\ref{app:cross_dataset}).

\begin{table}[h]
\caption{\textbf{Cross-dataset transfer}. MIMIC-III$\to$IV. MIMIC-IV was restricted to admissions from 2014 to remove patient and temporal overlap with MIMIC-III. 
For the Discrete backbone trained on MIMIC-III mortality, TI-SAE and \predselector{} are applied directly to MIMIC-IV.
Baselines recomputed on disjoint MIMIC-IV.
Neighborhood-based perturbation. 
Backbone AUROC: .861$\to$.864. ID = in-distribution; OOD = out-of-distribution.}
\label{tab:cross_dataset}
\centering
\small
\begin{tabular}{l cc cc}
\toprule
& \multicolumn{2}{c}{AOPC $\uparrow$} & \multicolumn{2}{c}{AUROC/AUPRC $\uparrow$} \\
\cmidrule(lr){2-3} \cmidrule(lr){4-5}
& ID (MIMIC-III) & OOD (MIMIC-IV) & ID (MIMIC-III) & OOD (MIMIC-IV) \\
\midrule
\tisae{} (soft) & .149 & .130 & .756/.386 & .756/.370 \\
Attention & .100 & .097 & .616/.259 & .631/.261 \\
IG & .086 & .045 & .567/.255 & .552/.224 \\
TSR & .088 & .046 & .593/.280 & .587/.245 \\
\bottomrule
\end{tabular}
\end{table}

\subsection{Verbalized explanation quality}
\label{sec:results_verbalization}

Table~\ref{tab:verbalization} reports pairwise win counts from the LLM judge ($N{=}50$ per baseline) and the clinician panel ($N{=}50$ on the IG comparison). \tisae{} wins concept-level reasoning and actionability across all baselines. 
Clinician majority votes match the LLM judge's ranking direction for concept-level reasoning and actionability, with faithfulness near-tied for both.
Per-rater details and inter-rater agreement analysis are reported in
Appendix~\ref{app:clinician_pairwise_eval}.

\begin{table}[h]
\caption{\textbf{Verbalized explanation quality.} Pairwise win counts:
TI-SAE / baseline (ties). LLM judge ($N{=}50$ per baseline) and
clinician majority vote (four independent raters) on the IG comparison.
}
\label{tab:verbalization}
\centering
\small
\setlength{\tabcolsep}{4pt}
\begin{tabular}{l cccc}
\toprule
& \multicolumn{3}{c}{LLM judge} & Clinician \\
\cmidrule(lr){2-4} \cmidrule(lr){5-5}
& vs.\ Att & vs.\ IG & vs.\ TSR & vs.\ IG \\
\midrule
Concept       & \textbf{30}--17 (3)  & \textbf{33}--15 (2)  & \textbf{33}--16 (1)  & \textbf{39}--4 (7)   \\
Actionability & \textbf{28}--22 (0)  & \textbf{32}--18 (0)  & \textbf{27}--23 (0)  & \textbf{23}--9 (18)  \\
Faithfulness  & \textbf{19}--15 (16) & \textbf{15}--11 (24) & \textbf{28}--9 (13)  & 4--4 (42)            \\
\bottomrule
\end{tabular}
\end{table}

An ablation isolating the role of the dictionary labels (Appendix~\ref{app:label_ablation}) shows that much of this verbalized advantage derives from the cross-patient concept labels: with these labels removed, \tisae{}'s concepts no longer clearly beat per-cell attribution. 
However, the concept labels are generated from cross-patient exemplar evidence, and only \tisae{}'s shared dictionary provides them with spatiotemporal grounding.

\subsection{Ablation study}
\label{sec:results_ablations}

We ablate \tisae{}'s structural mechanisms, dictionary capacity, and loss components (Appendix~\ref{app:ablations}).
Within \tisae{}, sparse variable routing is the critical mechanism for faithfulness (removing it costs 58--75\% of AOPC), while the temporal islands and channel organization have minimal impact on faithfulness but provide the concepts' clinical readability: without them, channel coherence drops from 91.9\% to 36.9\% and temporal localization collapses (Appendix~\ref{app:structural_ablation}).
Isolating the dictionary from the \predselector{} shows that the same selection objective applied to raw hidden states collapses (neighborhood-based AOPC .038; Appendix~\ref{app:structural_ablation}).
Composition over multiple windows matters, as a single island per concept ($M{=}1$) degrades both faithfulness and clinical plausibility (Appendix~\ref{app:capacity_sparsity}). The symmetric Gaussian window shape is not restrictive: a learned split-normal window changes neither metric (Appendix~\ref{app:island_shape}).

\section{Limitations}
\label{sec:limitations}

First, \tisae{} operates on a fixed-length time$\times$variable grid; variable-length sequences or observation windows beyond the 48-hour setting used here would require adaptation, which we have not validated.
Second, \tisae{}'s variable-routing encoder requires hidden representations that balance per-variable locality with contextual concept formation. Not all backbone layers achieve this.
Third, while \tisae{}'s latents are structurally defined by sparse variable composition and temporal extent, their identification as \emph{clinical concepts} relies on post-hoc validation.
Fourth, SOFA-derived severity is a coarse, organ-independent proxy that cannot capture the full complexity of clinical reasoning.
Fifth, \tisae{} assumes a single-token-per-event representation that maps to one cell in the time$\times$variable grid; multi-token strategies may require architectural adaptation.
Sixth, our verbalization evaluation is mediated by an LLM verbalizer whose strong clinical priors can produce fluent explanations from any source: high quality does not guarantee the explanation reflects model reasoning, and although concept-based inputs constrain the verbalizer more tightly, this asymmetry is not directly measured.

\section{Conclusion}
\label{sec:conclusion}

We introduced \tisae{}, a sparse autoencoder that decomposes frozen clinical time-series model representations into a dictionary of temporal clinical concepts. Each concept is defined by learned Gaussian temporal islands and sparse variable composition, enabling structured explanations that go beyond cell-level attribution scores.
Evaluated across five backbone architectures on MIMIC-III and MIMIC-IV with three different tasks, \tisae{} achieves the highest faithfulness on four of five backbones while providing a decomposable, clinically labeled dictionary that no baseline method can produce.

\begin{ack}
This work was supported by the National Institutes of Health and National Institute on Aging (project number R01AG084637 and R01AG082721); the BK21 FOUR program of the Education and Research Program for Future ICT Pioneers, Seoul National University in 2026; the National Research Foundation of Korea (NRF) grant funded by the Korea government (MSIT) (Nos. 2022R1A3B1077720 and 2022R1A5A7083908); the Institute of Information \& Communications Technology Planning \& Evaluation (IITP) grant funded by the Korea government (MSIT) (No. RS-2021-II211343, Artificial Intelligence Graduate School Program, Seoul National University); and the InnoCORE program of the Ministry of Science and ICT(1.260007.01).

The authors declare no competing interests.
\end{ack}

\bibliography{references}

@article{rajkomar2018scalable,
author = {Rajkomar, Alvin and Oren, Eyal and Chen, Kai and Dai, Andrew M. and Hajaj, Nissan and Hardt, Michaela and Liu, Peter J. and Liu, Xiaobing and Marcus, Jake and Sun, Mimi and Sundberg, Patrik and Yee, Hector and Zhang, Kun and Zhang, Yi and Flores, Gerardo and Duggan, Gavin E. and Irvine, Jamie and Le, Quoc and Litsch, Kurt and Mossin, Alexander and Tansuwan, Justin and Wang, De and Wexler, James and Wilson, Jimbo and Ludwig, Dana and Volchenboum, Samuel L. and Chou, Katherine and Pearson, Michael and Madabushi, Srinivasan and Shah, Nigam H. and Butte, Atul J. and Howell, Michael D. and Cui, Claire and Corrado, Greg S. and Dean, Jeffrey},
title = {Scalable and accurate deep learning with electronic health records},
journal = {npj Digital Medicine},
volume = {1},
number = {1},
pages = {18},
DOI = {10.1038/s41746-018-0029-1},
year = {2018},
type = {Journal Article}
}

@article{harutyunyan2019multitask,
author = {Harutyunyan, Hrayr and Khachatrian, Hrant and Kale, David C. and Ver Steeg, Greg and Galstyan, Aram},
title = {Multitask learning and benchmarking with clinical time series data},
journal = {Scientific Data},
volume = {6},
number = {1},
pages = {96},
DOI = {10.1038/s41597-019-0103-9},
year = {2019},
type = {Journal Article}
}

@article{tipirneni2022self,
author = {Tipirneni, Sindhu and Reddy, Chandan K.},
title = {Self-Supervised Transformer for Sparse and Irregularly Sampled Multivariate Clinical Time-Series},
journal = {ACM Trans. Knowl. Discov. Data},
volume = {16},
number = {6},
DOI = {10.1145/3516367},
year = {2022},
type = {Journal Article}
}

@inproceedings{tonekaboni2019clinicians,
title={What Clinicians Want: Contextualizing Explainable Machine Learning for Clinical End Use},
author={Tonekaboni, Sana and Joshi, Shalmali and McCradden, Melissa D. and Goldenberg, Anna},
booktitle={Proceedings of the 4th Machine Learning for Healthcare Conference},
pages={359--380},
year={2019},
url ={https://proceedings.mlr.press/v106/tonekaboni19a.html}
}

@article{ghassemi2021false,
author = {Ghassemi, Marzyeh and Oakden-Rayner, Luke and Beam, Andrew L.},
title = {The false hope of current approaches to explainable artificial intelligence in health care},
journal = {The Lancet Digital Health},
volume = {3},
number = {11},
pages = {e745-e750},
DOI = {10.1016/S2589-7500(21)00208-9},
year = {2021},
type = {Journal Article}
}

@article{johnson2016mimic,
title={MIMIC-III, a freely accessible critical care database},
author={Johnson, Alistair EW and Pollard, Tom J and Shen, Lu and Lehman, Li-wei H and Feng, Mengling and Ghassemi, Mohammad and Moody, Benjamin and Szolovits, Peter and Anthony Celi, Leo and Mark, Roger G},
journal={Scientific data},
volume={3},
number={1},
pages={1--9},
year={2016},
publisher={Nature Publishing Group}
}

@article{johnson2023mimiciv,
title={{MIMIC-IV}, a freely accessible electronic health record dataset},
author={Johnson, Alistair E W and Bulgarelli, Lucas and Shen, Lu and Gayles, Alvin and Shammout, Ayad and Horng, Steven and Pollard, Tom J and
Hao, Sicheng and Moody, Benjamin and Gow, Brian and Lehman, Li-wei H and Celi, Leo A and Mark, Roger G},
journal={Scientific Data},
volume={10},
number={1},
pages={1},
year={2023},
publisher={Nature Publishing Group}
}

@inproceedings{sundararajan2017axiomatic,
author = {Sundararajan, Mukund and Taly, Ankur and Yan, Qiqi},
title = {Axiomatic attribution for deep networks},
year = {2017},
booktitle = {Proceedings of the 34th International Conference on Machine Learning - Volume 70},
pages = {3319–3328},
numpages = {10},
location = {Sydney, NSW, Australia},
}

@inproceedings{lundberg2017unified,
author = {Lundberg, Scott M. and Lee, Su-In},
title = {A unified approach to interpreting model predictions},
year = {2017},
booktitle = {Proceedings of the 31st International Conference on Neural Information Processing Systems},
pages = {4768–4777},
numpages = {10},
location = {Long Beach, California, USA},
}

@inproceedings{shrikumar2017learning,
author = {Shrikumar, Avanti and Greenside, Peyton and Kundaje, Anshul},
title = {Learning important features through propagating activation differences},
year = {2017},
booktitle = {Proceedings of the 34th International Conference on Machine Learning - Volume 70},
pages = {3145–3153},
numpages = {9},
location = {Sydney, NSW, Australia},
}

@inproceedings{bento2021timeshap,
author = {Bento, Jo\~{a}o and Saleiro, Pedro and Cruz, Andr\'{e} F. and Figueiredo, M\'{a}rio A.T. and Bizarro, Pedro},
title = {TimeSHAP: Explaining Recurrent Models through Sequence Perturbations},
year = {2021},
doi = {10.1145/3447548.3467166},
booktitle = {Proceedings of the 27th ACM SIGKDD Conference on Knowledge Discovery \& Data Mining},
pages = {2565–2573},
numpages = {9},
location = {Virtual Event, Singapore},
}

@inproceedings{crabbe2021explaining,
title={Explaining Time Series Predictions with Dynamic Masks},
author={Crabb{\'e}, Jonathan and Van Der Schaar, Mihaela},
booktitle={Proceedings of the 38th International Conference on Machine Learning},
pages={2166--2177},
year={2021},
}

@inproceedings{tonekaboni2020went,
author = {Tonekaboni, Sana and Joshi, Shalmali and Campbell, Kieran R and Duvenaud, David and Goldenberg, Anna},
title = {What went wrong and when? instance-wise feature importance for time-series black-box models},
year = {2020},
booktitle = {Proceedings of the 34th International Conference on Neural Information Processing Systems},
articleno = {68},
numpages = {11},
location = {Vancouver, BC, Canada},
}

@inproceedings{leung2023temporal,
title={Temporal Dependencies in Feature Importance for Time Series Prediction},
author={Kin Kwan Leung and Clayton Rooke and Jonathan Smith and Saba Zuberi and Maksims Volkovs},
booktitle={The Eleventh International Conference on Learning Representations},
year={2023},
url={https://openreview.net/forum?id=C0q9oBc3n4}
}

@inproceedings{ismail2020benchmarking,
author = {Ismail, Aya Abdelsalam and Gunady, Mohamed and Bravo, H\'{e}ctor Corrada and Feizi, Soheil},
title = {Benchmarking deep learning interpretability in time series predictions},
year = {2020},
booktitle = {Proceedings of the 34th International Conference on Neural Information Processing Systems},
articleno = {540},
numpages = {12},
location = {Vancouver, BC, Canada},
}

@inproceedings{liu2024explaining,
title={Explaining Time Series via Contrastive and Locally Sparse Perturbations},
author={Zichuan Liu and Yingying ZHANG and Tianchun Wang and Zefan Wang and Dongsheng Luo and Mengnan Du and Min Wu and Yi Wang and Chunlin Chen and Lunting Fan and Qingsong Wen},
booktitle={The Twelfth International Conference on Learning Representations},
year={2024},
url={https://openreview.net/forum?id=qDdSRaOiyb}
}

@inproceedings{queen2023encoding,
author = {Queen, Owen and Hartvigsen, Thomas and Koker, Teddy and He, Huan and Tsiligkaridis, Theodoros and Zitnik, Marinka},
title = {Encoding time-series explanations through self-supervised model behavior consistency},
year = {2023},
booktitle = {Proceedings of the 37th International Conference on Neural Information Processing Systems},
articleno = {1395},
numpages = {31},
location = {New Orleans, LA, USA},
}

@inproceedings{liu2024timex++,
author = {Liu, Zichuan and Wang, Tianchun and Shi, Jimeng and Zheng, Xu and Chen, Zhuomin and Song, Lei and Dong, Wenqian and Obeysekera, Jayantha and Shirani, Farhad and Luo, Dongsheng},
title = {TIMEX++: learning time-series explanations with information bottleneck},
year = {2024},
booktitle = {Proceedings of the 41st International Conference on Machine Learning},
articleno = {1297},
numpages = {21},
location = {Vienna, Austria},
}

@inproceedings{chen2018learning,
title={Learning to Explain: An Information-Theoretic Perspective on Model Interpretation},
author={Chen, Jianbo and Song, Le and Wainwright, Martin and Jordan, Michael},
booktitle={Proceedings of the 35th International Conference on Machine Learning},
pages={883--892},
year={2018},
}

@article{bricken2023monosemanticity,
title={Towards Monosemanticity: Decomposing Language Models With Dictionary Learning},
author={Bricken, Trenton and Templeton, Adly and Batson, Joshua and Chen, Brian and Jermyn, Adam and Conerly, Tom and Turner, Nick and Anil, Cem and Denison, Carson and Askell, Amanda and Lasenby, Robert and Wu, Yifan and Kravec, Shauna and Schiefer, Nicholas and Maxwell, Tim and Joseph, Nicholas and Hatfield-Dodds, Zac and Tamkin, Alex and Nguyen, Karina and McLean, Brayden and Burke, Josiah E and Hume, Tristan and Carter, Shan and Henighan, Tom and Olah, Christopher},
year={2023},
journal={Transformer Circuits Thread},
url={https://transformer-circuits.pub/2023/monosemantic-features/index.html}
}

@article{templeton2024scaling,
title={Scaling Monosemanticity: Extracting Interpretable Features from Claude 3 Sonnet},
author={Templeton, Adly and Conerly, Tom and Marcus, Jonathan and Lindsey, Jack and Bricken, Trenton and Chen, Brian and Pearce, Adam and Citro, Craig and Ameisen, Emmanuel and Jones, Andy and Cunningham, Hoagy and Turner, Nicholas L and McDougall, Callum and MacDiarmid, Monte and Freeman, C. Daniel and Sumers, Theodore R. and Rees, Edward and Batson, Joshua and Jermyn, Adam and Carter, Shan and Olah, Chris and Henighan, Tom},
year={2024},
journal={Transformer Circuits Thread},
url={https://transformer-circuits.pub/2024/scaling-monosemanticity/index.html}
}

@inproceedings{
rajamanoharan2024improving,
title={Improving Sparse Decomposition of Language Model Activations with Gated Sparse Autoencoders},
author={Senthooran Rajamanoharan and Arthur Conmy and Lewis Smith and Tom Lieberum and Vikrant Varma and Janos Kramar and Rohin Shah and Neel Nanda},
booktitle={The Thirty-eighth Annual Conference on Neural Information Processing Systems},
year={2024},
url={https://openreview.net/forum?id=zLBlin2zvW}
}

@inproceedings{
gao2025scaling,
title={Scaling and evaluating sparse autoencoders},
author={Leo Gao and Tom Dupre la Tour and Henk Tillman and Gabriel Goh and Rajan Troll and Alec Radford and Ilya Sutskever and Jan Leike and Jeffrey Wu},
booktitle={The Thirteenth International Conference on Learning Representations},
year={2025},
url={https://openreview.net/forum?id=tcsZt9ZNKD}
}

@inproceedings{zhang2025large,
title={Large Multi-modal Models Can Interpret Features in Large Multi-modal Models},
author={Zhang, Kaichen and Shen, Yifei and Li, Bo and Liu, Ziwei},
booktitle={Proceedings of the IEEE/CVF International Conference on Computer Vision (ICCV)},
pages={3650--3661},
year={2025}
}

@inproceedings{
mishra2026dissecting,
title={Dissecting Chronos: Sparse Autoencoders Reveal Causal Feature Hierarchies in Time Series Foundation Models},
author={Anurag Mishra},
booktitle={1st ICLR Workshop on Time Series in the Age of Large Models},
year={2026},
url={https://openreview.net/forum?id=Q5ox302DLN}
}

@inproceedings{
bhalla2026temporal,
title={Temporal Sparse Autoencoders: Leveraging the Sequential Nature of Language for Interpretability},
author={Usha Bhalla and Alex Oesterling and Claudio Mayrink Verdun and Himabindu Lakkaraju and Flavio Calmon},
booktitle={The Fourteenth International Conference on Learning Representations},
year={2026},
url={https://openreview.net/forum?id=bojVI4l9Kn}
}

@inproceedings{kim2018interpretability,
title={Interpretability Beyond Feature Attribution: Quantitative Testing with Concept Activation Vectors ({TCAV})},
author={Kim, Been and Wattenberg, Martin and Gilmer, Justin and Cai, Carrie and Wexler, James and Viegas, Fernanda and sayres, Rory},
booktitle={Proceedings of the 35th International Conference on Machine Learning},
year={2018}
}

@inproceedings{ghorbani2019towards,
title={Towards automatic concept-based explanations},
author={Ghorbani, Amirata and Wexler, James and Zou, James and Kim, Been},
booktitle={Proceedings of the 33rd International Conference on Neural Information Processing Systems},
year={2019}
}

@inproceedings{koh2020concept,
title={Concept Bottleneck Models},
author={Koh, Pang Wei and Nguyen, Thao and Tang, Yew Siang and Mussmann, Stephen and Pierson, Emma and Kim, Been and Liang, Percy},
booktitle={Proceedings of the 37th International Conference on Machine Learning},
pages={5338--5348},
year={2020},
}

@inproceedings{
yuksekgonul2023posthoc,
title={Post-hoc Concept Bottleneck Models},
author={Mert Yuksekgonul and Maggie Wang and James Zou},
booktitle={The Eleventh International Conference on Learning Representations },
year={2023},
url={https://openreview.net/forum?id=nA5AZ8CEyow}
}

@article{li2020behrt,
title={{BEHRT}: Transformer for Electronic Health Records},
author={Li, Yikuan and Rao, Shishir and Solares, Jose Roberto Ayala and Hassaine, Abdelaali and Ramakrishnan, Rema and Canoy, Dexter and Zhu,
Yajie and Rahimi, Kazem and Salimi-Khorshidi, Gholamreza},
journal={Scientific Reports},
volume={10},
number={1},
pages={7155},
year={2020},
publisher={Nature Publishing Group}
}

@inproceedings{
steinberg2024motor,
title={{MOTOR}: A Time-to-Event Foundation Model For Structured Medical Records},
author={Ethan Steinberg and Jason Alan Fries and Yizhe Xu and Nigam Shah},
booktitle={The Twelfth International Conference on Learning Representations},
year={2024},
url={https://openreview.net/forum?id=NialiwI2V6}
}

@article{che2018grud,
title={Recurrent Neural Networks for Multivariate Time Series with Missing Values},
author={Che, Zhengping and Purushotham, Sanjay and Cho, Kyunghyun and Sontag, David and Liu, Yan},
journal={Scientific Reports},
volume={8},
number={1},
pages={6085},
year={2018},
publisher={Nature Publishing Group}
}

@inproceedings{mottalib2025hymate,
title={{HyMaTE}: A Hybrid Mamba and Transformer Model for {EHR} Representation Learning},
author={Mottalib, Md Mozaharul and Phan, Thao-Ly T. and Beheshti, Rahmatollah},
booktitle={Proceedings of the 16th ACM International Conference on Bioinformatics, Computational Biology, and Health Informatics},
pages={1--9},
year={2025}
}

@article{renc2024ethos,
title={Zero Shot Health Trajectory Prediction Using Transformer},
author={Renc, Pawel and Jia, Yugang and Samir, Anthony E. and Was, Jaroslaw and Li, Quanzheng and Bates, David W. and Sitek, Arkadiusz},
journal={npj Digital Medicine},
volume={7},
number={1},
pages={256},
year={2024},
publisher={Nature Publishing Group}
}

@inproceedings{fallahpour2024ehrmamba,
title={{EHRMamba}: Towards Generalizable and Scalable Foundation Models for Electronic Health Records},
author={Fallahpour, Adibvafa and Alinoori, Mahshid and Ye, Wenqian and Cao, Xu and Afkanpour, Arash and Krishnan, Amrit},
booktitle={Proceedings of the 4th Machine Learning for Health Symposium},
volume={259},
pages={291--307},
year={2025}
}

@misc{devries2025bat,
title={Bi-Axial Transformers: Addressing the Increasing Complexity of {EHR} Classification},
author={Rachael DeVries and Casper Christensen and Marie Lisandra Zepeda Mendoza and Ole Winther},
year={2025},
eprint={2508.12418},
archivePrefix={arXiv},
primaryClass={cs.LG},
url={https://arxiv.org/abs/2508.12418}, 
}

@inproceedings{labach2023duett,
title={{DuETT}: Dual Event Time Transformer for Electronic Health Records},
author={Labach, Alex and Pokhrel, Aslesha and Huang, Xiao Shi and Zuberi, Saba and Yi, Seung Eun and Volkovs, Maksims and Poutanen, Tomi and Krishnan, Rahul G.},
booktitle={Proceedings of the 8th Machine Learning for Healthcare Conference},
volume={219},
year={2023}
}

@article{vincent1996sofa,
title={The {SOFA} (Sepsis-related Organ Failure Assessment) score to describe organ dysfunction/failure},
author={Jean Louis Vincent and Rui Paolo Jin{\'o} Moreno and Jukka Takala and Sheila M. Willatts and Arnaldo de Mendonça and Hajo A. Bruining and Camelia Reinhart and Peter M. Suter and L. G. Thijs},
journal={Intensive Care Medicine},
volume={22},
number={7},
pages={707--710},
year={1996}
}

@article{zheng2023judging,
title={Judging LLM-as-a-judge with MT-bench and Chatbot Arena},
author={Zheng, Lianmin and Chiang, Wei-Lin and Sheng, Ying and Zhuang, Siyuan and Wu, Zhanghao and Zhuang, Yonghao and Lin, Zi and Li, Zhuohan and Li, Dacheng and Xing, Eric P. and Zhang, Hao and Gonzalez, Joseph E. and Stoica, Ion},
journal={Proceedings of the 37th International Conference on Neural Information Processing Systems},
year={2023}
}

@misc{bengio2013estimating,
title={Estimating or Propagating Gradients Through Stochastic Neurons for Conditional Computation}, 
author={Yoshua Bengio and Nicholas L{\'e}onard and Aaron Courville},
year={2013},
eprint={1308.3432},
archivePrefix={arXiv},
primaryClass={cs.LG},
url={https://arxiv.org/abs/1308.3432}, 
}

@misc{kokhlikyan2020captum,
title={Captum: A unified and generic model interpretability library for PyTorch}, 
author={Narine Kokhlikyan and Vivek Miglani and Miguel Martin and Edward Wang and Bilal Alsallakh and Jonathan Reynolds and Alexander Melnikov and Natalia Kliushkina and Carlos Araya and Siqi Yan and Orion Reblitz-Richardson},
year={2020},
eprint={2009.07896},
archivePrefix={arXiv},
primaryClass={cs.LG},
url={https://arxiv.org/abs/2009.07896}, 
}

@inproceedings{saadfalcon2025lmunit,
title = "{LMUNIT}: Fine-grained Evaluation with Natural Language Unit Tests",
author = "Saad-Falcon, Jon  and Vivek, Rajan Pathe  and Berrios, William  and Naik, Nandita Shankar  and Franklin, Matija  and Vidgen, Bertie  and
Singh, Amanpreet  and Kiela, Douwe  and Mehri, Shikib",
booktitle = "Findings of the Association for Computational Linguistics: EMNLP 2025",
year = "2025",
doi = "10.18653/v1/2025.findings-emnlp.176",
pages = "3303--3324",
}

@article{steinberg2021language,
title = {Language models are an effective representation learning technique for electronic health record data},
author = {Ethan Steinberg and Ken Jung and Jason A. Fries and Conor K. Corbin and Stephen R. Pfohl and Nigam H. Shah},
journal = {Journal of Biomedical Informatics},
volume = {113},
pages = {103637},
year = {2021},
doi = {https://doi.org/10.1016/j.jbi.2020.103637},
}

@InProceedings{enguehard2023learning,
title = 	 {Learning Perturbations to Explain Time Series Predictions},
author =       {Enguehard, Joseph},
booktitle = 	 {Proceedings of the 40th International Conference on Machine Learning},
pages = 	 {9329--9342},
year = 	 {2023},
volume = 	 {202},
month = 	 {23--29 Jul},
}
\bibliographystyle{plainnat}

\appendix

\section{Training and hyperparameter details}
\label{app:training}

All models are trained on NVIDIA A100-SXM4-80GB GPUs.
Deterministic seeding (seed 42) is used throughout.

\paragraph{Backbones.}
Table~\ref{tab:backbone_configs} summarizes the five backbone configurations. All backbones are trained for binary prediction via cross-entropy with inverse-prevalence class weighting.
The discrete transformer and its pretrained variant share the same architecture ($D{=}128$, 4 layers, 8 heads); the pretrained variant additionally undergoes 2-hour value-bin forecast pretraining before finetuning.
STraTS and HyMaTE use continuous value embedding (CVE) with $D{=}64$; STraTS uses 4 transformer layers with 16 heads, while HyMaTE uses 2 Mamba layers iterated over 3 blocks.
BAT uses 1 bi-axial layer with embed\_out${=}44$ (total $D{=}88$ after track concatenation), 1 attention head, and input masking for unobserved cells.
All backbones are frozen for subsequent stages.

\begin{table}[h]
\caption{Backbone configurations.}
\label{tab:backbone_configs}
\centering
\small
\begin{tabular}{lcccccl}
\toprule
Backbone & $D$ & Layers & Heads & Dropout & LR & Tokenization \\
\midrule
Discrete      & 128 & 4 & 8  & 0.3 & $10^{-4}$ & Binned (single-token) \\
Discrete+PT   & 128 & 4 & 8  & 0.3 & $10^{-4}$ & Binned (single-token) \\
STraTS        & 64  & 4 & 16 & 0.2 & $5{\times}10^{-5}$ & CVE (single-token) \\
HyMaTE        & 64  & 2${\times}$3 & 4 & 0.2 & $5{\times}10^{-5}$ & CVE (single-token) \\
BAT           & 88  & 1 & 1  & 0.2 & $5{\times}10^{-4}$ & Grid-native \\
\bottomrule
\end{tabular}
\end{table}

\paragraph{TI-SAE (stage A).}
Table~\ref{tab:tisae_hparams} lists the full \tisae{} hyperparameter configuration, applied identically across all backbones.
The model has ${\approx}856$K trainable parameters (encoder: 502K, IslandParamNet: 27K, decoder: 328K).
The variable routing matrix $\mathbf{W}$ is shared between encoder and decoder, with independent sparsification ($\kappa_{\text{enc}}{=}5$ and $\kappa_{\text{dec}}{=}3$), along with row-normalization and non-negativity.
Before encoding, a per-variable population mean $\bar{\mathbf{H}} \in \reals^{V \times D}$ is subtracted from $\Hgrid$ (computed from 100 training batches) and added back after decoding.

Training uses AdamW (lr $10^{-4}$, weight decay $10^{-2}$), cosine annealing, batch size 32, gradient norm clipping at 1.0, and early stopping on total validation loss (patience 15).
No prediction loss is used in Stage~A ($\lambda_{\text{pred}}{=}0$); prediction relevance is handled by the \predselector{} in Stage~B.

\begin{table}[t]
\caption{TI-SAE hyperparameters.}
\label{tab:tisae_hparams}
\centering
\small
\begin{tabular}{llr}
\toprule
Category & Parameter & Value \\
\midrule
\multirow{4}{*}{Architecture}
  & Latents $K$ & 1280 \\
  & Islands per latent $M$ & 4 \\
  & Selection budget $n_{\text{global}}$ & 128 \\
\midrule
\multirow{2}{*}{Encoder / Decoder}
  & Encoder Variable Routing sparsity $\kappa_{\text{enc}}$ & 5 \\
  & Decoder Variable Routing sparsity $\kappa_{\text{dec}}$ & 3 \\
  & Atom normalization & L2-normalize \\
\midrule
\multirow{5}{*}{IslandParamNet}
  & Hidden dimension $H$ & 32 \\
  & Conv kernel size & 25 \\
  & EM iterations $n_{\text{iter}}$ & 3 \\
  & $\sigma$ bounds $[\sigma_{\min}, \sigma_{\max}]$ & $[0.5,\ 12.0]$ \\
  & Score mode & energy-gated \\
\midrule
\multirow{2}{*}{Selection}
  & Gumbel noise scale & 0.05 \\
  & Dead boost scale & 0.5 \\
  & Dead threshold (batches) & 5 \\
\midrule
\multirow{6}{*}{Loss weights}
  & $\lambda_{\text{recon}}$ & 1.0 \\
  & $\lambda_{\text{auxk}}$ (dead latent revival, $k_{\text{aux}}{=}256$) & 0.3 \\
  & $\lambda_{\text{repul}}$ (island center repulsion, $\delta{=}2.0$) & 0.01 \\
  & $\lambda_{\text{mcov}}$ (mask coverage) & 0.005 \\
  & $\lambda_{\text{div}}$ (decoder diversity) & 0.01 \\
  & $\lambda_{\text{ch\_purity}}$ (channel purity) & 0.1 \\
\midrule
\multirow{5}{*}{Optimization}
  & Optimizer & AdamW \\
  & Learning rate & $10^{-4}$ \\
  & Weight decay & $10^{-2}$ \\
  & Batch size & 32 \\
  & LR schedule & Cosine annealing \\
  & Max gradient norm & 1.0 \\
  & Max epochs / patience & 100 / 15 \\
\bottomrule
\end{tabular}
\end{table}

\paragraph{Prediction selector (stage B).}
The \predselector{} is trained on the frozen \tisae{} dictionary with the backbone also frozen (${\approx}38$K trainable parameters).
Each island is represented by a $6{+}2V$-dimensional feature vector (Appendix~\ref{app:predselector}).
Candidates are pre-filtered to the top-2560 islands by energy score, from which the \predselector{} selects $n_{\text{global}}^{\text{pred}}{=}64$ islands per sample via soft (softmax-weighted) selection with temperature $\tau{=}1.0$.
The reference logit $z_{\text{full}}$ uses a separate pool of 256 top-energy islands, selected as the pool size that minimized $|z_{\text{full}} - z_{\text{teacher}}|^2$ on the validation set.

Training uses AdamW (lr $10^{-3}$, no weight decay), batch size 52, gradient norm clipping at 1.0, and early stopping on validation $\loss_B$ (patience 20, up to 100 epochs).
Loss weights are $\lambda_{\text{logit}}{=}1.0$, $\lambda_{\text{comp}}{=}0.1$, and $\delta_{\text{comp}}{=}0.5$.
Projection dimension $d_{\text{sel}}{=}64$, MLP hidden dimension 64.

\paragraph{Pretraining details.}
STraTS, HyMaTE, and Discrete+PT use a 48-hour sliding window with 2-hour value forecast pretraining objective before finetuning.
Pretraining uses a higher learning rate ($5{\times}10^{-4}$) than finetuning ($5{\times}10^{-5}$); architecture hyperparameters
(dimensions, layers, dropout) are shared.
The pretrained checkpoint is loaded and all layers are finetuned end-to-end on the downstream task.

\paragraph{Implementation notes.}
Our STraTS uses 4 transformer layers (vs.\ 2 in the original) to provide comparable model depth across backbones.
For HyMaTE, we correct a tensor dimension mismatch in the original supervised forward path.
BAT uses input masking (setting unobserved grid cell embeddings to zero before attention) to produce representations compatible with \tisae{}'s grid-based encoding.

\section{Baseline method adaptation}
\label{app:baseline_adaptation}

Our backbone consumes irregular event sequences $(v_j, t_j, x_j)_{j=1}^L$, not a regular $[T, V]$ grid.
Some baseline attribution methods require adaptation to this input format.

\paragraph{Gradient-based methods (IG, GradSHAP, TSR).}
The backbone embeds discrete token indices into continuous vectors
$\mathbf{e}_j = \mathbf{e}_{\text{val}}(x_j) + \mathbf{e}_{\text{var}}(v_j) + \mathbf{e}_{\text{time}}(t_j)$,
so we attribute in the continuous embedding space: \texttt{captum}~\citep{kokhlikyan2020captum}
treats the embedding tensor as the input and everything downstream (transformer layers, pooling, prediction head) as the forward function.
This yields per-event, per-dimension attributions $a_{j,d}$.
We first sum over embedding dimensions to obtain a signed group attribution
$\bar{a}_j = \sum_d a_{j,d}$ for each event. For ranking-based faithfulness evaluation,
we use its magnitude $a_j = |\bar{a}_j|$ as the event-importance score, and place each
$a_j$ at its grid position $(t_j, v_j)$ to obtain a $[T, V]$ attribution map.

The choice of reference embedding (baseline) determines what the attribution measures. Ideally, a null baseline representing ``no event'' would be used, but unobserved cells have no corresponding token in the sequence.
A zero embedding is the closest approximation, attributing importance to the observation's existence.
A per-variable median baseline $\mathbf{e}_j^{\text{ref}} = \mathbf{e}_{\text{val}}(b_{\text{med}}(v_j)) + \mathbf{e}_{\text{var}}(v_j) +
\mathbf{e}_{\text{time}}(t_j)$
instead attributes importance to the event's \emph{value} deviating from a typical value for that variable.
We evaluated both and selected the median baseline, for yielding better faithfulness for each method.

\paragraph{Dynamask.}
Dynamask optimizes a per-sample soft mask $\mathbf{m} \in [0,1]^{T \times V}$ to identify features whose removal most changes the prediction.
Since the backbone takes discrete token indices, the optimization requires a differentiable approximation.
For removal masking, we multiply event embeddings by the soft mask value: $\tilde{\mathbf{e}}_j = m_{t_j, v_j} \cdot \mathbf{e}_j$.
For marginal and neighborhood masking, we pre-compute a replacement embedding per event (by re-embedding a sampled or nearest-neighbor value) and
interpolate: $\tilde{\mathbf{e}}_j = m_{t_j, v_j} \cdot \mathbf{e}_j + (1 - m_{t_j, v_j}) \cdot \mathbf{e}_j^{\text{ref}}$.
At intermediate mask values these blended embeddings have no token-level equivalent, but the final attribution is binary-thresholded during
faithfulness evaluation, where embedding interpolation and token-level replacement coincide.
We fit masks at four sparsity levels (10\%, 20\%, 30\%, 50\%) and select the one with lowest prediction error.

\paragraph{TimeSHAP.}
TimeSHAP~\citep{bento2021timeshap} computes cell-level Shapley values via hierarchical KernelSHAP: event-level pruning identifies the relevant time
window, then event-level and feature-level SHAP values are computed, and finally cell-level values are estimated for the top events $\times$
features.
The package requires a callable from $[T, V]$ observation grids to predictions.
Each coalition determines which observed cells are included; excluded cells are handled according to the masking strategy (removed, marginally
replaced, or neighborhood-replaced).

\paragraph{Cluster-TCAV.}
We construct a TCAV~\citep{kim2018interpretability}-inspired prototype baseline using ACE-style unsupervised concept discovery~\citep{ghorbani2019towards}.
Contextual event embeddings $\mathbf{h}_j \in \mathbb{R}^D$ from all training samples are clustered into $K{=}50$ concepts via Mini-Batch K-means.
For each concept, we use the normalized cluster center as a CAV proxy and compute the average directional derivative of the prediction logit along this direction over training samples;
concepts whose importance is not statistically significant (one-sample $t$-test against 20 random directions, $p \geq 0.05$) are zeroed out.
At inference, each event is assigned to its nearest concept and receives that concept's importance as its attribution score, placed at its grid position $(t_j, v_j)$.

\paragraph{FIT and WinIT.}
FIT~\citep{tonekaboni2020went} and WinIT~\citep{leung2023temporal} require a conditional generator that models $P(\mathbf{x}_t \mid \mathbf{x}_{<t})$.
The standard implementations assume dense, regularly-sampled time series data.
We evaluated several adaptations: mask-aware generator training, mask-concatenated input, and event-removal counterfactual without generator training. We report results from the event-removal variant, which performed best; instead of replacing other features' values with generator samples, it removes their events by zeroing the observation mask.
For FIT, the counterfactual at time $t$ retains only feature $i$'s event; for WinIT, feature $f$'s events are removed in windows $[t,t+n]$ with finite-difference decomposition.
Both methods' attributions are computed only at observed $(t, v)$ cells.

\paragraph{TimeX, TimeX++, and ExtremalMask.}
TimeX~\citep{queen2023encoding} trains an interpretable surrogate that produces discrete attribution maps while preserving the behavior of a pretrained time-series model, and TimeX++~\citep{liu2024timex++} reformulates time-series explanation through an information-bottleneck objective.
ExtremalMask~\citep{enguehard2023learning} learns a per-sample mask and perturbation for time-series attribution.
All three methods were designed for dense, regularly-sampled multivariate time series.
Their default architectures fail on sparse clinical data (most grid cells unobserved); we adapted TimeX/TimeX++ by replacing their internal encoder with the frozen backbone (using only the \texttt{MaskGenerator} component conditioned on backbone-derived grid features), and ExtremalMask by wrapping the backbone in a grid-to-embedding adapter with tuned regularization ($\lambda_1{=}0.01$, lr${=}0.01$).
Despite these adaptations, all three produced near-random faithfulness, suggesting that the adaptations did not yield reliable implementations for sparse event-sequence data. We therefore exclude them from the main comparison and report the adaptation details here for completeness.

\paragraph{Observation masking.}
All methods' final attribution maps are masked to observed cells.
Without this, methods that produce full-grid attributions (Dynamask, TimeSHAP) waste top-$f$ budget on unobserved cells that have no causal
effect when perturbed.

\paragraph{CVE backbone adaptation.}
For CVE backbones (STraTS, HyMaTE), attributions can additionally be computed at the value level by treating the raw scalar input to the CVE MLP as the attributed input, yielding per-event value-sensitivity scores. We report the variant (embedding-level or value-level) with
higher faithfulness per method.

\section{Regularizer formulas}
\label{app:regularizers}

\paragraph{Island center repulsion.}
Prevents islands within each latent from collapsing to the same center:
\begin{equation}
    \loss_{\text{repul}} = \left\langle \exp\!\left(-\frac{(\mu_{k,m} - \mu_{k,m'})^2}{2\delta^2}\right) \cdot \mathbf{1}[m \neq m'] \right\rangle_{b,k,m,m'}
\end{equation}
averaged over all latents, island pairs, and samples, with $\delta{=}2.0$.

\paragraph{Mask coverage.}
Penalizes total temporal mass of selected islands:
\begin{equation}
    \loss_{\text{mcov}} = \frac{1}{n_{\text{global}}} \sum_{k,m} \text{sel}_{k,m} \cdot \overline{g_{k,m}}
\end{equation}
where $\overline{g_{k,m}} = \frac{1}{T}\sum_t g_{k,m}(t)$ is the temporal average of the Gaussian envelope, averaged over samples.

\paragraph{Decoder diversity.}
Penalizes cosine similarity between decoder weight rows:
\begin{equation}
    \loss_{\text{div}} = \left\langle \left(\frac{\mathbf{W}_k \cdot \mathbf{W}_{k'}}{\|\mathbf{W}_k\| \|\mathbf{W}_{k'}\|} - \mathbf{1}[k{=}k']\right)^2 \right\rangle_{k,k'}
\end{equation}
When channel organization is active, diversity is computed per-channel to avoid penalizing cross-channel differences.

\paragraph{Channel purity.}
Encourages each latent to concentrate its variable weight mass on a single clinical channel:
\begin{equation}
    \loss_{\text{ch\_purity}} = 1 - \frac{1}{K} \sum_k \frac{\max_j \sum_v W[k,v] \cdot M_{\text{ch}}[j,v]}{\sum_v W[k,v]}
\end{equation}

\section{Architecture details of IslandParamNet}
\label{app:island_param_net}

\paragraph{Temporal feature extraction.}
The IslandParamNet transforms per-latent activation traces $A_0[b,k,:] \in \reals^T$ into island parameters.
A two-layer 1D convolutional network extracts temporal features:
\begin{equation}
    \mathbf{h}_k = \text{GELU}(\text{Conv1D}_2(\text{GELU}(\text{Conv1D}_1(A_0[b,k,:])))) \in \reals^{H \times T}
\end{equation}
where both layers have hidden dimension $H{=}32$ and kernel size $25$ (with zero-padding to preserve length).

\paragraph{Island center localization.}
For each island $m$, a pointwise projection $\boldsymbol{\ell}_\mu^{(m)} \in \reals^H$ produces attention logits over time.
The center $\mu_{k,m}$ is the softmax-weighted average time position (Eq.~\ref{eq:mu} in the main text).

\paragraph{EM sigma estimation.}
Island widths $\sigma_{k,m}$ are estimated via expectation-maximization, treating time steps as data points and islands as mixture components:

\begin{algorithmic}
\STATE Initialize $\sigma_{k,m}^{(0)} = \min(T / 2M,\ 3.0)$
\FOR{$\text{iter} = 1, \dots, n_{\text{iter}}$ (default 3)}
    \STATE Gaussian kernel: $g^{\text{em}}_{m,t} = \exp\!\left(-\frac{(t - \mu_{k,m})^2}{2(\sigma_{k,m}^{(\text{iter}-1)})^2}\right)$
    \STATE Responsibilities: $r_{m,t} = g^{\text{em}}_{m,t} \big/ \sum_{m'} g^{\text{em}}_{m',t}$ \hfill (soft assignment)
    \STATE Weights: $w_{m,t} = A_0[b,k,t] \cdot r_{m,t}$ \hfill (activation-weighted)
    \STATE Update: $\sigma_{k,m}^{(\text{iter})} = \text{clamp}\!\left(\sqrt{\frac{\sum_t w_{m,t}(t - \mu_{k,m})^2}{\sum_t w_{m,t}}},\ \sigma_{\min},\ \sigma_{\max}\right)$
\ENDFOR
\end{algorithmic}

The EM procedure has no learnable parameters; $\sigma$ emerges from the shape of $A_0$.
We use $\sigma_{\min}{=}0.5$, $\sigma_{\max}{=}12.0$, and $n_{\text{iter}}{=}3$.

\paragraph{Energy-gated scoring.}
Each island $(k,m)$ is scored by its activation energy within the Gaussian window, normalized by effective support length:
\begin{equation}
    s_{k,m} = \frac{\sum_t A_0[b,k,t] \cdot g_{k,m}(t)}{\sqrt{\sum_t g_{k,m}(t)}} \cdot \text{gate}_{k,m}
\end{equation}
The $\sqrt{\cdot}$ normalization prevents wider islands from scoring higher purely due to coverage.
The gate $\text{gate}_{k,m} \in [0,1]$ is produced by a separate linear layer on \emph{detached} temporal features (gradient from $\loss_{\text{mcov}}$ reaches only the gate weights, not $\mu$ or $\sigma$), providing learned budget control.

\paragraph{Dead latent revival (AuxK).}
Latents not selected for $>\delta$ consecutive batches are declared dead ($\delta$ chosen so that approximately 150 training episodes elapse before revival).
The top $k_{\text{aux}}{=}256$ dead latents (by energy) are trained to reconstruct the residual $\Hgrid - \Hhat$ via an auxiliary loss:
\begin{equation}
    \loss_{\text{auxk}} = \frac{\|(\hat{\mathbf{H}}_{\text{aux}} - (\Hgrid - \Hhat))\ \odot\ \mathbf{O}\|^2}{\|(\Hgrid - \Hhat)\ \odot\ \mathbf{O}\|^2}
\end{equation}
where $\hat{\mathbf{H}}_{\text{aux}}$ is the dead-latent reconstruction and $\mathbf{O}$ is the observation mask.
Latents with low gate values ($< 0.1$) are excluded from the dead set, as they are budget-paused rather than truly dead.

\section{Feature vector in the prediction selector}
\label{app:predselector}

For each candidate island $(k, m)$, we assemble a feature vector $\mathbf{e}_{k,m} \in \reals^{6+2V}$:
\begin{equation}
    \mathbf{e}_{k,m} = \left[\frac{\mu_{k,m}}{T},\ \frac{\sigma_{k,m}}{T},\ \widetilde{E}_{k,m},\ \widetilde{P}_{k,m},\ \text{gate}_{k,m},\ \widetilde{G}_{k,m},\ \mathbf{W}[k,:],\ \mathbf{o}_{k,m}\right]
    \label{eq:island_features}
\end{equation}

\begin{table}[t]
\caption{Components of the island feature vector $\mathbf{e}_{k,m} \in \reals^{6+2V}$.}
\label{tab:island_features}
\centering
\small
\begin{tabular}{lccl}
\toprule
Component & Dim & Range & Description \\
\midrule
$\mu_{k,m}/T$ & 1 & $[0, 1]$ & Normalized island center position \\
$\sigma_{k,m}/T$ & 1 & $[0, 0.25]$ & Normalized island width \\
$\widetilde{E}_{k,m}$ & 1 & $[0, 1]$ & Gaussian-weighted activation energy (batch-normalized) \\
$\widetilde{P}_{k,m}$ & 1 & $[0, 1]$ & Peak activation at $\mu_{k,m}$ (batch-normalized) \\
$\text{gate}_{k,m}$ & 1 & $[0, 1]$ & Budget gate value from IslandParamNet \\
$\widetilde{G}_{k,m}$ & 1 & $[0, 1]$ & Gradient$\times$activation prior score (batch-normalized) \\
$\mathbf{W}[k,:]$ & $V$ & $[0, 1]$ & Variable-routing row (top-$\kappa_{\text{dec}}$ sparse, row-normalized) \\
$\mathbf{o}_{k,m}$ & $V$ & $[0, 1]$ & Observation profile within island window \\
\bottomrule
\end{tabular}
\end{table}

The observation profile $\mathbf{o}_{k,m} \in \reals^V$ measures what fraction of each variable's observations fall within the island's temporal window:
\begin{equation}
    o_{k,m}[v] = \frac{\sum_t g_{k,m}(t) \cdot O[b,t,v]}{\sum_t g_{k,m}(t)}
\end{equation}
This allows the selector to distinguish islands with dense observations from those covering sparsely observed regions.

\section{Clinical ontology mapping}
\label{app:ontology}

Table~\ref{tab:ontology} lists the 10 clinical channels and their constituent variables for MIMIC-III ($V{=}135$).
MIMIC-IV uses $V{=}132$ (three CareVue-era variables absent: Troponin I, Ultrafiltrate, Unknown); channel assignments are otherwise identical.

Of the $V{=}135$ variables, 24 are assigned to two channels (e.g., blood gases $\to$ Respiratory + Metabolic; IV fluids $\to$ Cardiovascular + Renal).
The hard mask $\mathbf{M}_{\text{ch}} \in \{0,1\}^{C \times V}$ used for channel purity and decoder masking reflects these shared assignments.

\begin{table}[t]
\caption{Clinical ontology mapping: 10 channels with all constituent variables ($V{=}135$). Counts include shared variables. Variables marked with $^\ast$ are assigned to two channels. Variables marked with $^\dagger$ are absent in MIMIC-IV.} 
\label{tab:ontology}
\centering
\scriptsize
\setlength{\tabcolsep}{3pt}
\begin{tabular}{clp{10.5cm}}
\toprule
Ch & Name & Variables \\
\midrule
0 & Respiratory (12) &
O$_2$ Saturation, RR, FiO$_2$, Intubated, Chest Tube,
PO$_2^\ast$, PCO$_2^\ast$, pH Blood$^\ast$, Total CO$_2^\ast$, Base Excess$^\ast$,
Propofol$^\ast$, Fentanyl$^\ast$ \\
\midrule
1 & Cardiovascular (27) &
MBP, HR, SBP, DBP, Vasopressin, Neosynephrine, Nitroglycerine, Amiodarone, Norepinephrine, Epinephrine, Milrinone, Nitroprusside, Troponin T$^\dagger$, Troponin I, NTproBNP, Hydralazine, Metoprolol, CRR,
Lactate$^\ast$, Dopamine$^\ast$, Furosemide$^\ast$, Heparin$^\ast$, Normal Saline$^\ast$, Lactated Ringers$^\ast$, Colloid$^\ast$, OR/PACU Crystalloid$^\ast$, Weight$^\ast$ \\
\midrule
2 & Coagulation (7) &
PTT, Fibrinogen,
Heparin$^\ast$, Platelet Count$^\ast$, Fresh Frozen Plasma$^\ast$, INR$^\ast$, PT$^\ast$ \\
\midrule
3 & Liver (13) &
Bilirubin (Total), Bilirubin (Direct), Bilirubin (Indirect), ALP, ALT, AST, Albumin, Ammonia,
LDH$^\ast$, Albumin 25\%$^\ast$, Albumin 5\%$^\ast$, INR$^\ast$, PT$^\ast$ \\
\midrule
4 & Renal (16) &
Creatinine Blood, Creatinine Urine, BUN, Urine, SG Urine, pH Urine, Ultrafiltrate$^\dagger$,
Furosemide$^\ast$, Dopamine$^\ast$, Normal Saline$^\ast$, Lactated Ringers$^\ast$, Colloid$^\ast$, OR/PACU Crystalloid$^\ast$, Weight$^\ast$, Albumin 25\%$^\ast$, Albumin 5\%$^\ast$ \\
\midrule
5 & Neurological (9) &
GCS\_eye, GCS\_verbal, GCS\_motor, Midazolam, Lorazepam, Hydromorphone, Morphine Sulfate,
Propofol$^\ast$, Fentanyl$^\ast$ \\
\midrule
6 & Metabolic (27) &
Anion Gap, Glucose (Serum), Glucose (Whole Blood), Glucose (Blood), Sodium, Calcium Total, Chloride, Potassium, Bicarbonate, Phosphate, Magnesium, Calcium Free, Calcium Gluconate, Magnesium Sulphate, Magnesium Sulfate (Bolus), KCl, KCl (Bolus), Insulin Regular, Insulin Humalog, Insulin Glargine, Insulin NPH,
Lactate$^\ast$, PO$_2^\ast$, PCO$_2^\ast$, pH Blood$^\ast$, Total CO$_2^\ast$, Base Excess$^\ast$ \\
\midrule
7 & Hematologic (12) &
Hgb, Hct, RBC, MCV, MCH, MCHC, RDW, Packed RBC,
WBC$^\ast$, Platelet Count$^\ast$, Fresh Frozen Plasma$^\ast$, LDH$^\ast$ \\
\midrule
8 & Infection (12) &
Temperature, Monocytes, Neutrophils, Basophils, Lymphocytes, Lymphocytes (Absolute), Eosinophils, Cefazolin, Vancomycin, Piperacillin, CRP,
WBC$^\ast$ \\
\midrule
9 & Other (24) &
Unknown$^\dagger$, Solution, D5W, Dextrose Other, PO Intake, Gastric, Gastric Meds, GT Flush, Sterile Water, Stool, Jackson-Pratt, Fiber, TPN, Half Normal Saline, Pantoprazole, Free Water, Pre-admission Intake, Piggyback, Famotidine, Pre-admission Output, Height, Emesis, Residual, EBL \\
\bottomrule
\end{tabular}
\end{table}

\section{Dictionary quality}
\label{app:dictionary}

\subsection{Reconstruction and island stability}

\tisae{} achieves cosine similarity $\cos_H$ of $0.91$ (Discrete), $0.83$ (Disc.+PT), $0.88$ (HyMaTE), $0.87$ (BAT), and $0.87$ (STraTS) between reconstructed and original backbone representations.
Island identity stability under 5\% event dropout yields Jaccard similarity 0.837--0.903 across backbones (vs.\ random baseline ${\sim}0.01$).

\subsection{Latent utilization}
\label{app:dead_latents}

Table~\ref{tab:dead_latents} reports latent utilization for the mortality prediction task on both cohorts. A dead latent here is defined as a latent that never selected across the full evaluation cohort.

\begin{table}[h]
\centering
\caption{Latent utilization across backbones ($K{=}1280$, mortality task).
``Dead'' latents are never selected across the full cohort.
``${>}1\%$'' counts latents selected in more than 1\% of episodes.}
\label{tab:dead_latents}
\small
\begin{tabular}{l rrr rrr}
\toprule
& \multicolumn{3}{c}{MIMIC-III ($N{=}28{,}626$)} & \multicolumn{3}{c}{MIMIC-IV ($N{=}46{,}140$)} \\
\cmidrule(lr){2-4} \cmidrule(lr){5-7}
Backbone & Dead & Dead\% & ${>}1\%$ & Dead & Dead\% & ${>}1\%$ \\
\midrule
Discrete    &  26 &  2.0\% & 471 &  40 &  3.1\% & 554 \\
Discrete+PT &  54 &  4.2\% & 524 &  21 &  1.6\% & 537 \\
HyMaTE      & 104 &  8.1\% & 413 &  36 &  2.8\% & 458 \\
BAT         & 154 & 12.0\% & 381 & 251 & 19.6\% & 365 \\
STraTS      & 295 & 23.0\% & 467 & 311 & 24.3\% & 404 \\
\bottomrule
\end{tabular}
\end{table}

\subsection{Dictionary overview}
\label{app:dict_overview}

\begin{figure}[h]
\centering
\includegraphics[width=\linewidth]{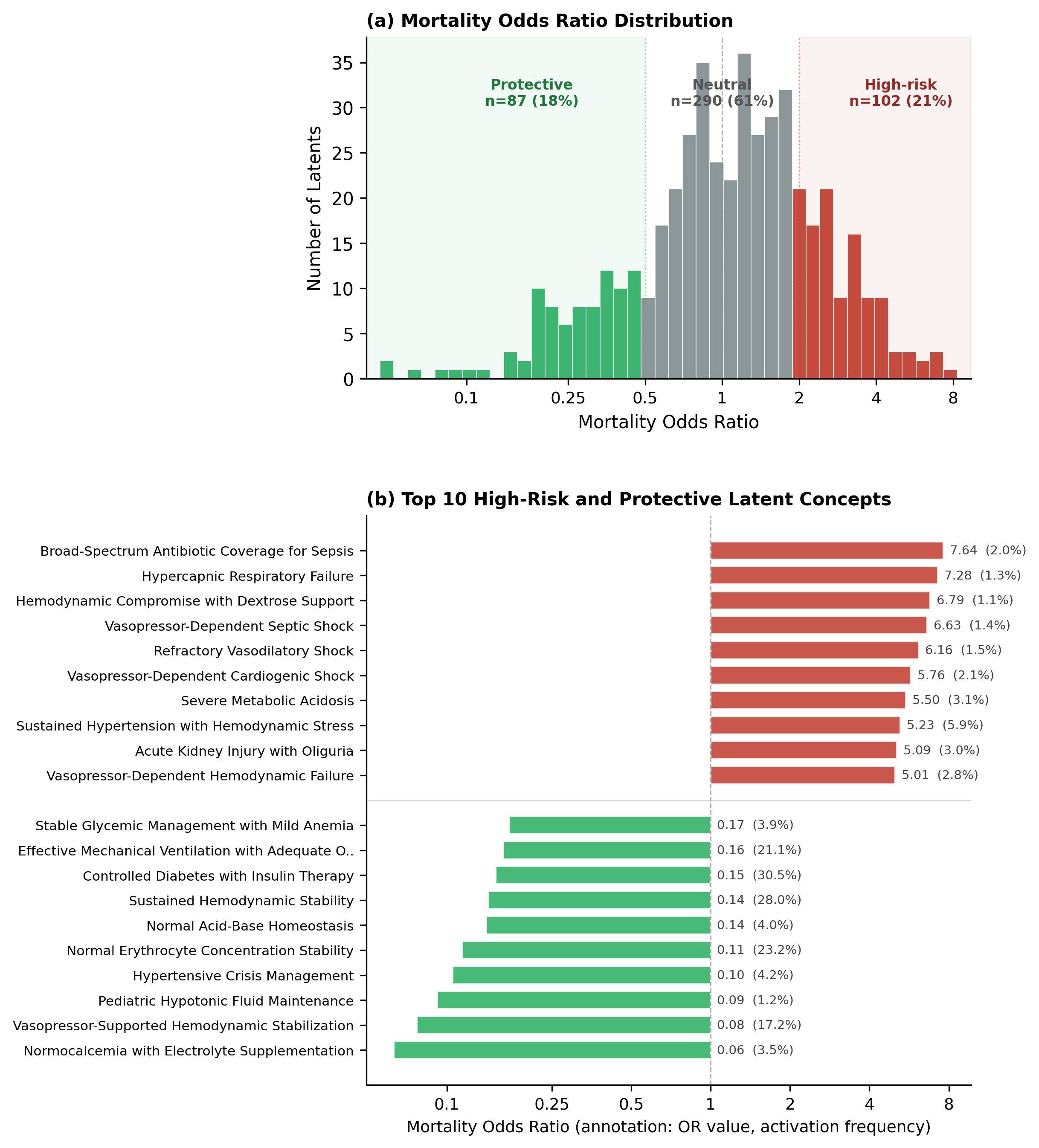}
\caption{Dictionary overview (Discrete backbone, MIMIC-III mortality).
(a)~Mortality odds ratio distribution colored by risk tier: high-risk (OR~${>}$~2, red), neutral (grey), protective (OR~${<}$~0.5, green).
(b)~Top~10 high-risk and protective latent concepts with LLM-generated clinical labels. Annotations show OR and activation frequency.}
\label{fig:dict_overview}
\end{figure}

Figure~\ref{fig:dict_overview} summarizes the 479 latents with activation frequency ${>}1\%$ from the primary dictionary (Discrete backbone, MIMIC-III
mortality).

High-risk latents capture clinically coherent critical-care states (vasopressor dependence, metabolic acidosis, acute kidney injury), while protective latents reflect hemodynamic stability and normal laboratory values.

\paragraph{Concept separability and groupings.}
We measure concept separability by how correlated two concepts' activations are across patients (decoder cosines are uninformative because the sparse atoms are near-orthogonal by construction). The mean absolute pairwise activation correlation is .125, and only 1.4\% of concept pairs exceed .5, indicating that concepts fire in distinct contexts rather than duplicating one another.
The dictionary is fine-grained rather than a few coarse clusters ($k$-means silhouette .073). There is a mild organ-level tendency (concepts within the same organ channel have more similar variable composition, .107 vs.\ .008 across organs), but this partly follows from the channel-purity regularizer.

\paragraph{Statistical significance of outcome association.}
We tested each active concept for association with the outcome (active vs. inactive mortality odds ratio with a 95\% confidence interval; significant if the interval excludes 1). Of 479 concepts, 385 (80\%) are statistically significant: 223 high-risk (OR\,$>$\,1) and 162 protective (OR\,$<$\,1). The remaining 20\% we report as non-significant.

\subsection{Exemplar latent concepts}
\label{app:exemplar_latents}

\begin{figure}[h!]
\centering
\includegraphics[width=\textwidth]{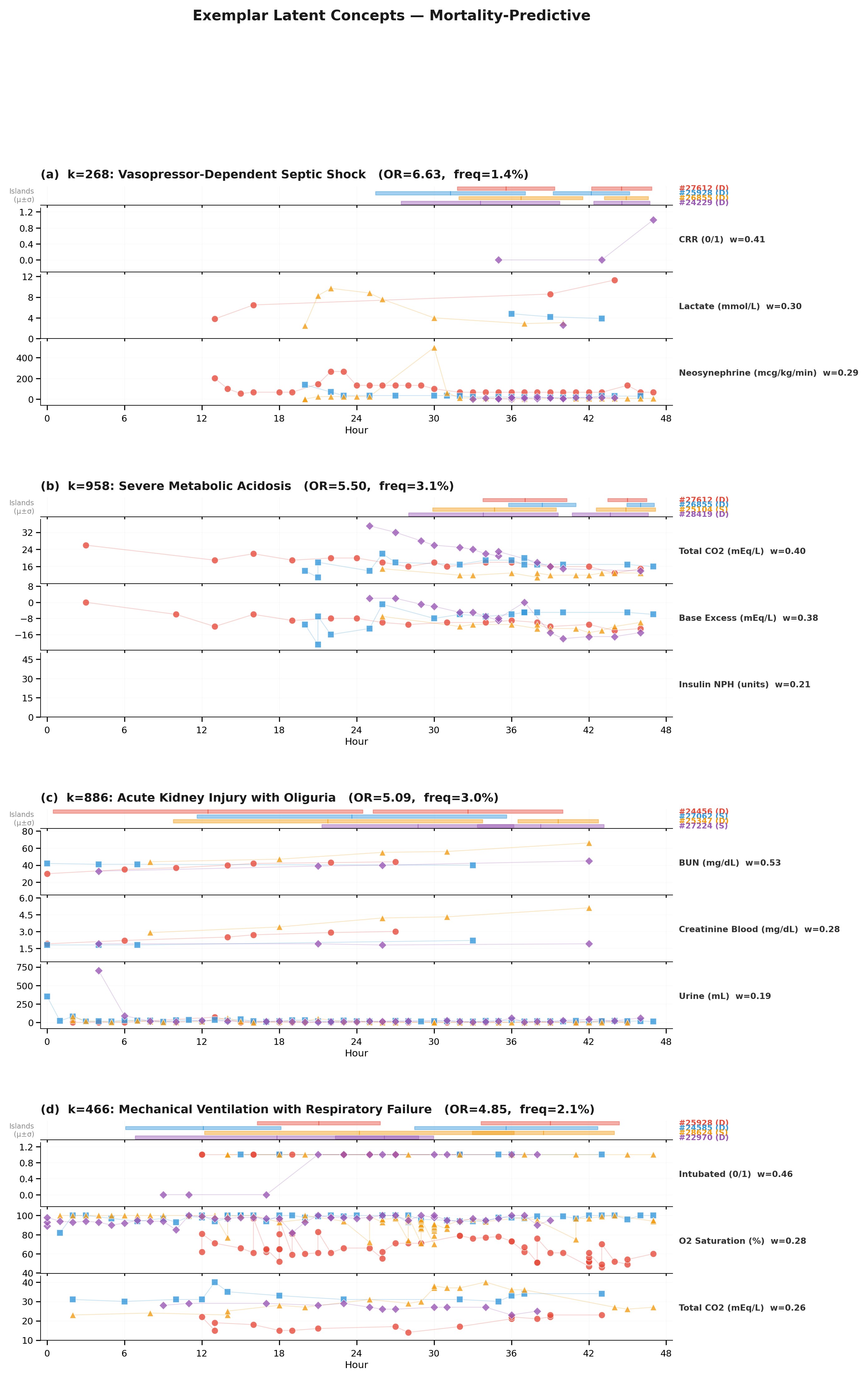}
\caption{Each row: island temporal window (top, colored bars = $\mu \pm \sigma$ per episode) and raw variable traces (bottom) on clinical y-axes. (D)~=~died, (S)~=~survived.}
\label{fig:latent_mortality}
\end{figure}

\begin{figure}[h!]
\centering
\includegraphics[width=\textwidth]{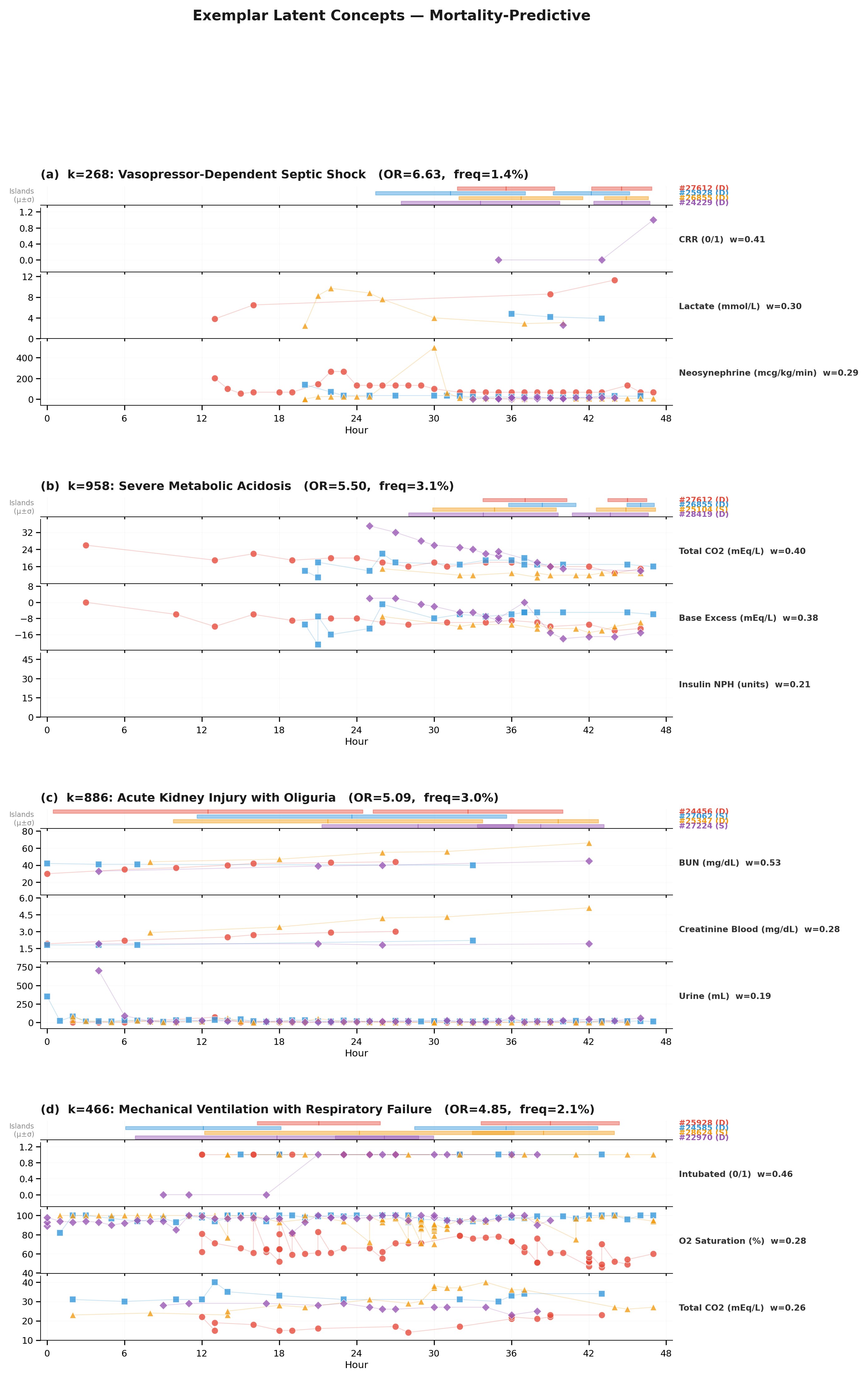}
\caption{Protective exemplar latents.
Same format as Figure~\ref{fig:latent_mortality}.}
\label{fig:latent_protective}
\end{figure}

Figures~\ref{fig:latent_mortality} and~\ref{fig:latent_protective} present exemplar latents from the high-risk and protective groups of the dictionary.
Each panel shows 4 episodes per latent with the island temporal window ($\mu \pm \sigma$, top track) and raw variable traces on clinical y-axes.

\paragraph{Mortality-predictive concepts (Figure~\ref{fig:latent_mortality}).}
(a)~\textbf{Vasopressor-Dependent Hyperlactatemia} (OR=6.63): abnormal capillary refill (CRR$=$1) co-occurs with elevated lactate and vasopressor (neosynephrine) administration, capturing a circulatory shock state with impaired tissue perfusion.
(b)~\textbf{Severe Metabolic Acidosis} (OR=5.50): persistently low total CO$_2$ and deeply negative base excess capture progressive acid--base derangement. The third variable (Insulin NPH) may reflect statistical co-occurrence in diabetic patients, rather than a mechanistic relationship.
(c)~\textbf{Acute Kidney Injury with Oliguria} (OR=5.09): rising BUN and creatinine with declining urine output form a coherent renal failure trajectory across all four episodes.
(d)~\textbf{Mechanical Ventilation with Respiratory Failure}
(OR=4.85): sustained intubation with monitored oxygen saturation and total CO$_2$, capturing the ventilated respiratory failure state.

\paragraph{Protective concepts (Figure~\ref{fig:latent_protective}).}
(a)~\textbf{Stable Acid--Base Balance} (OR=0.14): total CO$_2$ and base excess remain within normal ranges---a direct contrast to the acidosis concept above.
(b)~\textbf{Sustained Hemodynamic Stability} (OR=0.14): dense SBP and DBP readings in normal ranges across the full 48 hours. The third decoder variable (NTproBNP, $w{=}0.18$) has zero observations in all selected episodes, illustrating sparse variables
can contribute decoder weight without contributing observations.
(c)~\textbf{Effective Mechanical Ventilation with Adequate Oxygenation} (OR=0.16): low FiO$_2$ (${\leq}0.5$) with adequate O$_2$ saturation and non-acidotic base excess, indicating controlled ventilation without deterioration.
(d)~\textbf{Sustained Normal Temperature} (OR=0.19): temperature remains in the normal range (${\sim}97$--$99^\circ$F) throughout the stay. Cefazolin and eosinophils are extremely sparse, making this
latent effectively a single-variable detector.

\paragraph{Structural observations.}
These exemplars illustrate three properties of unsupervised concept discovery that merit discussion:
(1)~\emph{Spurious variable grouping}: decoder weights capture statistical co-occurrence, sometimes pairing clinically unrelated variables (e.g., Insulin NPH in the acidosis concept). Whether such groupings reflect genuine latent confounders (e.g., diabetic patients presenting with more severe acidosis) or artifacts of the training distribution is unknown.
(2)~\emph{Observation-frequency imbalance}: when constituent variables differ substantially in measurement frequency, the concept's observable behavior is dominated by the densely sampled variable (e.g., temperature in concept~(d)), even though the decoder assigns meaningful weight to rarer variables. The learned composition remains multi-variable, but its empirical grounding becomes uneven.
(3)~\emph{Latent-space variables without direct observations}: a decoder variable can carry substantial weight yet have no raw measurements in a given episode (e.g., NTproBNP in concept~(b)).
Because the SAE operates on the backbone's contextual
representations---which encode cross-variable dependencies even for unmeasured variables---this may reflect genuine imputed signal rather than a vacuous weight. However, such contributions are not directly verifiable from the clinical record, complicating interpretability.

\clearpage

\subsection{Clinician evaluation of dictionary labels}
\label{app:clinician_eval}

To validate that LLM-generated dictionary labels are clinically meaningful, four independent tertiary-hospital clinicians (all $\geq$10 years of experience) rated a stratified random sample of $N{=}30$ labels (Discrete backbone, MIMIC-III mortality) on a 5-point Likert scale (1: completely wrong; 2: partially related but misleading; 3: vaguely correct but imprecise; 4: mostly accurate, minor issues; 5: accurate and clinically meaningful). Each rater saw the latent's top-3
decoder variables, mortality odds ratio, and up to 10 exemplar episodes with raw variable traces --- the same information shown to the LLM labeler.

\paragraph{Results.}
Mean ratings were 4.20 (rater~1), 4.07 (rater~2), 4.17 (rater~3), and 4.40 (rater~4); 16 of 30 labels
($53\%$) were rated $\geq 4$ by all three raters, 4 of 30 ($20\%$) were rated 5 unanimously, and no label received a rating of 1. Inter-rater agreement on the ordinal scale was fair (quadratic-weighted Cohen's $\kappa = 0.11$--$0.47$, Pearson $r = 0.12$--$0.51$, mean absolute rating difference $0.77$ points).

\begin{table}[h]
\centering
\caption{Per-rater Likert score distributions ($N{=}30$ labels).}
\label{tab:clinician_dict_eval}
\small
\begin{tabular}{l ccccc c c}
\toprule
Rater & 1 & 2 & 3 & 4 & 5 & Mean \\
\midrule
Rater 1 & 0 & 5 & 2 & 6 & 17 & 4.17 \\
Rater 2 & 0 & 1 & 8 & 9 & 12 & 4.07 \\
Rater 3 & 0 & 2 & 5 & 8 & 15 & 4.20 \\
Rater 4 & 0 & 0 & 3 & 12 & 15 & 4.40 \\
\bottomrule
\end{tabular}
\end{table}

Of the six labels rated $\leq 2$ by at least one rater, the most common disagreements were: (i)~labels naming a specific etiology (e.g., \emph{Cardiogenic Shock}) where decoder variables alone could not distinguish among shock subtypes; (ii)~labels for protective concepts where exemplar episodes showed mixed trajectories.

\subsection{Concept ablation}
\label{app:concept_knockout}

We zero-ablate the top-$k$ importance-ranked concepts (by \predselector{} score) and measure the prediction shift $\Delta p = |\sigma(\ell) - \sigma(\ell_{\text{ko}})|$, where $\ell$ is the full backbone prediction and $\ell_{\text{ko}}$ is the prediction after reconstructing from the remaining concepts through the SAE decoder and backbone prediction head.
As a control, we repeat the experiment with $k$ concepts removed in random order (averaged over 10 trials).

\begin{figure}[h]
  \centering
  \begin{subfigure}[t]{0.65\textwidth}
      \centering
      \includegraphics[width=\textwidth]{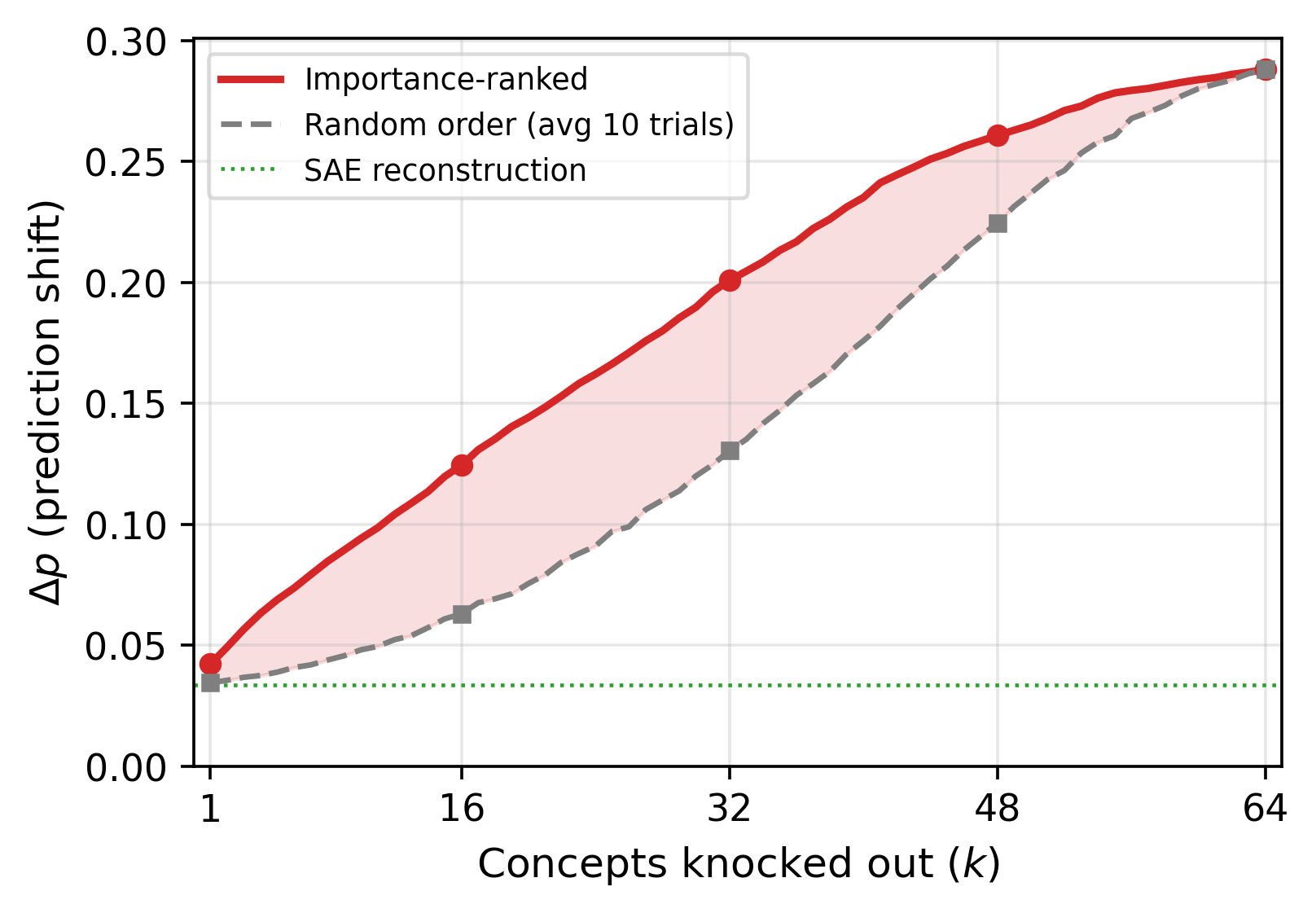}
      \caption{In-distribution (MIMIC-III)}
      \label{fig:knockout_id}
  \end{subfigure}

  \vspace{0.3em}
  
  \begin{subfigure}[t]{0.65\textwidth}
      \centering
      \includegraphics[width=\textwidth]{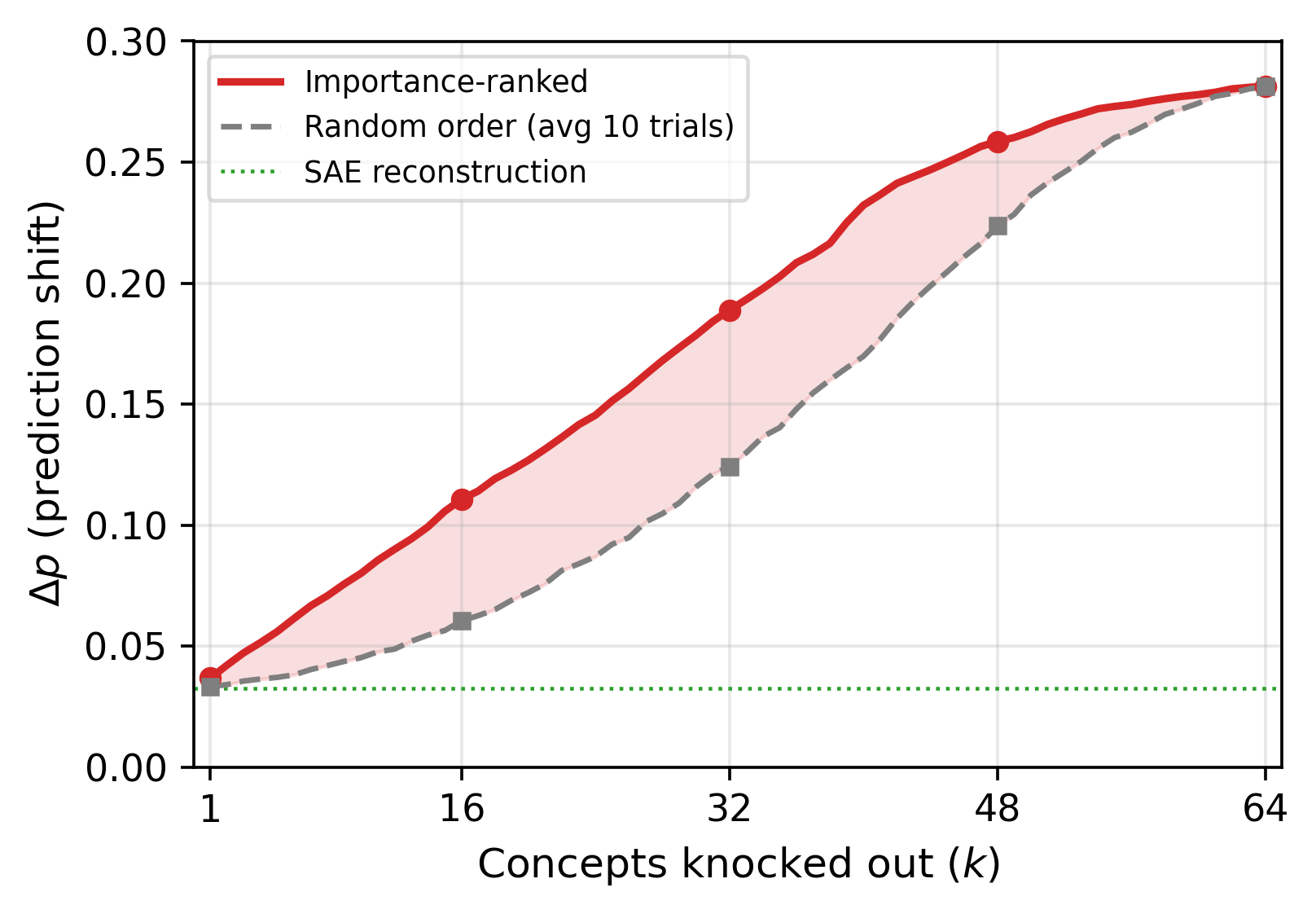}
      \caption{Out-of-distribution (MIMIC-IV)}
      \label{fig:knockout_ood}
  \end{subfigure}
  \caption{
  Concept ablation.
  Prediction shift $\Delta p$ when zero-ablating the top-$k$ concepts from the selected set. Importance-ranked removal (red) degrades predictions more than random-order removal (grey) at all $k$ in both settings. Dashed green: SAE reconstruction baseline (no concepts removed).
  Discrete backbone, mortality task, $N{=}500$.
  }
  \label{fig:knockout}
\end{figure}

Figure~\ref{fig:knockout_id} shows monotonically increasing $\Delta p$ as more concepts are removed.
Importance-ranked removal consistently exceeds random-order removal across all $k$, with the largest gap at $k{=}26$ ($\Delta p = 0.170$ vs.\ $0.099$).
The two curves converge at $k{=}64$ (all selected concepts removed), showing that the gap reflects ranking quality.

\clearpage

\subsection{Concept seed reproducibility}
\label{app:seed_stability}

We assess whether \tisae{}'s learned concepts are reproducible across random seeds by training three models (seeds 42, 43, 44) on the same backbone (Discrete, MIMIC-III mortality) with identical hyperparameters.

In an overcomplete dictionary ($K{=}1280$), multiple latents may share similar decoder weights but differ in temporal behavior.
Matching by decoder similarity can pair temporally unrelated latents, therefore, we use two-stage alignment: optimal (Hungarian) assignment on decoder cosine similarity, refined by selecting the best temporal trace
correlation among structurally similar candidates ($\cos > 0.7$).
The random baseline applies the same temporal search with the same per-latent candidate counts, but draws candidates at random instead of by decoder similarity.

We evaluate five metrics on matched pairs:
\emph{Decoder cosine} measures the cosine similarity of matched decoder weight vectors $\mathbf{w}_k$.
\emph{Variable overlap} reports the fraction of pairs sharing at least two of three top decoder variables (by $|\mathbf{w}_k|$).
\emph{Latent energy correlation} is the Pearson correlation of per-latent activation energy across test samples, measuring whether matched concepts activate on the same patients.
\emph{Latent temporal cosine} is the mean cosine similarity of per-sample activation traces $\hat{A}[k, :]$ ($T{=}48$ time steps), measuring whether matched concepts fire at the same times on the same patients.
\emph{Prediction selector NDCG} is binary NDCG@64: treating one seed's top-64 \predselector{}-ranked latents as relevant, it measures how well the other seed's ranking recovers them.

\begin{table}[h!]
\caption{Concept seed reproducibility ($K{=}1280$, Discrete backbone, MIMIC-III mortality). 
Mean $\pm$ std over 3 seed pairs.}
\label{tab:seed_stability}
\centering
\small
\begin{tabular}{lcc}
\toprule
Metric & Actual & Random \\
\midrule
Decoder cosine                       & $.724 {\pm} .003$ & $.034$ \\
Variable overlap (${\geq}2/3$)       & $77 {\pm} 1\%$    & $1.2\%$ \\
Latent energy correlation            & $.408 {\pm} .006$ & $.093$ \\
Latent temporal cosine               & $.724 {\pm} .003$ & $.382$ \\
Prediction selector NDCG             & $.401 {\pm} .016$ & $.074$ \\
\bottomrule
\end{tabular}
\end{table}

Structural reproducibility is strong: 77\% of matched latent pairs share at least two of three decoder variables (vs.\ 1.2\% random), with mean decoder cosine similarity of .724.
Temporal firing patterns are moderately reproducible, indicating that matched concepts activate at similar time steps on the same patients.
The \predselector{} preserves importance ranking across seeds, though competitive top-$k$ selection introduces variability.

\paragraph{Per-stage reproducibility.}
The analysis above retrains both stages per seed, so dictionary and selector variabilities are mixed.
Here we separate the stages: (i) retraining only the dictionary across five seeds (42--46) on the fixed backbone; and (ii) fixing the seed-42 dictionary and retraining only the \predselector{} across five seeds, measuring per-patient consistency of the selected concepts against a random-subset null.

\begin{table}[h]
\caption{Per-stage seed reproducibility (K=1280, Discrete backbone, MIMIC-III mortality). Stage~A: dictionary retrained, selector excluded. Stage~B: selector retrained on the fixed seed-42 dictionary. Stage~B metrics are computed per patient and averaged over patients and seed pairs. Mean $\pm$ std over 5 seed pairs.}
\label{tab:seed_stability_per_stage}
\centering
\small
\begin{tabular}{l l cc}
\toprule
Stage & Metric & Actual & Random null \\
\midrule
A --- dictionary & Decoder cosine & $.735 \pm .002$ & $.029$ \\
 & Variable overlap (${\geq}2/3$) & $82.4\% \pm 1.1\%$ & $0.8\%$ \\
 & Latent energy correlation & $.361 \pm .010$ & $.083$ \\
 & Latent temporal cosine & $.731 \pm .013$ & $.398$ \\
\midrule
B --- selector (dict.\ fixed) & Top-64 concept Jaccard & $.708 \pm .017$ & $.026$ \\
 & PS-score rank corr.\ (Spearman) & $.870 \pm .029$ & ${\sim}0$ \\
\bottomrule
\end{tabular}
\end{table}

Every metric is well above its matched null.
With the dictionary fixed, the selector's importance ranking is highly consistent across seeds (Spearman .87) and its faithfulness is stable (neighborhood-based AOPC $.151 \pm .008$).

\subsection{Cross-dataset generalization}
\label{app:cross_dataset}

We evaluate whether the dictionary learned on MIMIC-III transfers to out-of-distribution data without retraining.
Because MIMIC-III and MIMIC-IV share patients admitted in 2008--2012, we construct a disjoint cohort using MIMIC-IV's \texttt{anchor\_year\_group}, restricting it to 2014--2016 and 2017--2019 bands.
The MIMIC-III--trained Discrete backbone, TI-SAE, and prediction selector are applied directly to MIMIC-IV mortality data (132 of 135 variables shared; 3 MIMIC-III--only variables receive no observations). Raw values are re-quantized using MIMIC-III quantile boundaries to match the backbone's learned value embeddings.

Table~\ref{tab:ood} reports in-distribution and out-of-distribution performance. Backbone prediction is maintained (AUROC .861$\to$.864). Faithfulness drops modestly (AOPC .149$\to$.130), while clinical plausibility is nearly unchanged (AUROC .756$\to$.756). Of 1280 latents, 500 are selected in ${\geq}$1\% of MIMIC-IV episodes (vs.\ 479 on MIMIC-III), while 440 are never selected---indicating that some MIMIC-III--specific concepts do not appear in the MIMIC-IV population.
The concept knockout curve (Figure~\ref{fig:knockout_ood}) retains the same monotonic shape and importance-ranked vs.\ random-order gap on OOD data (Appendix~\ref{app:concept_knockout}), showing that individual concepts remain causally meaningful in a new patient population. While the disjoint cohort eliminates patient and temporal overlap, both datasets originate from the same institution, so this evaluation captures temporal and population shift rather than fully external, cross-institutional generalization.

\begin{table}[h]
\centering
\caption{Cross-dataset generalization. MIMIC-III--trained Discrete backbone evaluated on disjoint MIMIC-IV mortality ($N{=}500$ for faithfulness and plausibility; prediction AUROC on the full sets).
Neighborhood-based perturbation strategy.
Baselines recomputed on MIMIC-IV.}
\label{tab:ood}
\small
\begin{tabular}{l cc}
\toprule
& In-distribution & Out-of-distribution \\
& (MIMIC-III) & (MIMIC-IV) \\
\midrule
\multicolumn{3}{l}{\textit{Backbone}} \\
Prediction AUROC & .861 & .864 \\
\midrule
\multicolumn{3}{l}{\textit{Faithfulness (AOPC $\uparrow$, neighborhood-based)}} \\
\tisae{} (soft) & .149 & .130 \\
\tisae{} (hard) & .134 & .121 \\
Attention & .100 & .097 \\
IG & .086 & .045 \\
TSR & .088 & .046 \\
DynaMask & .028 & .021 \\
TimeSHAP & .009 & .012 \\
WinIT & .035 & .028 \\
Random & .001 & .005 \\
\midrule
\multicolumn{3}{l}{\textit{Clinical Plausibility (AUROC / AUPRC $\uparrow$)}} \\
\tisae{} (soft) & .756 / .386 & .756 / .370 \\
\tisae{} (hard) & .639 / .315 & .652 / .306 \\
Attention & .616 / .259 & .631 / .261 \\
IG & .567 / .255 & .552 / .224 \\
TSR & .593 / .280 & .587 / .245 \\
DynaMask & .507 / .213 & .505 / .201 \\
TimeSHAP & .521 / .212 & .516 / .203 \\
WinIT & .601 / .265 & .601 / .253 \\
Random & .502 / .193 & .503 / .186 \\
\midrule
\multicolumn{3}{l}{\textit{Dictionary}} \\
$\geq$1\% & 479 & 500 \\
Dead & 26 & 440 \\
\bottomrule
\end{tabular}
\end{table}

\subsection{Cross-backbone concept reuse}
\label{app:cross_backbone}

We investigate whether \tisae{} discovers similar latent concepts across different backbones by comparing the Discrete backbone's dictionary against each of the four remaining backbones (Disc.+PT, HyMaTE, STraTS, BAT), all trained on the MIMIC-III mortality task. 
We apply the same two-stage matching used for seed reproducibility

Table~\ref{tab:cross_backbone} reports five metrics for each backbone pair (Discrete vs.\ target). \tisae{} consistently discovers similar
variable-level groupings across backbones. 
However, latent energy correlation, temporal trace cosine, and prediction selector agreement are substantially lower, indicating that the temporal concepts identified by models and predictive importance of matched concepts diverge across backbones.

\begin{table}[h]
\centering
\caption{Cross-backbone concept reuse (reference: Discrete, MIMIC-III mortality)}
\label{tab:cross_backbone}
\small
\begin{tabular}{l ccccc}
\toprule
Backbone & Dec.\ cos & Var ${\geq}$2/3 & Energy corr & Temp.\ cos & PS NDCG \\
\midrule
Disc.+PT & .801 & 88.2\% & .230 & .674 & .163 \\
HyMaTE   & .731 & 77.5\% & .246 & .592 & .155 \\
BAT      & .716 & 82.0\% & .228 & .583 & .089 \\
STraTS   & .712 & 82.0\% & .108 & .596 & .099 \\
\midrule
Random   & .029 &  0.7\% & .051 & .390 & .055 \\
\bottomrule
\end{tabular}
\end{table}

Disc.+PT shows the strongest agreement (which shares same architecture).
The results suggest that the decoder's variable groupings are largely data-driven, but the temporal pattern grouping and functional importance over those groupings are shaped by each backbone's learned representation.

\clearpage

\section{Extended evaluation results}
\label{app:extended_results}

\subsection{Bootstrap standard deviations and alternative perturbation strategies}

Table~\ref{tab:aopc_neighborhood_full} reproduces the faithfulness results from Table~\ref{tab:main_results} with bootstrap standard deviations (2000 resamples over tasks $\times$ datasets $\times$ samples). Table~\ref{tab:plausibility_std} reports the same for plausibility AUROC and AUPRC.
Tables~\ref{tab:aopc_removal} and~\ref{tab:aopc_marginal} present AOPC under removal and marginal perturbation, averaged across tasks and datasets per backbone.

\begin{table}[h!]
\caption{AOPC under \textbf{neighborhood-based} perturbation (all methods), averaged across tasks and datasets. Best per column in \textbf{bold},
second-best \underline{underlined}. $\pm$ bootstrap standard deviation (2000 within-condition resamples).}
\label{tab:aopc_neighborhood_full}
\centering
\small
\begin{tabular}{lccccc}
\toprule
& Discrete & Disc.+PT & HyMaTE & BAT & STraTS \\
\midrule
\tisae{} (soft) & \textbf{.086}$\pm$.003 & \textbf{.076}$\pm$.003 & \textbf{.097}$\pm$.004 & \textbf{.043}$\pm$.002 & .016$\pm$.001 \\
\tisae{} (hard) & \underline{.079}$\pm$.003 & \underline{.068}$\pm$.002 & \underline{.090}$\pm$.003 & \underline{.041}$\pm$.002 & .016$\pm$.001 \\
Attention & .053$\pm$.002 & .019$\pm$.002 & .051$\pm$.003 & -- & .016$\pm$.001 \\
IG & .060$\pm$.003 & .051$\pm$.002 & .049$\pm$.003 & .031$\pm$.003 & .067$\pm$.003 \\
TSR & .063$\pm$.003 & .054$\pm$.003 & .062$\pm$.003 & .037$\pm$.003 & \underline{.071}$\pm$.003 \\
DynaMask & .019$\pm$.002 & .032$\pm$.002 & .012$\pm$.001 & .028$\pm$.002 & .034$\pm$.002 \\
TimeSHAP & .009$\pm$.001 & .011$\pm$.001 & .012$\pm$.001 & .016$\pm$.002 & .029$\pm$.002 \\
FIT & .011$\pm$.002 & .011$\pm$.001 & .022$\pm$.002 & .012$\pm$.001 & .021$\pm$.002 \\
WinIT & .020$\pm$.002 & .030$\pm$.002 & .032$\pm$.002 & .021$\pm$.002 & .032$\pm$.002 \\
Cluster-TCAV & .046$\pm$.002 & .026$\pm$.001 & .031$\pm$.002 & .023$\pm$.001 & .022$\pm$.001 \\
Random & .002$\pm$.001 & .003$\pm$.000 & .002$\pm$.001 & .006$\pm$.001 & .007$\pm$.001 \\
\bottomrule
\end{tabular}
\end{table}

\begin{table}[h!]
\caption{Plausibility with bootstrap standard deviations, averaged across tasks and datasets. Best per column in \textbf{bold}, second-best
\underline{underlined}. $\pm$ bootstrap standard deviation (2000 within-condition resamples).}
\label{tab:plausibility_std}
\centering
\small
\begin{tabular}{lccccc}
\toprule
& Discrete & Disc.+PT & HyMaTE & BAT & STraTS \\
\midrule
\multicolumn{6}{l}{\textit{AUROC $\uparrow$}} \\
\midrule
\tisae{} (soft) & \textbf{.593}$\pm$.002 & \underline{.589}$\pm$.002 & \underline{.613}$\pm$.002 & \underline{.588}$\pm$.002 & .536$\pm$.002 \\
\tisae{} (hard) & .569$\pm$.002 & .568$\pm$.002 & .580$\pm$.002 & .567$\pm$.002 & .535$\pm$.002 \\
Attention & .570$\pm$.002 & .539$\pm$.002 & .554$\pm$.002 & -- & .552$\pm$.002 \\
IG & .545$\pm$.002 & .548$\pm$.002 & .596$\pm$.002 & .583$\pm$.002 & .576$\pm$.002 \\
TSR & .566$\pm$.002 & .573$\pm$.002 & \textbf{.625}$\pm$.002 & \textbf{.608}$\pm$.002 & \underline{.594}$\pm$.002 \\
DynaMask & .510$\pm$.001 & .504$\pm$.001 & .494$\pm$.002 & .493$\pm$.001 & .497$\pm$.002 \\
\bottomrule
\end{tabular}
\end{table}

\begin{table}[h!]
\caption{AOPC under \textbf{removal} perturbation, averaged across tasks and datasets. Best per column in \textbf{bold}, second-best \underline{underlined}. $\pm$ bootstrap standard deviation (2000 within-condition resamples).}
\label{tab:aopc_removal}
\centering
\small
\begin{tabular}{lccccc}
\toprule
& Discrete & Disc.+PT & HyMaTE & BAT & STraTS \\
\midrule
\tisae{} (soft)  & \textbf{.197}$\pm$.005 & \textbf{.071}$\pm$.003 & \underline{.118}$\pm$.003 & .074$\pm$.004 & .067$\pm$.003 \\
\tisae{} (hard)  & .172$\pm$.004 & .066$\pm$.003 & .102$\pm$.003 & .081$\pm$.003 & .053$\pm$.002 \\
Attention        & \underline{.191}$\pm$.006 & .068$\pm$.004 & \textbf{.143}$\pm$.004 & -- & .038$\pm$.002 \\
IG               & .112$\pm$.005 & .061$\pm$.003 & .066$\pm$.004 & .091$\pm$.005 & .095$\pm$.003 \\
TSR              & .112$\pm$.005 & .060$\pm$.004 & .068$\pm$.003 & \textbf{.100}$\pm$.005 & \underline{.100}$\pm$.003 \\
DynaMask         & .070$\pm$.003 & .068$\pm$.003 & .098$\pm$.003 & .043$\pm$.003 & \textbf{.252}$\pm$.004 \\
TimeSHAP         & .038$\pm$.002 & .019$\pm$.001 & .033$\pm$.002 & .037$\pm$.002 & .036$\pm$.002 \\
FIT              & .036$\pm$.004 & .028$\pm$.003 & .041$\pm$.003 & .046$\pm$.003 & .025$\pm$.002 \\
WinIT            & .036$\pm$.004 & .027$\pm$.003 & .053$\pm$.002 & .064$\pm$.004 & .044$\pm$.002 \\
Cluster-TCAV     & .143$\pm$.003 & \underline{.069}$\pm$.002 & .042$\pm$.002 & \underline{.092}$\pm$.003 & .050$\pm$.002 \\
Random           & .006$\pm$.001 & .005$\pm$.001 & .004$\pm$.001 & .012$\pm$.002 & .009$\pm$.001 \\
\bottomrule
\end{tabular}
\end{table}

\begin{table}[h!]
\caption{AOPC under \textbf{marginal} perturbation, averaged across tasks and datasets. Best per column in \textbf{bold}, second-best \underline{underlined}. $\pm$ bootstrap standard deviation (2000 within-condition resamples).}
\label{tab:aopc_marginal}
\centering
\small
\begin{tabular}{lccccc}
\toprule
& Discrete & Disc.+PT & HyMaTE & BAT & STraTS \\
\midrule
\tisae{} (soft)  & .092$\pm$.003 & .071$\pm$.003 & \textbf{.099}$\pm$.004 & .038$\pm$.003 & .015$\pm$.003 \\
\tisae{} (hard)  & .084$\pm$.003 & .066$\pm$.003 & .090$\pm$.004 & .047$\pm$.003 & .013$\pm$.001 \\
Attention        & .085$\pm$.004 & .033$\pm$.003 & .082$\pm$.004 & -- & .019$\pm$.001 \\
IG               & \textbf{.100}$\pm$.004 & \textbf{.086}$\pm$.003 & \underline{.098}$\pm$.005 & \underline{.069}$\pm$.004 & \textbf{.106}$\pm$.004 \\
TSR              & \underline{.098}$\pm$.004 & \underline{.085}$\pm$.003 & .098$\pm$.005 & \textbf{.071}$\pm$.004 & \underline{.105}$\pm$.004 \\
DynaMask         & .080$\pm$.003 & .043$\pm$.002 & .041$\pm$.003 & .049$\pm$.003 & .076$\pm$.003 \\
TimeSHAP         & .058$\pm$.003 & .047$\pm$.003 & .063$\pm$.004 & .053$\pm$.004 & .087$\pm$.004 \\
FIT              & .048$\pm$.003 & .033$\pm$.002 & .055$\pm$.003 & .024$\pm$.002 & .038$\pm$.002 \\
WinIT            & .052$\pm$.003 & .040$\pm$.003 & .058$\pm$.004 & .031$\pm$.003 & .041$\pm$.002 \\
Cluster-TCAV     & .064$\pm$.003 & .025$\pm$.002 & .027$\pm$.002 & .023$\pm$.002 & .027$\pm$.002 \\
Random           & .022$\pm$.001 & .014$\pm$.001 & .023$\pm$.002 & .014$\pm$.002 & .025$\pm$.002 \\
\bottomrule
\end{tabular}
\end{table}

\subsection{Per-fraction detail}

Tables~\ref{tab:frac_discrete}--\ref{tab:frac_strats_readm} present
per-fraction comprehensiveness and sufficiency under neighborhood-based perturbation for each backbone (MIMIC-III and MIMIC-IV).
Best AOPC per sub-table in \textbf{bold}, second-best \underline{underlined}.

\begin{table}[h!]
\caption{Per-fraction faithfulness: \textbf{Discrete} backbone (neighborhood, mortality).}
\label{tab:frac_discrete}
\centering
\scriptsize
\setlength{\tabcolsep}{3pt}
\begin{tabular}{l ccccc c ccccc}
\toprule
& \multicolumn{5}{c}{Comprehensiveness $\uparrow$} & & \multicolumn{5}{c}{Sufficiency $\downarrow$} \\
\cmidrule(lr){2-6} \cmidrule(lr){8-12}
Method & @.10 & @.20 & @.30 & @.40 & @.50 & AOPC & @.10 & @.20 & @.30 & @.40 & @.50 \\
\midrule
\multicolumn{12}{l}{\textit{MIMIC-III}} \\
\midrule
\tisae{} (soft)  & .082 & .166 & .200 & .176 & .146 & \textbf{.149} & $-$.004 & .009 & .019 & .017 & .011 \\
\tisae{} (hard)  & .074 & .149 & .168 & .160 & .150 & \underline{.134} & .007 & .021 & .028 & .014 & .009 \\
Attention & .045 & .091 & .131 & .132 & .128 & .100 & .091 & .059 & .045 & .033 & .021 \\
IG & .057 & .082 & .101 & .108 & .110 & .086 & .164 & .123 & .072 & .055 & .050 \\
TSR & .052 & .079 & .099 & .111 & .121 & .088 & .093 & .072 & .064 & .054 & .050 \\
\bottomrule
\end{tabular}
\end{table}

\begin{table}[h!]
\caption{Per-fraction faithfulness: \textbf{Discrete+PT} backbone (neighborhood, mortality).}
\label{tab:frac_discpt}
\centering
\scriptsize
\setlength{\tabcolsep}{3pt}
\begin{tabular}{l ccccc c ccccc}
\toprule
& \multicolumn{5}{c}{Comprehensiveness $\uparrow$} & & \multicolumn{5}{c}{Sufficiency $\downarrow$} \\
\cmidrule(lr){2-6} \cmidrule(lr){8-12}
Method & @.10 & @.20 & @.30 & @.40 & @.50 & AOPC & @.10 & @.20 & @.30 & @.40 & @.50 \\
\midrule
\multicolumn{12}{l}{\textit{MIMIC-III}} \\
\midrule
\tisae{} (soft)  & .033 & .080 & .109 & .129 & .137 & \textbf{.091} & .022 & .002 & .003 & $-$.001 & $-$.003 \\
\tisae{} (hard)  & .032 & .071 & .093 & .110 & .123 & \underline{.081} & .022 & .008 & .004 & .003 & .015 \\
Attention & .012 & .030 & .042 & .052 & .060 & .036 & .250 & .244 & .220 & .171 & .117 \\
IG  & .035 & .062 & .086 & .103 & .113 & .076 & .065 & .052 & .041 & .036 & .037 \\
TSR  & .038 & .068 & .092 & .108 & .119 & .079 & .032 & .035 & .031 & .033 & .031 \\
\bottomrule
\end{tabular}
\end{table}

\begin{table}[h!]
\caption{Per-fraction faithfulness: \textbf{HyMaTE} backbone (neighborhood, mortality).}
\label{tab:frac_hymate}
\centering
\scriptsize
\setlength{\tabcolsep}{3pt}
\begin{tabular}{l ccccc c ccccc}
\toprule
& \multicolumn{5}{c}{Comprehensiveness $\uparrow$} & & \multicolumn{5}{c}{Sufficiency $\downarrow$} \\
\cmidrule(lr){2-6} \cmidrule(lr){8-12}
Method & @.10 & @.20 & @.30 & @.40 & @.50 & AOPC & @.10 & @.20 & @.30 & @.40 & @.50 \\
\midrule
\multicolumn{12}{l}{\textit{MIMIC-III}} \\
\midrule
\tisae{} (soft)  & .082 & .141 & .172 & .180 & .182 & \textbf{.143} & .071 & .053 & .041 & .026 & .009 \\
\tisae{} (hard)  & .065 & .124 & .156 & .170 & .165 & \underline{.127} & .048 & .040 & .057 & .051 & .036 \\
Attention        & .041 & .086 & .115 & .136 & .140 & .096 & .035 & .042 & .027 & .017 & .015 \\
IG               & .049 & .079 & .094 & .108 & .116 & .085 & .118 & .088 & .050 & .030 & .018 \\
TSR              & .051 & .086 & .110 & .120 & .133 & .094 & .070 & .054 & .031 & .016 & .011 \\
\bottomrule
\end{tabular}
\end{table}

\begin{table}[h!]
\caption{Per-fraction faithfulness: \textbf{BAT} backbone (neighborhood, mortality).}
\label{tab:frac_bat}
\centering
\scriptsize
\setlength{\tabcolsep}{3pt}
\begin{tabular}{l ccccc c ccccc}
\toprule
& \multicolumn{5}{c}{Comprehensiveness $\uparrow$} & & \multicolumn{5}{c}{Sufficiency $\downarrow$} \\
\cmidrule(lr){2-6} \cmidrule(lr){8-12}
Method & @.10 & @.20 & @.30 & @.40 & @.50 & AOPC & @.10 & @.20 & @.30 & @.40 & @.50 \\
\midrule
\multicolumn{12}{l}{\textit{MIMIC-III}} \\
\midrule
\tisae{} (soft)  & .022 & .052 & .076 & .091 & .109 & \underline{.065} & .065 & .067 & .064 & .067 & .052 \\
\tisae{} (hard)  & .021 & .055 & .080 & .099 & .121 & \textbf{.070} & .055 & .039 & .050 & .048 & .055 \\
Attention        & -- & -- & -- & -- & -- & -- & -- & -- & -- & -- & -- \\
IG               & .023 & .055 & .064 & .075 & .089 & .057 & .095 & .079 & .056 & .031 & .024 \\
TSR              & .024 & .050 & .066 & .081 & .095 & .059 & .059 & .045 & .030 & .017 & .012 \\
\bottomrule
\end{tabular}
\end{table}

\begin{table}[h!]
\caption{Per-fraction faithfulness: \textbf{STraTS} backbone (neighborhood, mortality).}
\label{tab:frac_strats}
\centering
\scriptsize
\setlength{\tabcolsep}{3pt}
\begin{tabular}{l ccccc c ccccc}
\toprule
& \multicolumn{5}{c}{Comprehensiveness $\uparrow$} & & \multicolumn{5}{c}{Sufficiency $\downarrow$} \\
\cmidrule(lr){2-6} \cmidrule(lr){8-12}
Method & @.10 & @.20 & @.30 & @.40 & @.50 & AOPC & @.10 & @.20 & @.30 & @.40 & @.50 \\
\midrule
\multicolumn{12}{l}{\textit{MIMIC-III}} \\
\midrule
\tisae{} (soft)  & .004 & .007 & .023 & .024 & .025 & .015 & .089 & .080 & .070 & .060 & .044 \\
\tisae{} (hard)  & .003 & .008 & .017 & .021 & .032 & .013 & .080 & .072 & .059 & .051 & .045 \\
Attention        & .004 & .008 & .008 & .010 & .020 & .010 & .089 & .098 & .090 & .083 & .073 \\
IG               & .037 & .058 & .073 & .084 & .084 & \textbf{.064} & .051 & .037 & .021 & .014 & .008 \\
TSR              & .034 & .058 & .071 & .079 & .082 & .061 & .023 & .021 & .012 & .007 & .004 \\
\bottomrule
\end{tabular}
\end{table}

\begin{table}[h!]
\caption{Per-fraction faithfulness: \textbf{Discrete} backbone (neighborhood, length-of-stay).}
\label{tab:frac_discrete_los}
\centering
\scriptsize
\setlength{\tabcolsep}{3pt}
\begin{tabular}{l ccccc c ccccc}
\toprule
& \multicolumn{5}{c}{Comprehensiveness $\uparrow$} & & \multicolumn{5}{c}{Sufficiency $\downarrow$} \\
\cmidrule(lr){2-6} \cmidrule(lr){8-12}
Method & @.10 & @.20 & @.30 & @.40 & @.50 & AOPC & @.10 & @.20 & @.30 & @.40 & @.50 \\
\midrule
\multicolumn{12}{l}{\textit{MIMIC-III}} \\
\midrule
\tisae{} (soft)      & .034 & .040 & .036 & .046 & .062 & .041 & .064 & .046 & .037 & .030 & .020 \\
\tisae{} (hard)      & .032 & .043 & .037 & .042 & .051 & .039 & .060 & .050 & .043 & .034 & .015 \\
Attention            & .011 & .036 & .050 & .051 & .055 & .037 & .047 & .025 & .016 & .015 & .013 \\
IG                   & .021 & .038 & .057 & .071 & .086 & \underline{.050} & .016 & $-$.012 & $-$.024 & $-$.023 & $-$.017 \\
TSR                  & .021 & .035 & .050 & .071 & .089 & .049 & .011 & $-$.007 & $-$.013 & $-$.012 & $-$.011 \\
\bottomrule
\end{tabular}
\end{table}

\begin{table}[h!]
\caption{Per-fraction faithfulness: \textbf{Disc.+PT} backbone (neighborhood, length-of-stay).}
\label{tab:frac_discpt_los}
\centering
\scriptsize
\setlength{\tabcolsep}{3pt}
\begin{tabular}{l ccccc c ccccc}
\toprule
& \multicolumn{5}{c}{Comprehensiveness $\uparrow$} & & \multicolumn{5}{c}{Sufficiency $\downarrow$} \\
\cmidrule(lr){2-6} \cmidrule(lr){8-12}
Method & @.10 & @.20 & @.30 & @.40 & @.50 & AOPC & @.10 & @.20 & @.30 & @.40 & @.50 \\
\midrule
\multicolumn{12}{l}{\textit{MIMIC-III}} \\
\midrule
\tisae{} (soft)      & .010 & .025 & .043 & .057 & .068 & \textbf{.037} & .031 & .019 & .013 & .006 & .001 \\
\tisae{} (hard)      & .007 & .022 & .038 & .055 & .061 & .034 & .033 & .023 & .016 & .007 & .003 \\
Attention            & $-$.012 & $-$.006 & .000 & .010 & .025 & .002 & .076 & .050 & .047 & .041 & .029 \\
IG                   & .010 & .022 & .034 & .049 & .058 & .032 & .035 & .017 & .007 & .003 & .003 \\
TSR                  & .011 & .028 & .036 & .049 & .061 & .034 & .024 & .012 & .007 & .005 & .004 \\
\bottomrule
\end{tabular}
\end{table}

\begin{table}[h!]
\caption{Per-fraction faithfulness: \textbf{HyMaTE} backbone (neighborhood, length-of-stay).}
\label{tab:frac_hymate_los}
\centering
\scriptsize
\setlength{\tabcolsep}{3pt}
\begin{tabular}{l ccccc c ccccc}
\toprule
& \multicolumn{5}{c}{Comprehensiveness $\uparrow$} & & \multicolumn{5}{c}{Sufficiency $\downarrow$} \\
\cmidrule(lr){2-6} \cmidrule(lr){8-12}
Method & @.10 & @.20 & @.30 & @.40 & @.50 & AOPC & @.10 & @.20 & @.30 & @.40 & @.50 \\
\midrule
\multicolumn{12}{l}{\textit{MIMIC-III}} \\
\midrule
\tisae{} (soft)      & .112 & .129 & .142 & .138 & .130 & \underline{.124} & $-$.039 & $-$.029 & $-$.023 & $-$.018 & $-$.008 \\
\tisae{} (hard)      & .117 & .131 & .147 & .149 & .144 & \textbf{.130} & $-$.028 & $-$.030 & $-$.017 & .004 & .005 \\
Attention            & .053 & .064 & .074 & .074 & .085 & .067 & .016 & .024 & .017 & .011 & .008 \\
IG                   & .033 & .045 & .058 & .069 & .082 & .054 & .055 & .045 & .027 & .019 & .012 \\
TSR                  & .065 & .089 & .096 & .107 & .112 & .089 & .034 & .024 & .013 & .010 & .007 \\
\bottomrule
\end{tabular}
\end{table}

\begin{table}[h!]
\caption{Per-fraction faithfulness: \textbf{BAT} backbone (neighborhood, length-of-stay).}
\label{tab:frac_bat_los}
\centering
\scriptsize
\setlength{\tabcolsep}{3pt}
\begin{tabular}{l ccccc c ccccc}
\toprule
& \multicolumn{5}{c}{Comprehensiveness $\uparrow$} & & \multicolumn{5}{c}{Sufficiency $\downarrow$} \\
\cmidrule(lr){2-6} \cmidrule(lr){8-12}
Method & @.10 & @.20 & @.30 & @.40 & @.50 & AOPC & @.10 & @.20 & @.30 & @.40 & @.50 \\
\midrule
\multicolumn{12}{l}{\textit{MIMIC-III}} \\
\midrule
\tisae{} (soft)      & .011 & .017 & .024 & .028 & .038 & .022 & .040 & .026 & .019 & .010 & .002 \\
\tisae{} (hard)      & .010 & .017 & .025 & .025 & .034 & .020 & .036 & .028 & .018 & .011 & .003 \\
Attention            & -- & -- & -- & -- & -- & -- & -- & -- & -- & -- & -- \\
IG                   & .016 & .023 & .029 & .032 & .036 & .025 & .037 & .023 & .016 & .011 & .008 \\
TSR                  & .017 & .034 & .037 & .043 & .047 & \textbf{.034} & .019 & .013 & .009 & .005 & .004 \\
\bottomrule
\end{tabular}
\end{table}

\begin{table}[h!]
\caption{Per-fraction faithfulness: \textbf{STraTS} backbone (neighborhood, length-of-stay).}
\label{tab:frac_strats_los}
\centering
\scriptsize
\setlength{\tabcolsep}{3pt}
\begin{tabular}{l ccccc c ccccc}
\toprule
& \multicolumn{5}{c}{Comprehensiveness $\uparrow$} & & \multicolumn{5}{c}{Sufficiency $\downarrow$} \\
\cmidrule(lr){2-6} \cmidrule(lr){8-12}
Method & @.10 & @.20 & @.30 & @.40 & @.50 & AOPC & @.10 & @.20 & @.30 & @.40 & @.50 \\
\midrule
\multicolumn{12}{l}{\textit{MIMIC-III}} \\
\midrule
\tisae{} (soft)      & .014 & .025 & .026 & .021 & .032 & .023 & .135 & .143 & .131 & .127 & .108 \\
\tisae{} (hard)      & .010 & .021 & .032 & .028 & .027 & .023 & .131 & .134 & .121 & .122 & .110 \\
Attention            & .012 & .012 & .015 & .029 & .040 & .020 & .120 & .106 & .094 & .080 & .061 \\
IG                   & .044 & .052 & .058 & .073 & .085 & .060 & .022 & .018 & .011 & .006 & .002 \\
TSR                  & .048 & .064 & .084 & .106 & .123 & \textbf{.080} & .013 & .007 & .005 & .003 & .002 \\
\bottomrule
\end{tabular}
\end{table}

\begin{table}[h!]
\caption{Per-fraction faithfulness: \textbf{Discrete} backbone (neighborhood, readmission).}
\label{tab:frac_discrete_readm}
\centering
\scriptsize
\setlength{\tabcolsep}{3pt}
\begin{tabular}{l ccccc c ccccc}
\toprule
& \multicolumn{5}{c}{Comprehensiveness $\uparrow$} & & \multicolumn{5}{c}{Sufficiency $\downarrow$} \\
\cmidrule(lr){2-6} \cmidrule(lr){8-12}
Method & @.10 & @.20 & @.30 & @.40 & @.50 & AOPC & @.10 & @.20 & @.30 & @.40 & @.50 \\
\midrule
\multicolumn{12}{l}{\textit{MIMIC-III}} \\
\midrule
\tisae{} (soft)      & .066 & .112 & .126 & .127 & .128 & \textbf{.106} & $-$.063 & $-$.049 & $-$.041 & $-$.029 & $-$.022 \\
\tisae{} (hard)      & .054 & .090 & .102 & .110 & .110 & \underline{.088} & $-$.050 & $-$.037 & $-$.019 & $-$.006 & $-$.006 \\
Attention            & .025 & .051 & .064 & .078 & .088 & .056 & .003 & .010 & .010 & .008 & .004 \\
IG                   & .037 & .054 & .058 & .059 & .060 & .051 & .007 & .017 & .018 & .015 & .010 \\
TSR                  & .035 & .052 & .058 & .060 & .064 & .051 & .008 & .009 & .005 & .005 & .004 \\
\bottomrule
\end{tabular}
\end{table}

\begin{table}[h!]
\caption{Per-fraction faithfulness: \textbf{Disc.+PT} backbone (neighborhood, readmission).}
\label{tab:frac_discpt_readm}
\centering
\scriptsize
\setlength{\tabcolsep}{3pt}
\begin{tabular}{l ccccc c ccccc}
\toprule
& \multicolumn{5}{c}{Comprehensiveness $\uparrow$} & & \multicolumn{5}{c}{Sufficiency $\downarrow$} \\
\cmidrule(lr){2-6} \cmidrule(lr){8-12}
Method & @.10 & @.20 & @.30 & @.40 & @.50 & AOPC & @.10 & @.20 & @.30 & @.40 & @.50 \\
\midrule
\multicolumn{12}{l}{\textit{MIMIC-III}} \\
\midrule
\tisae{} (soft)      & .050 & .095 & .117 & .123 & .129 & \textbf{.097} & .008 & $-$.011 & $-$.015 & $-$.009 & $-$.006 \\
\tisae{} (hard)      & .044 & .083 & .099 & .111 & .116 & \underline{.086} & .008 & .001 & .001 & .004 & .020 \\
Attention            & .003 & .005 & .003 & .007 & .010 & .005 & .119 & .103 & .096 & .090 & .083 \\
IG                   & .049 & .062 & .071 & .081 & .090 & .068 & .039 & .034 & .031 & .026 & .023 \\
TSR                  & .050 & .065 & .079 & .093 & .101 & .074 & .025 & .021 & .014 & .015 & .014 \\
\bottomrule
\end{tabular}
\end{table}

\begin{table}[h!]
\caption{Per-fraction faithfulness: \textbf{HyMaTE} backbone (neighborhood, readmission).}
\label{tab:frac_hymate_readm}
\centering
\scriptsize
\setlength{\tabcolsep}{3pt}
\begin{tabular}{l ccccc c ccccc}
\toprule
& \multicolumn{5}{c}{Comprehensiveness $\uparrow$} & & \multicolumn{5}{c}{Sufficiency $\downarrow$} \\
\cmidrule(lr){2-6} \cmidrule(lr){8-12}
Method & @.10 & @.20 & @.30 & @.40 & @.50 & AOPC & @.10 & @.20 & @.30 & @.40 & @.50 \\
\midrule
\multicolumn{12}{l}{\textit{MIMIC-III}} \\
\midrule
\tisae{} (soft)      & .025 & .057 & .077 & .069 & .058 & \textbf{.055} & $-$.024 & $-$.010 & .007 & .014 & .010 \\
\tisae{} (hard)      & .025 & .051 & .066 & .069 & .072 & \underline{.054} & $-$.028 & $-$.025 & $-$.017 & .003 & $-$.003 \\
Attention            & .033 & .035 & .039 & .043 & .048 & .038 & .048 & .044 & .038 & .026 & .016 \\
IG                   & .030 & .033 & .036 & .039 & .041 & .035 & .055 & .049 & .035 & .024 & .018 \\
TSR                  & .033 & .038 & .041 & .042 & .047 & .039 & .042 & .035 & .026 & .018 & .013 \\
\bottomrule
\end{tabular}
\end{table}

\begin{table}[h!]
\caption{Per-fraction faithfulness: \textbf{BAT} backbone (neighborhood, readmission).}
\label{tab:frac_bat_readm}
\centering
\scriptsize
\setlength{\tabcolsep}{3pt}
\begin{tabular}{l ccccc c ccccc}
\toprule
& \multicolumn{5}{c}{Comprehensiveness $\uparrow$} & & \multicolumn{5}{c}{Sufficiency $\downarrow$} \\
\cmidrule(lr){2-6} \cmidrule(lr){8-12}
Method & @.10 & @.20 & @.30 & @.40 & @.50 & AOPC & @.10 & @.20 & @.30 & @.40 & @.50 \\
\midrule
\multicolumn{12}{l}{\textit{MIMIC-III}} \\
\midrule
\tisae{} (soft)      & .005 & .019 & .028 & .038 & .051 & .025 & .084 & .067 & .059 & .047 & .040 \\
\tisae{} (hard)      & .006 & .018 & .030 & .041 & .057 & \underline{.028} & .084 & .074 & .061 & .046 & .035 \\
Attention            & -- & -- & -- & -- & -- & -- & -- & -- & -- & -- & -- \\
IG                   & $-$.002 & .011 & .020 & .033 & .042 & .018 & .143 & .111 & .088 & .063 & .045 \\
TSR                  & $-$.003 & .015 & .034 & .038 & .048 & .023 & .135 & .103 & .074 & .051 & .037 \\
\bottomrule
\end{tabular}
\end{table}

\begin{table}[h!]
\caption{Per-fraction faithfulness: \textbf{STraTS} backbone (neighborhood, readmission).}
\label{tab:frac_strats_readm}
\centering
\scriptsize
\setlength{\tabcolsep}{3pt}
\begin{tabular}{l ccccc c ccccc}
\toprule
& \multicolumn{5}{c}{Comprehensiveness $\uparrow$} & & \multicolumn{5}{c}{Sufficiency $\downarrow$} \\
\cmidrule(lr){2-6} \cmidrule(lr){8-12}
Method & @.10 & @.20 & @.30 & @.40 & @.50 & AOPC & @.10 & @.20 & @.30 & @.40 & @.50 \\
\midrule
\multicolumn{12}{l}{\textit{MIMIC-III}} \\
\midrule
\tisae{} (soft)      & .003 & .008 & .013 & .014 & .017 & .010 & .075 & .074 & .075 & .071 & .061 \\
\tisae{} (hard)      & .003 & .008 & .010 & .014 & .016 & .010 & .076 & .074 & .074 & .074 & .066 \\
Attention            & .002 & .004 & .009 & .013 & .020 & .008 & .073 & .064 & .057 & .047 & .035 \\
IG                   & .055 & .059 & .060 & .058 & .059 & .057 & .013 & .013 & .010 & .007 & .005 \\
TSR                  & .055 & .058 & .059 & .060 & .062 & \underline{.058} & .008 & .009 & .006 & .004 & .003 \\
\bottomrule
\end{tabular}
\end{table}

\clearpage

\subsection{Per-task and per-dataset plausibility}
\label{app:per_task_plaus}

Tables~\ref{tab:plaus_mort_m3}--\ref{tab:plaus_readm_m4} report per-task per-dataset plausibility. Best AUROC in \textbf{bold}, second-best \underline{underlined}.

\begin{table}[h!]
\caption{Per-task plausibility: \textbf{Mortality, MIMIC-III} (AUROC/AUPRC).}
\label{tab:plaus_mort_m3}
\centering
\small
\begin{tabular}{l ccccc}
\toprule
Method & Discrete & Disc.+PT & HyMaTE & BAT & STraTS \\
\midrule
\tisae{} (soft) & \textbf{.756}/.384 & \textbf{.631}/.303 & \textbf{.707}/.356 & .598/.249 & .516/.216 \\
\tisae{} (hard) & \underline{.639}/.313 & \underline{.625}/.289 & .621/.301 & .575/.240 & .506/.218 \\
Attention & .615/.258 & .468/.189 & .568/.231 & -- & .610/.249 \\
IG & .567/.254 & .555/.245 & .641/.278 & \underline{.620}/.265 & .621/.251 \\
TSR & .594/.279 & .586/.262 & \underline{.673}/.320 & \textbf{.649}/.296 & \underline{.636}/.264 \\
DynaMask & .507/.212 & .507/.205 & .492/.189 & .491/.189 & .492/.189 \\
TimeSHAP & .521/.212 & .521/.214 & .525/.214 & .526/.218 & .523/.214 \\
FIT & .523/.205 & .547/.219 & .558/.218 & .489/.191 & .564/.227 \\
WinIT & .601/.264 & .622/.283 & .657/.342 & .584/.261 & \textbf{.644}/.298 \\
Cluster-TCAV & .541/.223 & .520/.217 & .598/.256 & .569/.264 & .494/.207 \\
Random & .502/.193 & .502/.193 & .502/.193 & .502/.193 & .502/.193 \\
\bottomrule
\end{tabular}
\end{table}

\begin{table}[h!]
\caption{Per-task plausibility: \textbf{Mortality, MIMIC-IV} (AUROC/AUPRC).}
\label{tab:plaus_mort_m4}
\centering
\small
\begin{tabular}{l ccccc}
\toprule
Method & Discrete & Disc.+PT & HyMaTE & BAT & STraTS \\
\midrule
\tisae{} (soft) & \textbf{.705}/.362 & .569/.256 & \underline{.611}/.255 & \textbf{.638}/.296 & .545/.248 \\
\tisae{} (hard) & .632/.314 & .533/.245 & .556/.232 & \underline{.621}/.276 & .561/.252 \\
Attention & \underline{.668}/.312 & \textbf{.624}/.268 & .523/.203 & -- & .497/.190 \\
IG & .562/.231 & .530/.218 & .595/.246 & .572/.230 & .541/.208 \\
TSR & .578/.244 & .552/.232 & \textbf{.619}/.269 & .592/.248 & \underline{.564}/.218 \\
DynaMask & .506/.199 & .491/.193 & .495/.188 & .515/.202 & .484/.203 \\
TimeSHAP & .514/.203 & .512/.203 & .520/.208 & .519/.207 & .513/.203 \\
FIT & .523/.204 & .532/.207 & .505/.189 & .495/.191 & .562/.221 \\
WinIT & .601/.270 & \underline{.607}/.276 & .576/.247 & .570/.256 & \textbf{.621}/.279 \\
Cluster-TCAV & .599/.257 & .471/.234 & .513/.222 & .559/.243 & .410/.176 \\
Random & .501/.191 & .501/.191 & .501/.191 & .501/.191 & .501/.191 \\
\bottomrule
\end{tabular}
\end{table}

\begin{table}[h!]
\caption{Per-task plausibility: \textbf{LOS>7d, MIMIC-III} (AUROC/AUPRC).}
\label{tab:plaus_los_m3}
\centering
\small
\begin{tabular}{l ccccc}
\toprule
Method & Discrete & Disc.+PT & HyMaTE & BAT & STraTS \\
\midrule
\tisae{} (soft) & \textbf{.699}/.360 & .514/.197 & \underline{.623}/.295 & \textbf{.633}/.279 & .559/.242 \\
\tisae{} (hard) & \underline{.659}/.335 & .508/.193 & .614/.275 & \underline{.612}/.266 & .555/.241 \\
Attention & .617/.282 & .453/.184 & .560/.243 & -- & \underline{.637}/.260 \\
IG & .543/.222 & .543/.234 & .587/.257 & .591/.243 & .593/.239 \\
TSR & .569/.242 & \underline{.581}/.248 & \textbf{.626}/.290 & .611/.264 & .629/.261 \\
DynaMask & .510/.203 & .505/.204 & .492/.189 & .493/.189 & .492/.189 \\
TimeSHAP & .513/.203 & .521/.216 & .517/.210 & .517/.209 & .517/.210 \\
FIT & .510/.199 & .538/.214 & .528/.209 & .504/.195 & .556/.221 \\
WinIT & .575/.242 & \textbf{.612}/.279 & .602/.265 & .562/.253 & \textbf{.640}/.310 \\
Cluster-TCAV & .463/.196 & .453/.195 & .423/.198 & .556/.225 & .421/.199 \\
Random & .502/.193 & .502/.193 & .502/.193 & .502/.193 & .502/.193 \\
\bottomrule
\end{tabular}
\end{table}

\begin{table}[h!]
\caption{Per-task plausibility: \textbf{LOS>7d, MIMIC-IV} (AUROC/AUPRC).}
\label{tab:plaus_los_m4}
\centering
\small
\begin{tabular}{l ccccc}
\toprule
Method & Discrete & Disc.+PT & HyMaTE & BAT & STraTS \\
\midrule
\tisae{} (soft) & .312/.181 & \textbf{.629}/.263 & .588/.255 & .525/.219 & .601/.244 \\
\tisae{} (hard) & .417/.217 & .572/.237 & .541/.244 & .499/.217 & \underline{.606}/.249 \\
Attention & .523/.220 & \underline{.619}/.260 & .577/.236 & -- & .477/.181 \\
IG & .516/.214 & .535/.232 & \underline{.606}/.263 & \underline{.564}/.239 & .552/.220 \\
TSR & \underline{.544}/.230 & .565/.249 & \textbf{.644}/.298 & \textbf{.587}/.259 & .580/.237 \\
DynaMask & .500/.198 & .500/.199 & .495/.188 & .474/.186 & .494/.188 \\
TimeSHAP & .512/.203 & .521/.212 & .525/.217 & .517/.212 & .514/.204 \\
FIT & .519/.210 & .542/.220 & .521/.193 & .466/.180 & .551/.222 \\
WinIT & \textbf{.567}/.241 & .618/.296 & .595/.252 & .563/.247 & \textbf{.610}/.272 \\
Cluster-TCAV & .420/.171 & .341/.153 & .348/.162 & .382/.165 & .465/.188 \\
Random & .501/.191 & .501/.191 & .501/.191 & .501/.191 & .501/.191 \\
\bottomrule
\end{tabular}
\end{table}

\begin{table}[h!]
\caption{Per-task plausibility: \textbf{Readmission, MIMIC-III} (AUROC/AUPRC).}
\label{tab:plaus_readm_m3}
\centering
\small
\begin{tabular}{l ccccc}
\toprule
Method & Discrete & Disc.+PT & HyMaTE & BAT & STraTS \\
\midrule
\tisae{} (soft) & .533/.213 & \textbf{.608}/.261 & .593/.261 & .577/.241 & .522/.195 \\
\tisae{} (hard) & .534/.211 & .596/.251 & .616/.262 & .558/.232 & .515/.194 \\
Attention & .464/.175 & .463/.175 & .573/.203 & -- & \textbf{.627}/.249 \\
IG & .553/.220 & .574/.223 & .593/.227 & \underline{.590}/.240 & \underline{.605}/.219 \\
TSR & \textbf{.573}/.245 & .584/.228 & \underline{.618}/.249 & \textbf{.615}/.258 & .603/.218 \\
DynaMask & .506/.200 & .515/.208 & .494/.183 & .495/.184 & .497/.192 \\
TimeSHAP & .513/.212 & .512/.194 & .510/.197 & .515/.197 & .505/.189 \\
FIT & .493/.180 & .532/.201 & .553/.204 & .525/.209 & .535/.197 \\
WinIT & .565/.261 & \underline{.602}/.267 & \textbf{.645}/.306 & .557/.228 & .599/.259 \\
Cluster-TCAV & \underline{.567}/.232 & .498/.213 & .617/.391 & .534/.215 & .536/.240 \\
Random & .503/.188 & .503/.188 & .503/.188 & .503/.188 & .503/.188 \\
\bottomrule
\end{tabular}
\end{table}

\begin{table}[h!]
\caption{Per-task plausibility: \textbf{Readmission, MIMIC-IV} (AUROC/AUPRC).}
\label{tab:plaus_readm_m4}
\centering
\small
\begin{tabular}{l ccccc}
\toprule
Method & Discrete & Disc.+PT & HyMaTE & BAT & STraTS \\
\midrule
\tisae{} (soft) & \underline{.556}/.215 & .581/.247 & .558/.234 & .559/.220 & .471/.215 \\
\tisae{} (hard) & .534/.210 & .572/.240 & .534/.228 & .537/.224 & .464/.214 \\
Attention & .532/.200 & \textbf{.606}/.239 & .524/.189 & -- & .462/.171 \\
IG & .527/.210 & .553/.219 & .552/.191 & \underline{.564}/.215 & .544/.193 \\
TSR & .542/.219 & .571/.229 & \underline{.569}/.199 & \textbf{.592}/.235 & \underline{.554}/.197 \\
DynaMask & .530/.195 & .503/.191 & .495/.182 & .488/.182 & .521/.201 \\
TimeSHAP & .498/.180 & .512/.196 & .501/.184 & .500/.183 & .505/.187 \\
FIT & .499/.179 & .530/.193 & .503/.172 & .488/.176 & .536/.192 \\
WinIT & \textbf{.563}/.230 & \underline{.603}/.279 & \textbf{.589}/.225 & .552/.233 & \textbf{.610}/.258 \\
Cluster-TCAV & .553/.244 & .490/.226 & .439/.191 & .470/.195 & .530/.202 \\
Random & .502/.183 & .502/.183 & .502/.183 & .502/.183 & .502/.183 \\
\bottomrule
\end{tabular}
\end{table}

\clearpage
\section{Clinical plausibility ground truth generation}
\label{app:plausibility}

We construct soft ground-truth severity maps from established ICU scoring systems to evaluate whether attributions align with clinically meaningful observations.
The ground truth is a per-cell score $\mathbf{S} \in [0, 1]^{B \times T \times V}$ defined only at observed cells, representing clinical severity.

\paragraph{SOFA scoring.}
We implement the Sequential Organ Failure Assessment
(SOFA)~\citep{vincent1996sofa} scoring for six organ systems:

\begin{itemize}
  \item \textbf{Respiratory}: PaO$_2$/FiO$_2$ ratio. Thresholds $< 400, 300, 200, 100$ yield SOFA 1--4. Requires co-observation of PO$_2$ and FiO$_2$ within 6 time bins.
  \item \textbf{CNS}: Glasgow Coma Scale total (eye + verbal + motor). GCS $\leq 5, 9, 12, 14$ yield SOFA 4, 3, 2, 1.
  \item \textbf{Cardiovascular}: MAP $< 70$ yields SOFA 1; vasopressor presence and dose (Dopamine, Norepinephrine, Epinephrine) yield SOFA 2--4.
  \item \textbf{Coagulation}: Platelet count $< 150, 100, 50, 20 \times 10^3$ yields SOFA 1--4.
  \item \textbf{Liver}: Bilirubin $> 1.2, 2.0, 6.0, 12.0$ mg/dL yields SOFA 1--4.
  \item \textbf{Renal}: Creatinine $> 1.2, 2.0, 3.5, 5.0$ mg/dL yields SOFA 1--4; urine output $< 500, 200$ mL/day yields SOFA 3--4.
\end{itemize}

Each cell's score is the organ sub-score divided by 4 (normalizing to $[0, 1]$). For multi-variable organs (respiratory, CNS, cardiovascular), all contributing variables at a given time bin receive the same organ score.

\paragraph{Metrics.}
Given an attribution map $\mathbf{A} \in \reals^{B \times T \times V}$
and severity map $\mathbf{S}$, both restricted to observed cells, we compute:
\begin{itemize}
  \item \textbf{AUROC}: ROC-AUC for classifying $S > 0$ (any severity) vs.\ $S = 0$.
  \item \textbf{AUPRC}: Precision-recall AUC for the same binary classification.
\end{itemize}
Higher scores indicate that the attribution method assigns higher importance to clinically severe cells.

This ground truth is a minimal surrogate for clinical relevance.
Actual clinical concepts are more complex, involving entangled multi-variable interactions and temporal dynamics that static threshold-based scoring cannot
capture.
The SOFA score evaluates organ systems independently and does not reflect inter-system dependencies or context-dependent severity.
We use this metric as a necessary but not sufficient condition for clinical plausibility.

\section{Ablation study}
\label{app:ablations}

\subsection{Layer selection}
\label{app:layer_selection}

\tisae{} operates on a single backbone layer's hidden states.
The choice of layer involves a tradeoff: early layers preserve per-variable locality but contain less contextual information; deeper layers form richer concepts but may lose variable distinctiveness.

Table~\ref{tab:layer_ablation} reports AOPC (neighborhood) across layers for each backbone.
For the discrete transformer, layer $-1$ (last) outperforms other layers across all tasks.
For Disc.+PT, earlier layers outperform, suggesting that pretraining shifts local representations to earlier depths. We select layer $-3$, however, as layer $-4$ plausibility degrades to near-random on several tasks.
For HyMaTE, layer $-2$ is optimal.
For STraTS, all layers yield low \tisae{} AOPC. We attribute this to STraTS's residual-averaging architecture ($\mathbf{h} \leftarrow (\mathbf{x} + \mathbf{h})/2$) they used instead of layernorm for simplification, which causes representations converge toward uniformity at all depths, eliminating the per-variable per-time distinctiveness that \tisae{}'s routing requires.
For BAT, layer $-2$ is optimal, suggesting that round-1 representations (before cross-axis mixing) better preserve separability.

\begin{table}[h!]
\caption{Layer selection: AOPC (neighborhood) by backbone layer,
MIMIC-III mortality. Best per backbone in \textbf{bold}.}
\label{tab:layer_ablation}
\centering
\small
\begin{tabular}{lcccc}
\toprule
Backbone & Layer $-1$ & Layer $-2$ & Layer $-3$ & Layer $-4$ \\
\midrule
Discrete & \textbf{.149} & .121 & .114 & .105 \\
Discrete+PT & .029 & .037 & .091 & \textbf{.216} \\
HyMaTE & .109 & \textbf{.143} & .121 & -- \\
STraTS & .002 & .003 & .003 & \textbf{.015} \\
BAT & .061 & \textbf{.065} & -- & -- \\
\bottomrule
\end{tabular}
\end{table}

\subsection{Selection method}

We compare island selection strategies at evaluation time.
``SAE (energy)'' uses the energy-gated score from Stage~A (Eq.~\ref{eq:score}).
``Grad$\times$Act'' uses gradient-times-activation scoring as a non-learned heuristic prior.
``\predselector{} (soft)'' uses the learned prediction-aware selector from Stage~B.

\begin{table}[h!]
\caption{Selection method: AOPC (neighborhood), averaged across tasks.}
\label{tab:selection_ablation}
\centering
\small
\begin{tabular}{lccccc}
\toprule
& Discrete & Disc.+PT & HyMaTE & BAT & STraTS \\
\midrule
Activation (energy) & .037 & .008 & .028 & .006 & .008 \\
Gradient$\times$Activation & .035 & .027 & .032 & .022 & .014 \\
Prediction selector (hard) & .079 & .068 & .090 & .035 & .020 \\
Prediction selector (soft) & .086 & .076 & .097 & .043 & .016 \\
\bottomrule
\end{tabular}
\end{table}

\subsection{Structural mechanism ablation}
\label{app:structural_ablation}

We ablate the structural mechanisms of \tisae{} by progressively removing components: temporal Gaussian islands, variable routing ($W_k{=}3 \to 135$), channel purity regularization ($\lambda_\text{ch}{=}0$), and grid structure entirely (standard event-level top-$k$ SAE on $\mathbf{H}[\![B,L,D]\!]$). Additionally, to isolate the dictionary's contribution from the prediction-aware selector, we train the selector directly on raw hidden states with no dictionary. All conditions use the Discrete backbone on MIMIC-III mortality.

\paragraph{Quantitative results.}
Table~\ref{tab:structural_ablation} reports faithfulness (neighborhood-based AOPC and sufficiency at 10\%) and plausibility (AUROC and AUPRC against SOFA-derived ground truth). Removing temporal structure or channel purity has minimal impact on faithfulness. Removing channel purity slightly improves AOPC, likely because the model gains more freedom in variable grouping. In contrast, removing variable routing causes a 58--75\% AOPC drop, showing that decoder sparsity is the critical mechanism for faithful attribution. 
The standard event-level top-$k$ SAE produces near-random attributions (AOPC .023) based on activation.

\begin{table}[h]
\centering
\caption{Structural mechanism ablation (Discrete backbone, MIMIC-III mortality). AOPC = comprehensiveness (fractions 0.05--0.5); Suff = sufficiency at 10\%; AUROC/AUPRC = plausibility against SOFA prior. All metrics use PS-soft selection. For Standard top-$k$ SAE, activation(dictionary)-based AOPC is shown.}
\label{tab:structural_ablation}
\small
\begin{tabular}{l cc cc c}
\toprule
& \multicolumn{2}{c}{Faithfulness} & \multicolumn{2}{c}{Plausibility} & \\
\cmidrule(lr){2-3} \cmidrule(lr){4-5}
Condition & AOPC$\uparrow$ & Suff@.1$\downarrow$ & AUROC$\uparrow$ & AUPRC$\uparrow$ \\
\midrule
Full \tisae{} & .149 & $-$.004 & .756          & .386 \\
$-$ Temporal & .145 & .006 & .777 & .418 \\
$-$ Channel org & .184 & .003 & .734 & .358 \\
$-$ Var routing & .061 & .052 & .629 & .280 \\
Standard SAE & .023 (.155) & .126 (.041) & .487 (.568) & .189 (.224) \\
\bottomrule
\end{tabular}
\end{table}

However, temporal structure, channel purity, and variable routings are all important components for generating a meaningful clinical dictionary. We examine the interpretability of the dictionary by \emph{channel coherence} and \emph{temporal localization}.

\paragraph{Channel coherence.}
Without channel purity ($\lambda_\text{ch}{=}0$), latents freely mix variables across organ systems. In the full dictionary, only 36.9\% of latents have at least two decoder variables within a single clinical channel (compared to 91.9\% in the default model), directly degrading the clinical interpretability of individual concepts.

Table~\ref{tab:channel_example} illustrates this on a representative mortality episode. \tisae{} discovers organ-specific concepts (e.g., a pure Renal latent: BUN + Creatinine + Creatinine Urine, labeled ``Chronic Kidney Disease''). Without channel purity, the corresponding latents mix variables from unrelated systems, making them difficult to label or act upon clinically.

\begin{table}[h]
\centering
\caption{Top-5 prediction-relevant latents for episode 23020 (mortality): default vs.\ no channel purity. Channel annotations in brackets.}
\label{tab:channel_example}
\small
\begin{tabular}{c l l}
\toprule
\multicolumn{3}{l}{\textbf{Default}} \\
\midrule
k=113  & Temperature, CRP, Eosinophils [all Infection] & \textit{Hypothermia w/ Inflammatory Markers} \\
k=1246 & TPN [Other], RDW [Heme], Gastric [Other] & \textit{TPN-Dependent Feeding Intolerance} \\
k=1119 & GCS\_verbal, GCS\_eye [Neuro], PO2 [Resp] & \textit{Coma with Mechanical Ventilation} \\
k=931  & BUN, Creatinine, Creatinine Urine [all Renal] & \textit{Chronic Kidney Disease with Azotemia} \\
k=886  & BUN, Creatinine, Urine [all Renal] & \textit{Acute Kidney Injury with Oliguria} \\
\midrule
\multicolumn{3}{l}{\textbf{No channel organization}} \\
\midrule
k=136  & Platelets, Fibrinogen [Coag], \textbf{Bilirubin [Liver]} & --- \\
k=789  & Albumin [\textbf{Liver}], Saline [\textbf{Other}], MgSO4 [\textbf{Metabolic}] & --- \\
k=1270 & PCO2 [\textbf{Resp}], MCH [\textbf{Heme}], Platelets [\textbf{Coag}] & --- \\
k=1119 & GCS\_verbal, GCS\_eye [Neuro], \textbf{Cr Urine [Renal]} & --- \\
k=817  & BUN [\textbf{Renal}], GCS\_eye [\textbf{Neuro}], Insulin [\textbf{Metabolic}] & --- \\
\bottomrule
\end{tabular}
\end{table}

\paragraph{Temporal localization.}
Without Gaussian islands, the encoder still produces per-time-step activations $\hat{A}[k,t]$, but it does not learn temporally concentrated patterns. Latents activate nearly uniformly across all 48 time bins (temporal entropy 0.957 vs.\ default version's localized islands). While this does not substantially hurt faithfulness metrics (the model still identifies which variables matter), it eliminates the ability to say \emph{when} a clinical concept is active. In the same episode, the default model distinguishes early renal decline (k=931 at hour~20) from late-stage injury (k=886 at hour~33) and late coma onset (k=1119 at hour~46). The no-temporal model collapses this temporal narrative.

\paragraph{Variable routing.}
Removing decoder sparsity degrades both faithfulness and interpretability. Each latent activates 22--37 variables, making individual concepts unidentifiable.

\paragraph{Standard SAE.}
We attempted to label latents by showing an LLM the full clinical records of each latent's top-activating episodes. For the same mortality episode (23020) where \tisae{} identifies five distinct organ-system concepts (Table~\ref{tab:channel_example}), the standard SAE's top 10 latents all fire on 100\% of patients and receive contradictory labels---the same patient is simultaneously described as having ``stable vital signs,'' ``tachycardia
with hemodynamic instability,'' and ``hypotension requiring vasopressors.'' These labels seem to describe the common phenotypes of different top-activating patients, not the latent's mechanism. Unlike \tisae{}, where the decoder structurally binds each latent to specific variables and time, the standard SAE's latents have no such anchor, making labels unfalsifiable.
We also trained a modified \predselector{} that scores latents globally, recovering neighborhood-based AOPC.
However, this required injecting variable identity information that the SAE itself does not encode (using per-variable activation profiles computed from batch data), and weights each latent uniformly across all temporal positions. The low plausibility metrics imply that the attribution does not reflect genuine clinical relevance. Still, the small change in AOPC suggests that the specific \tisae{} architecture is not required for the faithfulness gain; the islands instead primarily improve concept plausibility and interpretability, which is the structural information needed to read the dictionary.

\paragraph{Isolating the dictionary from the selector.}
The dictionary plays an independent role in turning a sufficient set of cells into a decomposable, spatiotemporally grounded concept shared across patients, which the \predselector{} alone cannot produce.
Still, it is important to decompose where the performance gain comes from. 
We tried to isolate the \predselector{} from the dictionary, using the selector trained directly on raw hidden states with no dictionary, under the same selection objective and budget (cell-level selection). 
\tisae{} substantially improves AOPC over raw hidden states (.149 vs.\ .038), consistent with a representation-level contribution beyond prediction-aware selection alone. 

\subsection{Island window shape}
\label{app:island_shape}
A Gaussian window in this study is a temporal support gate that indicates when a concept is active, while the actual temporal trajectory (the activation strength) is carried by the encoder output.
Nevertheless, here we test whether it restricts the temporal patterns a concept can express, by replacing the symmetric Gaussian with a split-normal (skewed) window ($\sigma_L = \sigma(1{-}a)$, $\sigma_R = \sigma(1{+}a)$), with the skew $a$ either learned per island or fixed, and re-running the full pipeline over seeds 42--44 (Table~\ref{tab:island_shape}).

\begin{table}[h]
\centering
\caption{Island window shape ablation (Discrete backbone, MIMIC-III mortality; mean $\pm$ std over three seeds). AOPC = neighborhood-based comprehensiveness; plausibility = SOFA AUROC.}
\label{tab:island_shape}
\small
\begin{tabular}{l cc}
\toprule
Island shape & AOPC$\uparrow$ & Plausibility$\uparrow$ \\
\midrule
Symmetric Gaussian & $.140 \pm .006$ & $.739 \pm .013$ \\
Learned asymmetric & $.142 \pm .014$ & $.725 \pm .013$ \\
Fixed asymmetric ($a{=}0.5$) & $.141 \pm .016$ & $.715 \pm .031$ \\
\bottomrule
\end{tabular}
\end{table}

When the skew is free the model uses it heavily (mean $|a| \approx 0.34$, with 85--96\% of islands skewed), so asymmetry is available and actively used, yet performance is unchanged.
This supports that the symmetric window is not a restrictive choice for these tasks.

Also, the windows are already diverse at the concept level. A concept's mask is the clamped sum of its $M$ windows (Eq.~\ref{eq:mask}), and on the canonical model most concepts use more than one: 30/21/23/26\% of selected concepts use 1/2/3/4 windows (70\% use two or more). Therefore, most masks are composed from several windows rather than a single symmetric bump.

\subsection{Dictionary capacity and sparsity budget}
\label{app:capacity_sparsity}

We sweep the dictionary size $K$ with fixed selection budget $n_\text{global}{=}128$, sweep $n_\text{global}$ with fixed $K{=}1280$, and sweep the number of islands per latent $M$ with proportional budget $n_\text{global}{=}32M$. All conditions use the Discrete backbone on MIMIC-III mortality.

\begin{table}[h]
\centering
\caption{Dictionary capacity sweep ($n_\text{global}{=}128$ fixed). Dead\% = fraction of latents never selected across 28,626 episodes; $>$1\% = latents active in $>$1\% of episodes.}
\label{tab:k_sweep}
\small
\begin{tabular}{r cc cc cc c}
\toprule
& \multicolumn{2}{c}{Utilization} & \multicolumn{2}{c}{Faithfulness} & \multicolumn{2}{c}{Plausibility} & \\
\cmidrule(lr){2-3} \cmidrule(lr){4-5} \cmidrule(lr){6-7}
$K$ & Dead\% & $>$1\% & AOPC$\uparrow$ & Suff@.1$\downarrow$ & AUROC$\uparrow$ & AUPRC$\uparrow$ \\
\midrule
320  & 0.3\%  & 281 & .137 & .030 & .711 & .353 \\
640  & 0.0\%  & 417 & .165 & $-$.005 & .734 & .375 \\
1280 & 2.0\%  & 469 & .149 & $-$.004 & .756 & .386 \\
2560 & 23.5\% & 524 & .143 & $-$.005 & .700 & .333 \\
\bottomrule
\end{tabular}
\end{table}

\begin{table}[h]
\centering
\caption{Sparsity budget sweep ($K{=}1280$ fixed). $n/KM$ = fraction of island slots selected per episode.}
\label{tab:ng_sweep}
\small
\begin{tabular}{rr cc cc cc c}
\toprule
& & \multicolumn{2}{c}{Utilization} & \multicolumn{2}{c}{Faithfulness} & \multicolumn{2}{c}{Plausibility} & \\
\cmidrule(lr){3-4} \cmidrule(lr){5-6} \cmidrule(lr){7-8}
$n_\text{global}$ & $n/KM$ & Dead\% & $>$1\% & AOPC$\uparrow$ & Suff@.1$\downarrow$ & AUROC$\uparrow$ & AUPRC$\uparrow$ & \\
\midrule
32  & 0.6\%  & 26.2\% & 136  & .134 & .015 & .713 & .333 \\
64  & 1.2\%  & 8.2\%  & 259  & .151 & $-$.007 & .716 & .353 \\
128 & 2.5\%  & 2.0\%  & 469  & .149 & $-$.004 & .756 & .386 \\
256 & 5.0\%  & 0.1\%  & 767  & .141 & .000 & .714 & .364 \\
512 & 10.0\% & 0.2\%  & 1022 & .128 & .019 & .763 & .395 \\
\bottomrule
\end{tabular}
\end{table}

\begin{table}[h!]
\centering
\caption{Island count sweep ($K{=}1280$, $n_\text{global}{=}32M$ proportional).}
\label{tab:m_sweep}
\small
\begin{tabular}{rr cc cc cc}
\toprule
& \multicolumn{2}{c}{Faithfulness} & \multicolumn{2}{c}{Plausibility} \\
\cmidrule(lr){2-3} \cmidrule(lr){4-5}
$M$ & AOPC$\uparrow$ & Suff@.1$\downarrow$ & AUROC$\uparrow$ & AUPRC$\uparrow$ \\
\midrule
1  & .131 & $-$.021 & .723 & .346 \\
2  & .152 & $-$.004 & .731 & .341 \\
4  & .149 & $-$.004 & .756 & .386 \\
8  & .144 & .005 & .703 & .357 \\
\bottomrule
\end{tabular}
\end{table}

Faithfulness peaks at moderate capacity: AOPC is highest at $K{=}640$ and $n_\text{global}{=}64$, then declines as larger dictionaries
spread attribution across more latents. Plausibility improves with capacity up to $K{=}1280$ but drops at $K{=}2560$, and increases monotonically with $n_\text{global}$ as more active latents cover more clinically relevant cells.
Increasing $M$ from 1 to 4 improves faithfulness (AOPC .131$\to$.149) by enabling temporal decomposition within latents, but performance declines at $M{=}8$, suggesting overly fine fragmentation is counterproductive.
We select $K{=}1280$, $n_\text{global}{=}128$, and $M{=}4$ as the operating point where faithfulness and plausibility remain competitive and latent utilization is efficient.

\subsection{Loss component analysis}
\label{app:loss_components}

We ablate each regularization loss by setting its weight to zero (Discrete backbone, MIMIC-III mortality). Without auxK revival
($\lambda_\text{auxk}{=}0$), 80.0\% of latents are never selected across the full cohort (vs.\ 2.0\% with auxK), and temporal spread within latents collapses to 1.80h (vs.\ 5.46h). Without $\mu$-repulsion ($\lambda_\text{repul}{=}0$), islands within each latent overlap temporally, and the model compensates by widening them ($\sigma_\text{alive}$: 7.80 vs.\ 4.92). Removing mask coverage ($\lambda_\text{mcov}{=}0$) has minimal effect at the default operating point ($\sigma_\text{sel}$: 4.97 vs.\ 5.25); residualization and $\mu$-repulsion already constrain island width, and we retain mask coverage at a small $\lambda{=}0.005$ as a safeguard for higher-budget regimes.

\subsection{STraTS residual architecture}
\label{app:strats_residual}

STraTS replaces the standard additive residual with an averaging operation $\mathbf{h} \leftarrow (\mathbf{x} + \mathbf{h})/2$.
Each averaging step blends the cell representation toward a shared mean, reducing per-$(t,v)$ distinctiveness.
We hypothesize that this undermines \tisae{}'s structure, which requires distinct per-cell features to assign observations to latent channels.
To isolate this effect, we replace only the residual with $\mathbf{h} \leftarrow \text{LayerNorm}(\mathbf{x} + \mathbf{h})$,
keeping all other hyperparameters identical.

Table~\ref{tab:strats_residual} shows the comparison before and after the residual change. On MIMIC-III mortality (layer~$-4$, neighborhood-based masking), this change recovers \tisae{} to competitive faithfulness (PS-soft AOPC .015 $\to$ .044) and best plausibility (AUROC .516 $\to$ .652), showing that the averaging residual is the primary bottleneck for \tisae{} on this backbone.

\begin{table}[h]
\centering
\caption{STraTS residual architecture ablation (MIMIC-III mortality, layer $-4$, neighborhood-based masking).}
\label{tab:strats_residual}
\small
\begin{tabular}{l cc cc cc}
\toprule
& \multicolumn{2}{c}{AOPC $\uparrow$} & \multicolumn{2}{c}{AUROC $\uparrow$} & \multicolumn{2}{c}{AUPRC $\uparrow$} \\
\cmidrule(lr){2-3} \cmidrule(lr){4-5} \cmidrule(lr){6-7}
Method & Orig & LN & Orig & LN & Orig & LN \\
\midrule
\tisae{} (soft) & .015 & .044 & .516 & .652 & .216 & .296 \\
\tisae{} (hard) & .013 & .042 & .506 & .637 & .218 & .285 \\
Attention       & .010 & .046 & .611 & .542 & .250 & .220 \\
IG              & .064 & .048 & .622 & .608 & .252 & .244 \\
TSR             & .061 & .050 & .637 & .625 & .265 & .260 \\
DynaMask        & .033 & .033 & .492 & .492 & .189 & .189 \\
TimeSHAP        & .016 & .016 & .524 & .522 & .214 & .214 \\
FIT             & .010 & .011 & .564 & .547 & .227 & .218 \\
WinIT           & .025 & .026 & .644 & .619 & .299 & .278 \\
\bottomrule
\end{tabular}
\end{table}

\section{Backbone prediction performance}
\label{app:backbone_perf}

Table~\ref{tab:backbone_auroc} reports test AUROC for each backbone across prediction tasks and datasets.
All backbones achieve comparable performance on mortality
(.855--.870) and length-of-stay (.819--.843).
Readmission is consistently harder (.618--.652), which is typical for this task.

\begin{table}[h]
\caption{Backbone test AUROC per task and dataset.}
\label{tab:backbone_auroc}
\centering
\small
\begin{tabular}{lcccccc}
\toprule
& \multicolumn{2}{c}{Mortality} & \multicolumn{2}{c}{LOS $>$7d} & \multicolumn{2}{c}{Readmission} \\
\cmidrule(lr){2-3} \cmidrule(lr){4-5} \cmidrule(lr){6-7}
Backbone & III & IV & III & IV & III & IV \\
\midrule
Discrete & .861 & .867 & .832 & .824 & .632 & .652 \\
Discrete+PT & .861 & .865 & .834 & .827 & .618 & .651 \\
HyMaTE & .861 & .872 & .830 & .829 & .638 & .649 \\
STraTS & .873 & .867 & .843 & .823 & .644 & .649 \\
BAT & .858 & .868 & .826 & .822 & .612 & .648 \\
\bottomrule
\end{tabular}
\end{table}

\section{Latent dictionary labeling}
\label{app:dictionary_labeling}

Each active latent $k$ with activation frequency ${>}1\%$ in the learned dictionary is labeled by an LLM (Claude Sonnet 4.6) to produce a clinical concept name and description.
The system message instructs the model to act as an expert clinical informaticist interpreting \tisae{} latents from MIMIC-III, responding only with a JSON object.

The per-latent prompt contains five sections:
\begin{enumerate}
    \item \textbf{Context.} Describes the \tisae{} setup (48-hour irregularly-sampled ICU time-series, mortality prediction, latent concepts with Gaussian islands and channel assignments) and a note that zero observations within a window reflect clinical ordering patterns, not missing data.
    \item \textbf{Variable composition.} Top-3 decoder variables with weights from $\mathbf{W}[k,:]$.
    \item \textbf{Activation statistics.} Population activation frequency and mortality odds ratio.
    \item \textbf{Exemplar episodes.} Up to 10 episodes ranked by SAE activation strength, plus up to 10 ranked by \predselector{} weight. Each shows island windows ($\mu$, $\sigma$), raw variable traces, and unobserved variables. Example:
    \begin{quote}
    \small
    \texttt{Episode 4821 (mortality: YES, strength: 0.847)}\\
    \texttt{Island: hours 18--30 (mu=24.2, sigma=3.1)}\\
    \texttt{- Creatinine: t20h=1.82, t24h=2.14, t28h=2.61}\\
    \texttt{- BUN: t18h=34.0, t24h=42.0}\\
    \texttt{- Urine Output: NOT MEASURED in this window}
    \end{quote}
    \item \textbf{Task.} Requests JSON with \texttt{clinical\_label} (2--6 word name) and \texttt{clinical\_description} (1--3 sentence interpretation).
\end{enumerate}

\paragraph{LLM implementation details.}
Dictionary labeling uses Claude Sonnet 4.6 (Anthropic API, temperature 0.0).
Verbalization uses the same model. Pairwise judging uses GPT-4o (OpenAI API, temperature 0.0) as the judge model, to avoid self-preference bias.

\section{Verbalized explanation evaluation}
\label{app:verbalization}

\subsection{Verbalization and explanation quality evaluation protocol}
\label{app:verbalization_protocol}
The verbalization evaluation (Section~\ref{sec:layer3}) uses two LLM stages: a verbalizer (Claude Sonnet 4.6) that generates explanations, and a pairwise judge (GPT-4o) that compares them per dimension.

\paragraph{Verbalizer prompt.}
Both conditions share the same system instruction:
\begin{quote}
\small
\emph{You are an ICU physician writing a clinical explanation of a mortality prediction. In under 150 words, explain why this patient is at this risk level. Explain exclusively based on the model-provided analysis below. Plain text, no markdown.}
\end{quote}

\paragraph{Explanation conditions.}
Both conditions receive the prediction (risk level and probability) and the full 48-hour raw value table. 
They differ in what model-derived information is provided.

\smallskip\noindent\textbf{Attribution-augmented.} Top-$K$ cells ranked by attribution score:

\begin{quote}
\small\ttfamily
Patient prediction: HIGH mortality risk (probability: 0.82)\\[2pt]
The model's attribution analysis identified these observations\\
as most important for its prediction:\\
~~Hour 18: Creatinine (importance: 0.1842, value: 2.14)\\
~~Hour 24: Heart Rate (importance: 0.1203, value: 112)\\
~~...\\[2pt]
Clinical values over the 48-hour ICU stay:\\
~~[full raw value table]
\end{quote}

\smallskip\noindent\textbf{\tisae{}-augmented.} Top-5 \predselector{}-ranked concepts with dictionary labels
(Appendix~\ref{app:dictionary_labeling}):

\begin{quote}
\small\ttfamily
Patient prediction: HIGH mortality risk (probability: 0.82)\\[2pt]
The model identified these clinical concepts as most important\\
for its prediction:\\
~~Concept 1 [Renal] "Progressive AKI" (total weight=3.5):\\
~~~~Creatinine (wt=0.482), BUN (wt=0.318)\\
~~~~Rising creatinine and BUN suggesting acute kidney injury.\\
~~~~Island h18-30 (wt=2.10):\\
~~~~~~Creatinine: h20=1.82, h24=2.14, h28=2.61\\
~~~~~~BUN: h18=34.0, h24=42.0\\
~~...\\[2pt]
Clinical values over the 48-hour ICU stay:\\
~~[full raw value table]
\end{quote}

\paragraph{Pairwise judge protocol.}
For each episode and baseline, the judge compares the \tisae{} explanation against the baseline explanation. A/B order is randomized per pair. Each of the three dimensions is evaluated in an independent API call to avoid holistic bias~\citep{zheng2023judging}.

Concept-level reasoning and Clinical actionability dimensions are \emph{blinded}: the judge sees only the patient data table and the two explanations, with no indication of which method produced which. Faithfulness dimension is \emph{unblinded}.

\begin{enumerate}
  \item \textbf{Concept-level reasoning}: Which explanation better identifies clinical concepts (co-evolving variable groups) rather than listing individual abnormalities?
  \item \textbf{Clinical actionability}: Which explanation would be more useful for a clinician deciding on next steps for this patient?
  \item \textbf{Faithfulness}: The judge additionally sees each explanation's source model analysis (attribution cells or concept structure) and evaluates which explanation more faithfully represents the findings from its source, accurately conveying the key patterns, priorities, and values identified by the model.
\end{enumerate}

\paragraph{Evaluation protocol.}
$N{=}50$ test episodes selected via stratified random sampling (25~high-risk with $p > 0.5$, 25~low-risk).

\emph{Budget matching.}
$K$ for the attribution condition equals the mean number of distinct observed $(t,v)$ cells covered by \tisae{}'s top-5 concepts ($K{=}75$),
ensuring comparable information budgets.

\subsection{Clinician evaluation of verbalized explanations}
\label{app:clinician_pairwise_eval}

To validate the LLM judge alignment, four independent tertiary-hospital clinicians (all board-certified specialists with $\geq$10 years of experience; the same panel as the dictionary label evaluation in Appendix~\ref{app:clinician_eval}) re-ran
the pairwise comparison on the same 50 episodes used in the LLM judge evaluation, restricted to the IG baseline (for completely blinded evaluation). Each clinician saw two explanations (A and B, randomized order) and scored three dimensions: concept-level reasoning, actionability, and faithfulness. Phase~1 (concept, actionability) was blinded to the source method; faithfulness was scored after locking Phase~1, with source analyses revealed in a separate window. Each rater could optionally view raw patient data and
per-variable trend charts.

\paragraph{Aggregate marginal alignment.}
Aggregate clinician majority votes matched the LLM judge's ranking direction across all three dimensions (Table~\ref{tab:clinician_pairwise_alignment}).

\begin{table}[h]
\centering
\caption{Marginal preference distributions ($N{=}50$ episode pairs, IG baseline).
Format: TI-SAE preferred / IG preferred / Tie.}
\label{tab:clinician_pairwise_alignment}
\small
\begin{tabular}{l ccc}
\toprule
& Concept & Actionability & Faithfulness \\
\midrule
Clinician majority & 39 / 4 / 7  & 23 / 9 / 18 & 4 / 4 / 42 \\
LLM judge & 33 / 15 / 2  & 32 / 18 / 0  & 15 / 11 / 24 \\
\bottomrule
\end{tabular}
\end{table}

\paragraph{Per-rater marginals.}
Individual raters varied substantially in tie usage (Table~\ref{tab:clinician_pairwise_perrater}),
particularly on faithfulness.

\begin{table}[h]
\centering
\caption{Per-rater preference marginals ($N{=}50$ each).}
\label{tab:clinician_pairwise_perrater}
\small
\begin{tabular}{l ccc}
\toprule
Rater & Concept & Actionability & Faithfulness \\
\midrule
Rater 1 & 38 / 8 / 4  & 24 / 5 / 21  & 3 / 2 / 45  \\
Rater 2 & 30 / 20 / 0 & 22 / 22 / 6  & 21 / 29 / 0 \\
Rater 3 & 30 / 3 / 17 & 23 / 15 / 12 & 14 / 13 / 23 \\
Rater 4 & 36 / 9 / 5 & 25 / 10 / 15 & 6 / 3 / 41 \\
\bottomrule
\end{tabular}
\end{table}

\paragraph{Per-item agreement.}
Per-item inter-rater agreement was low across all rater pairs:
clinician-clinician Fleiss $\kappa$ was $+0.04$ for concept, $+0.18$ for actionability, and $-0.06$ for faithfulness; clinician-LLM agreement was in a simliar range (mean Cohen's $\kappa = -0.03$, $-0.01$, and $+0.08$ across raters for concept, actionability, and faithfulness, respectively). 
We attribute this to the inherent hardness in our setting, compounded by the verbalizer LLM's strong clinical prior, and the difficulty of consistently judging free-form text on multiple implicit quality criteria simultaneously~\citep{saadfalcon2025lmunit}. Small differences in explanation quality produce many borderline cases, so per-item agreement becomes dominated by rater-style differences such as individual tie-threshold preferences. In our data this manifested as substantial across-rater variation in tie usage (one rater tied $45/50$ faithfulness pairs while another never tied). 
When both raters committed to a direction on the concept dimension, they agreed in 68\% of comparisons (153/226; 50--87\% across the six rater pairs), indicating that the low $\kappa$ reflects heterogeneous tie-calling together with genuine disagreement concentrated on borderline cases.
We therefore report aggregate marginal preferences rather than per-item agreement as the primary clinician validation signal.

\subsection{Verbalization advantage from concept labels}
\label{app:label_ablation}

In the pairwise comparison of verbalized explanations, \tisae{} provides dictionary labels while baselines provide attributed cells (Section~\ref{sec:layer3}); here we isolate what the labels contribute.
We ran two controls (same LLM verbalizer and judge, 50 episodes):
\emph{labels removed} strips the label and description from \tisae{}'s concepts, keeping variables, temporal windows, weights, and per-window activation values;
\emph{+ cross-episode summary, no label} adds back only the label-free cross-episode characterization of each concept (cohort frequency, outcome odds ratio, typical temporal positions).
We focus on the concept dimension, where the label-inflation concern applies; without labels the concept format is also rated less actionable than per-cell attributions, and on faithfulness the two are mixed and tie-heavy.

\begin{table}[h]
\centering
\caption{Label ablation on \tisae{}'s input. Win counts: \tisae{} condition--baseline (ties), LLM judge, $N{=}50$ per comparison.}
\label{tab:label_ablation}
\small
\setlength{\tabcolsep}{4pt}
\begin{tabular}{l ccc}
\toprule
\tisae{} condition & vs.\ IG & vs.\ Attention & vs.\ TSR \\
\midrule
Label removed & 24--23 (3) & 16--29 (5) & 25--23 (2) \\
+ cross-episode summary, no label & 31--19 (0) & 26--20 (4) & 27--21 (2) \\
\bottomrule
\end{tabular}
\end{table}

With its label removed, \tisae{} no longer clearly beats raw attribution cells, so most of the verbalized advantage came from the label.
Adding back only the label-free cross-episode summary recovers a small concept lead. However, as noted in Section~\ref{sec:results_verbalization}, the labels are themselves generated from cross-patient exemplar evidence that only the shared dictionary provides.

In addition, we compare \tisae{} against different baseline conditions: \emph{LLM-summarized (per-episode)} lets the same LLM first group the baseline's attributed cells into per-episode concepts (channel, variables, time window, key values) before verbalization; \emph{channel-grouped, no label} groups the same cells into organ-channel concepts, isolating structure without LLM wording.

\begin{table}[h]
\centering
\caption{\tisae{} vs.\ different baseline conditions. Win counts: \tisae{}--baseline (ties), LLM judge, $N{=}50$ per comparison.}
\label{tab:matched_baseline}
\small
\setlength{\tabcolsep}{4pt}
\begin{tabular}{l ccc}
\toprule
Baseline condition & vs.\ IG & vs.\ Attention & vs.\ TSR \\
\midrule
LLM-summarized (per-episode) & 11--39 (0) & 11--37 (2) & 18--30 (2) \\
Channel-grouped, no label & 27--21 (2) & 29--20 (1) & 28--20 (2) \\
\bottomrule
\end{tabular}
\end{table}

Given a fresh per-episode LLM summary, the baseline beats \tisae{} on concept. However, that summary is a per-patient description the LLM can word freely for each case. Against the matched label-free baseline that groups the same cells into organ-channel concepts (isolating structure, not LLM wording), \tisae{} wins across all three methods.

\section{Example explanations}
\label{app:examples}

Below are verbalized explanations for representative high-risk and low-risk episodes, generated by the same verbalizer.

\paragraph{High-risk example (episode 23415, $p{=}0.646$).}
\hfill

\smallskip\noindent\textbf{Attention:}
\begin{quote}\small
This patient carries a 64.6\% predicted mortality risk driven by several converging physiological concerns over the 48-hour ICU stay. The most influential features include a low temperature at hour 47 (36.06\textdegree C), suggesting early hypothermia, and persistently impaired neurological status, with GCS eye opening of 3 and verbal score of 1 at multiple time points despite ongoing sedation with midazolam and fentanyl, making true
neurological assessment difficult. The patient required prolonged mechanical ventilation and vasopressor support with neosynephrine throughout the stay, indicating refractory hemodynamic instability. Markedly elevated neutrophils (92.8\%) with reciprocally low lymphocytes (4.9\%) at hour 15 reflect a severe stress or septic response. Persistently low urine output, ongoing broad-spectrum antibiotics, amiodarone use suggesting arrhythmia burden, and elevated heart rates up to 160 bpm further compound the risk.
\end{quote}

\smallskip\noindent\textbf{\tisae{}:}
\begin{quote}\small
This patient carries a 64.6\% predicted mortality risk driven by five converging critical illness patterns. First, persistent hypothermia early in the admission (temperatures 35.3–35.5°C) signals impaired thermoregulation consistent with severe physiologic stress. Second, the patient demonstrates oliguria with urine outputs frequently 10–30 mL/hr in the latter half of the stay, accompanied by acidemic arterial pH
values as low as 7.23, indicating acute kidney injury with metabolic acidosis. Third, the patient requires sustained mechanical ventilation throughout the 48-hour window with FiO2 ranging from 0.5 to 1.0, reflecting severe hypoxemic respiratory failure. Fourth, neurological assessment reveals a largely unresponsive state with GCS verbal scores of 1 and absent eye opening for much of the admission, indicating profound coma. Fifth, continuous amiodarone infusion signals serious cardiac arrhythmia requiring active suppression. The simultaneous presence of respiratory failure, renal failure with acidosis, coma, hemodynamic instability requiring vasopressors, and cardiac arrhythmia
represents a high-acuity multi-organ dysfunction syndrome portending poor survival.
\end{quote}

\paragraph{Low-risk example (episode 23134, $p{=}0.016$).}
\hfill

\smallskip\noindent\textbf{Attention:}
\begin{quote}\small
This patient is predicted to have a low mortality risk (1.6\%) over the 48-hour ICU stay. Despite markedly elevated transaminases (ALT peaking above 10,000 U/L) and persistently elevated INR consistent with acute hepatocellular injury, several reassuring features drove the model's favorable prediction. Temperature remained consistently normal throughout the stay, suggesting absence of significant infectious or inflammatory
deterioration. Platelet counts were maintained in an acceptable range (164–218), indicating preserved hematopoietic reserve. Albumin levels, while mildly reduced (3.0–3.5 g/dL), remained relatively stable. Neurological status was intact, with GCS verbal and motor scores consistently at 5 and 6 respectively throughout the admission. Urine output was adequate, reflecting preserved renal perfusion. Hemodynamics, though occasionally borderline, largely stabilized over time. The patient was tolerating oral intake by the latter part of the admission. Collectively, these findings suggest a patient with acute liver injury who is nonetheless maintaining end-organ function and neurological integrity, supporting a favorable
short-term prognosis.
\end{quote}

\smallskip\noindent\textbf{\tisae{}:}
\begin{quote}\small
This patient carries a low predicted mortality risk of 1.6\%, driven primarily by sustained neurological and physiologic stability throughout the 48-hour ICU stay. The most protective finding is a consistently maximal GCS motor score of 6 and full verbal responsiveness (GCS verbal 5) maintained across the entire observation window, indicating intact consciousness without deterioration. Eye-opening was spontaneous (GCS eye 4) from approximately hour 9 onward. Platelet counts remained normal throughout, reflecting reserved hematologic function. Urine output was sustained and adequate across all time intervals, suggesting intact renal perfusion. Hemodynamics, while occasionally labile,
generally remained acceptable without vasopressor requirement. Despite concerning laboratory findings — markedly elevated transaminases, elevated bilirubin, significantly prolonged INR, and initially very high LDH consistent with acute hepatocellular injury — these hepatic derangements showed a clear downward trend over 48 hours. The combination of preserved and stable neurological function, adequate renal output, and improving hepatic markers collectively drove the model's low mortality prediction.
\end{quote}

\end{document}